\pdfoutput=1
\documentclass[11pt]{article}

\usepackage{ACL2023}

\usepackage{times}
\usepackage{latexsym}
\usepackage{booktabs}
\usepackage{lipsum}
\usepackage{graphicx}
\usepackage{multirow}
\usepackage{float}

\usepackage[T1]{fontenc}
\usepackage[utf8]{inputenc}

\usepackage{microtype}

\usepackage{inconsolata}

\newcommand{\datalink}{\href{https://www.kaggle.com/datasets/googleai/fleurs-asl}{the original FLEURS-ASL dataset}}

\title{Fingerspelling within Sign Language Translation}

\author{Garrett Tanzer \\
  Google \\
  \texttt{gtanzer@google.com}}

\begin{document}
\maketitle
\begin{abstract}
Fingerspelling poses challenges for sign language processing due to its high-frequency motion and use for out-of-vocabulary terms. While prior work has studied fingerspelling recognition, there has been little attention to evaluating how well sign language translation models understand fingerspelling in the context of entire sentences---and improving this capability. We manually annotate instances of fingerspelling within FLEURS-ASL and use them to evaluate the effect of two simple measures on fingerspelling recognition within American Sign Language to English translation: 1) use a model family (ByT5) with character- rather than subword-level tokenization, and 2) mix fingerspelling recognition data into the translation training mixture. We find that 1) substantially improves understanding of fingerspelling (and therefore translation quality overall), but the effect of 2) is mixed.
\end{abstract}

\section{Introduction}

Fingerspelling is a system used within sign languages to borrow words from spoken languages by spelling them out using a manual alphabet~\citep{fspct}. Each sign language has its own manual alphabet, which is either 1-handed (e.g., in American Sign Language) or 2-handed (e.g., in British Sign Language)~\citep{bsl-fs}. While fingerspelling is only one component of sign languages, it is often an important one; for example, in American Sign Language, fingerspelling constitutes 12-35\% of signing~\citep{fspct}. Fingerspelling poses challenges for receptive sign language processing because a) its hand movements are quick, small, \& highly coarticulated/blended~\citep{patrie} and b) it is primarily used for out-of-vocabulary terms (such as proper nouns or domain-specific vocabulary) that don't have native signs~\citep{fspct}.

While there is a body of prior work studying handshape classification for fingerspelling~\citep{dreuw06smvp,kang2015realtime} and then fingerspelling recognition for entire phrases~\citep{kim2016lexiconfree,fs18slt,fs18iccv,shi-etal-2022-searching,prajwal2022weaklysupervisedfingerspellingrecognitionbritish,fsboard}, there has been little attention on evaluating how well translation models understand \textit{fingerspelling within the context of entire sentences}, and improving this specific important capability. In some ways, one expects this task to be harder than fingerspelling recognition (because it requires full sign language understanding), but in other ways it should also be easier (because the pace/neatness of natural fingerspelling is tailored to the available context and easier to understand given lexical context~\citep{patrie,thumann,wager,reconsideringsentence}.

We study this question---fingerspelling within sign language translation---in the context of translation from American Sign Language to English, trained on a noisy $\sim$2800-hour superset of YouTube-ASL~\citep{youtubeasl} and evaluated on FLEURS-ASL~\citep{fleursasl}. In order to measure the improvement in fingerspelling specifically, we manually annotate all the instances of fingerspelling within FLEURS-ASL\footnote{We have contributed these annotations as an addition to \datalink, under the same CC BY-SA 4.0 license, and present them in tabular form in Appendix~\ref{app:full-fleurs-asl-fs} for convenience.} and score the character-level accuracy of spans within predicted translations that are identified to correspond to the fingerspelled terms. (We automatically identify the relevant spans with an LLM.) We use this setup to evaluate two simple measures:

\begin{figure*}
    \centering
    \includegraphics[width=\textwidth]{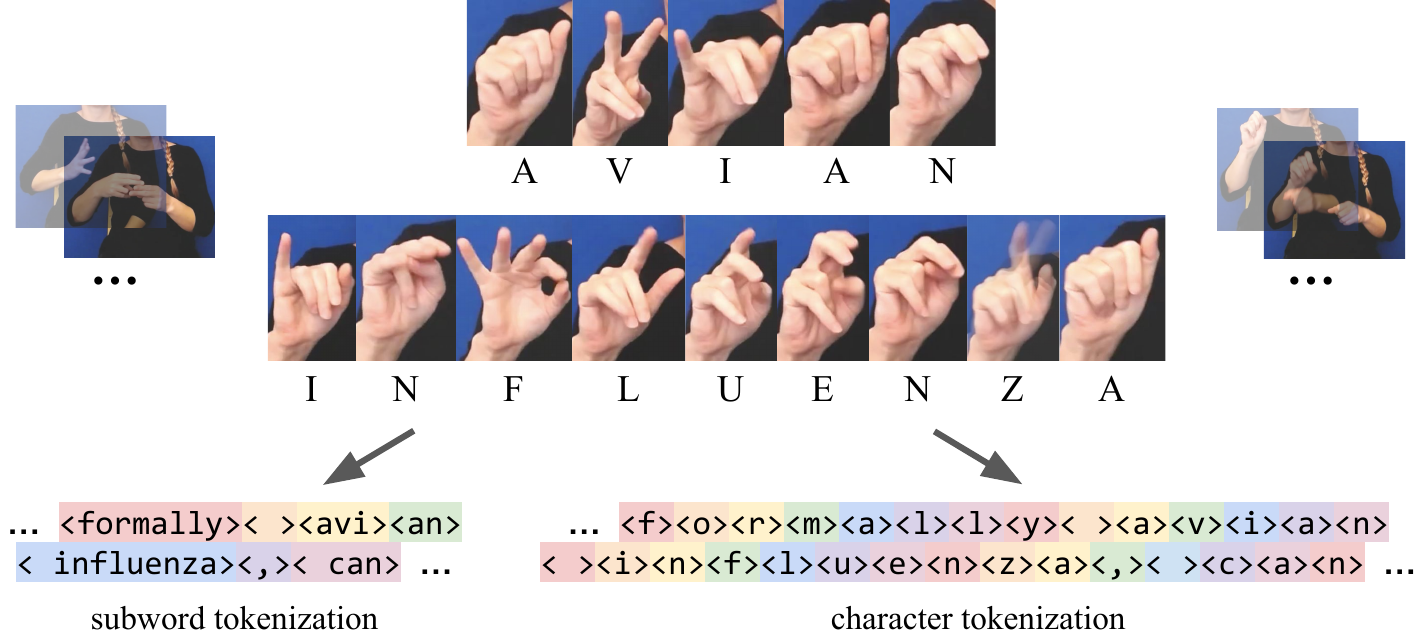}
    \caption{\textbf{Visual depiction of subword- vs. character-level tokenization as it relates to sign language translation.} Color highlights are a visual aid for token boundaries. We omit faces in the figure for privacy. Fingerspelled spans such as ``avian influenza'' within larger sentences must be mapped to sequences like \texttt{<avi><an>< influenza>} in the T5 subword vocabulary; the model can only know how these tokens are spelled through fingerspelling data coverage (less likely) or from text pretraining (more likely). Character-level tokenization makes the mapping much more straightforward, at the expense of increasing sequence length for the rest of the sentence.}
    \label{fig:tok-visualizatino}
\end{figure*}

1) Use a model family with character-level tokenization (ByT5~\citep{xue2022byt5}) rather than subword-level tokenization (T5~\citep{t5}), so the model actually knows how words are spelled.

2) Mix fingerspelling recognition data---specifically FSboard~\citep{fsboard}, a fingerspelling recognition dataset situated in a mobile text entry use case---into the training mixture, to make the most of the limited available resources for ASL.

We find that 1) provides substantial improvements in overall translation quality (38.8 $\rightarrow$ 45.4 BLEURT); more of this gain comes from sentences with fingerspelling (37.3 $\rightarrow$ 44.8 BLEURT) than without (42.5 $\rightarrow$ 46.5 BLEURT); and the accuracy of fingerspelled phrases within those translations improves dramatically (76.6\% CER $\rightarrow$ 41.6\% CER). On the other hand, adding 2) on top of 1) gives mixed or negative results (45.4 BLEURT $\rightarrow$ 45.1 BLEURT, 41.6\% CER $\rightarrow$ 42.1\% CER).

We hope that these findings and the FLEURS-ASL-FS evaluation protocol will be useful for understanding the behavior of fingerspelling within sign language machine translation. In particular, we think that character-level tokenization should become standard practice for sign language translation models, or at least that future work should prioritize achieving the same benefits with even fewer compromises.

\section{Related Work}

In this section we survey related work in three areas: spelling in LLMs, machine learning for fingerspelling, and datasets \& modeling approaches for learning from sign language data with weakly aligned captions.

\subsection{Spelling in LLMs}

LLMs are notoriously bad at spelling---as in, their ability to manipulate the characters within words is poor~\citep{shin2024large}. This is because they virtually always use word-level or subword-level tokenization, which reduces sequence length but obscures the composition of each token. The models can generate perfectly spelled text with ease, but must learn to introspect on spelling information from only cooccurrences between tokens in the training data, such as between ``dog'', ``DOG'', and ``d o g''. This capability, like virtually all others, improves with scale~\citep{itzhak2022models}.

So-called token-free models (models with character-level or byte-level tokenization)~\citep{kim2015characteraware,xue2022byt5,Clark_2022,wang2024mambabyte} expose this information to the model and therefore perform much better at spelling tasks, such as adding spaces between characters in a string and rendering text in images~\citep{liu2023characteraware}. Fingerspelling recognition is much like text rendering, in that it requires the model to establish a correspondence between text tokens and continuous spelled data in another modality. The main downside of token-free models is increased sequence length and therefore higher latency for inference by autoregressive generation, but techniques like speculative decoding~\citep{leviathan2023fast,chen2023accelerating} may mitigate this regression. Besides, current sign language translation models are so far from usability that it would be premature to prioritize inference latency at the expense of quality.

\subsection{Fingerspelling Recognition}

While many prior works study handshape classification for fingerspelling~\citep{dreuw06smvp,isl-fs,kang2015realtime,rsl-fs,arabic-fs} (classifying a single frame for static handshapes or several frames for dynamic handshapes), this is a toy task without significant practical applications. The relevant task that approaches usefulness is continuous fingerspelling recognition (transcribing a clip containing only fingerspelling, but in its complete form with coarticulation etc.), studied by works such as ChicagoFSVid~\citep{kim2016lexiconfree}, ChicagoFSWild~\citep{fs18slt}, ChicagoFSWild+~\citep{fs18iccv}, and FSboard~\citep{fsboard}. Another line of work~\citep{shi-etal-2022-searching,prajwal2022weaklysupervisedfingerspellingrecognitionbritish} expands the task to include fingerspelling span detection within longer videos. These tasks may be useful on their own in narrow contexts, but they are intended primarily as a stepping stone to full translation. All of these works use baseline models with character-level vocabularies; the choice of character-level targets is treated as so obvious that it is barely even discussed.

In the context of sign language translation, on the other hand,~\citet{slt_orig} raise the different options for tokenization granularity but opt to inherit word-level tokenization by analogy to mainstream machine translation work. The paper makes no mention of fingerspelling; it is possible that this was less salient for the RWTH-PHOENIX-2014T dataset, which has a narrow domain (weather forecasts) and limited vocabulary (including extremely small numbers of out-of-vocabulary terms across splits). Since then, methods have become more dependent on pretrained language models~\citep{de-coster-etal-2021-frozen,youtubeasl}, where subword-level tokenization is dominant. To the best of our knowledge, OpenASL~\citep{openasl} is the only work that has evaluated fingerspelling in the context of sentence-level translation (and understandably, as it comes from the same authors as ChicagoFSWild). They find that translation quality is slightly worse on clips with fingerspelling than without fingerspelling, and observe that the task is challenging due to out-of-vocabulary proper nouns (in their case triggering \texttt{<unk>} tokens). They pretrain their vision encoder on sign spotting and (more relevant here) fingerspelling spotting, but the clips come from within the translation dataset itself rather than a separate dataset, so they do not separately mix them into the translation model's training mixture. In contrast, FSboard is strictly a fingerspelling dataset, so we treat its fingerspelling recognition task as a special case of translation and add the new data into our training mixture.

\subsection{Weakly Aligned Sign Language Pretraining}

In order to accommodate weakly aligned data for pretraining, we use the unified multitask approach introduced in the FLEURS-ASL baselines~\citep{fleursasl}, which samples random 34-second video clips from YouTube-ASL videos and predicts timed caption tracks given surrounding text context. The YouTube-SL-25 baselines~\citep{youtubesl25} apply this to data that is slightly noisier than YouTube-ASL but still the result of a human filtering process. We extend this to the noisiest data source yet, the raw unfiltered list of ASL video IDs that was pared down by human annotators in YouTube-ASL~\citep{youtubeasl}.

Other datasets like BOBSL~\citep{bobsl} and SRF23~\citep{wmt_slt_23} have only weak, audio-derived alignments, but they have not been utilized to full effect. A line of work surrounding BOBSL has aimed to realign sentence boundaries by spotting signs or other signals in the video content~\citep{bull1,bull2,momeni,raude}. Meanwhile, in WMT23~\citep{wmt_slt_23} participants simply used the clip boundaries that were known to be misaligned (excluding relevant content or including extraneous content at the boundaries). Instead, the method introduced in FLEURS-ASL aims to tolerate misaligned data at training time without special learned preprocessing.

See YouTube-ASL~\citep{youtubeasl}, FLEURS-ASL~\citep{fleursasl}, or YouTube-SL-25~\citep{youtubesl25} for discussion of different modeling approaches at a lower level of abstraction (input preprocessing, architectures, etc.). We use their methodology without modification. 

\begin{table*}
    \centering
    \begin{tabular}{p{8cm}c}
    \toprule
        \bf reference & \bf fingerspelled phrases  \\
        \midrule
        Bird flu, or more formally avian influenza, can infect both birds and mammals. & \{ {\it flu, avian influenza, mammals} \} \\
        \midrule
        Tournament top seeds South Africa started on the right note when they had a comfortable 26 - 00 win against 5th seeded Zambia. & \{ \} \\
        \midrule
        During the 1980s he worked on shows such as Taxi, Cheers, and The Tracy Ullman Show. & \{ {\it Taxi, Cheers, The Tracy Ullman} \} \\
    \bottomrule
    \end{tabular}
    \caption{\textbf{Examples of annotations from FLEURS-ASL-FS.} Each sentence is paired with an ordered multiset of human-annotated fingerspelled phrases. See the complete set of annotations in Appendix~\ref{app:full-fleurs-asl-fs}.}
    \label{tab:fleurs-asl-fs}
\end{table*}

\section{FLEURS-ASL-FS}

FLEURS-ASL~\citep{fleursasl} is an extension of the massively multilingual FLORES~\citep{flores200} / FLEURS~\citep{fleurs} translation benchmarks for text and speech respectively into American Sign Language video. Because FLORES is drawn from Wikipedia articles, it is relatively rich in proper nouns and other specialized vocabulary, making it a good testbed for fingerspelling.

We annotate each of the 1749 sentences in FLEURS-ASL with the sequence of phrases that are fingerspelled within them; we refer to these annotations as ``FLEURS-ASL-FS''.\footnote{Note that we do not annotate the fingerspelled timespans within the videos, because this kind of annotation is extremely time-consuming.} The first author (a non-native proficient signer) performed these annotations given access to the entire video context and the reference captions; the intent was to recover the ground truth fingerspelled terms, not measure the ability of a human to recognize them from the video alone. Subjectively, this task was easy and not especially time-consuming; it could generally be performed in real time while watching the video at $\geq$1x speed. In the vast majority of cases, these fingerspelled terms were subspans of the reference text, but sometimes the translation into ASL used different or extra terms. The definition of fingerspelling is somewhat fuzzy at the margins, so we made the following decisions on inclusion in annotations:

\begin{itemize}
    \item[(a)] We included fingerspelled abbreviations, such as months like \texttt{NOV}, transcribed in abbreviated form (what was actually spelled) rather than the meaning. However, but we excluded lexicalized signs that have significant movement. For example, we excluded a lexicalized sign for ``back''', which roughly moves from a \texttt{B} to \texttt{K} along a directional path. Some examples of borderline signs are a lexicalized sign for ``company'', \texttt{CO}, and an initialized sign for ``north'', \texttt{N}.
    \item[(b)] Unlike FSboard, we did not consider numbers to be a form of fingerspelling; we included numbers only when they were contiguous with fingerspelled letters. From a modeling perspective, due to issues with learning arithmetic, many tokenizers already force digits to have individual tokens~\citep{chowdhery2022palmscalinglanguagemodeling}, though this is not the case for T5's tokenizer.\footnote{In retrospect, we should have perhaps annotated numbers but labeled them as a separate category from fingerspelling. We hope that future work will be able to annotate FLEURS-ASL with even more granular annotations, which were out of scope for this work.}
\end{itemize}

The result is 2845 total fingerspelled terms, with 72.73\% of sentences having at least one fingerspelled word, and an average of 11.56 fingerspelled characters per sentence. If we disaggregate by signer (following~\citet{reconsideringsentence}), we see some variety in frequency of fingerspelling, from 63\% to 81\% of sentences and 8.80 to 14.11 average fingerspelled characters per sentence.\footnote{Qualitatively, this corresponds to variation in translation strategies across interpreters in several respects (this list is non-exhaustive): a) whether certain words without perfectly corresponding signs are translated with fingerspelling for precision, or translated with a synonymous sign (often mouthing the term at the same time); b) whether proper nouns with less commonly known signs (such as endonyms for certain cities or countries) are fingerspelled or signed (sometimes after introducing with fingerspelling); and c) whether certain fingerspelled functional words like ``by'' or ``of'' are used.} See Table~\ref{tab:fleurs-asl-fs} for examples of the FLEURS-ASL-FS annotations and Appendix~\ref{app:disagg} for disaggregated summary statistics and more analysis.

\section{Experiments}
\label{sec:evaluation}

In this section, we describe our experiments on two simple measures that could improve the quality of fingerspelling within sign language translation: using character-level tokenization and jointly training on independently collected fingerspelling data.

\subsection{Setup}
\label{subsec:training}

We use two training datasets:
\begin{enumerate}
    \item FSboard~\citep{fsboard}, a $>$250-hour ASL fingerspelling recognition dataset recorded from smartphone selfie cameras, or ``FS'' for short in tables. This data was elicited specifically for fingerspelling recognition, rather than being clipped from full signing data~\citep{fs18slt,fs18iccv}.
    \item An internal dataset that we refer to as ``noisy YouTube-ASL'', or ``YT'' for short in tables. This is the $\sim$2800-hour pre-human filtering superset of YouTube-ASL from~\citet{youtubeasl}, with the sign language heuristically classified as ASL based on metadata as in~\citet{youtubesl25}. We use only ASL-English data to avoid conflation with sign-multilinguality \& text-multilingual pretraining, and we use a larger (albeit noisier) dataset than the final YouTube-ASL dataset to achieve higher-quality results where we can reliably identify text spans corresponding to fingerspelled phrases.
\end{enumerate}

We use the same modeling setup as the FSboard and FLEURS-ASL baselines. In brief, 85 3D MediaPipe Holistic~\citep{lugaresi2019mediapipe,mediapipeholistic} are linearly projected into a T5-class model. For FSboard, these inputs are fingerspelled phrases at 15 Hz and the targets are the transcription. For noisy YouTube-ASL, these inputs are random clips up to 34 seconds long at half frame rate, where control tokens and caption track context beyond the clip are provided to accommodate weakly aligned captions. We pretrain on noisy YouTube-ASL then finetune on the 1:1 mixture of caption-level and multitask clean YouTube-ASL data from FLEURS-ASL.\footnote{Note that results are still ``zero-shot'' after this finetuning, because it is just continued training on a subset of the pretraining data, not the downstream benchmark data.}

For fingerspelling recognition, we evaluate on FSboard's test set (which has nonoverlapping signers and phrases with the train set) using the same metrics as its baselines: Character Error Rate (CER, $\downarrow$) and phrase-level top-1 accuracy ($\uparrow$).

For translation, we evaluate zero-shot sentence-level translation on FLEURS-ASL, measuring quality with BLEURT(-20)~\citep{sellam2020bleurt,pu2021learning} (overall, and for the subset of phrases with and without fingerspelling) and fingerspelling recognition-within-translation with CER (using \href{https://www.tensorflow.org/api_docs/python/tf/edit_distance}{TensorFlow's implementation}~\citep{tensorflow2015-whitepaper}) on spans within predicted translations that are detected to correspond with the FLEURS-ASL-FS annotations. In order to make the span extraction for relevant spans scalable and reproducible, we use an LLM, Gemini 1.5 Pro, as the annotator rather than humans; see Figure~\ref{fig:prompt-excerpt} for a summary.\footnote{Note that we do not expect all fingerspelled terms to be reflected faithfully in translation outputs, such as abbreviations. These tend to be extremely short, often closed-vocabulary terms which do not exercise the open-vocabulary challenges we wish to test. We manually create a denylist of fingerspelled terms that may reasonably not be translated in the literal way (e.g., ``mph'' may be translated into ``mph'' or ``miles per hour'') and exclude them from the CER evaluation. We still release annotations for these terms in case they are useful to others.}
Because large language models such as Gemini are not unambiguously superior to human raters, and in particular because they conventionally do not use character-level tokenization~\citep{gemmateam2024gemmaopenmodelsbased}, we expect that they will cause downstream overestimates of character-level error by missing or mislabeling corresponding sequences. However, these estimates should be unbiased and can be used to compare experimental conditions relative to each other. See Appendix~\ref{app:prompt} for full details, including validation of the autorater quality.

\begin{figure}
\begin{verbatim}
Reference: {reference}
Prediction: {prediction}

Fingerspelled terms:
1. {term 1}
...
N. {term n}

Output:
1. [span for term 1]
...
N. [span for term N]
\end{verbatim}
\caption{\textbf{Framing for the span extraction task performed by an LLM in our evaluation framework}. The LLM sees a) a reference translation, b) the SLT model's predicted translation, and c) a list of fingerspelled terms known to be in the original video. Then it identifies the spans within the predicted translation that best correspond to the fingerspelled terms (or \texttt{''''} for terms where none are found). The role of the fingerspelled terms within the reference sentence often helps to identify the correspondence in the predicted translation when the character-level correspondence is weak. See Appendix~\ref{app:prompt} for the full prompt, which includes instructions and 3 examples in context.}
\label{fig:prompt-excerpt}
\end{figure}

\subsection{Ablations}
\label{subsec:ablations}

We run ablations to answer the following four research questions:
\begin{enumerate}
    \item \textbf{RQ1:} \textit{What happens if we train a fingerspelling recognition model with a subword vocabulary?}\\
    \textbf{Ablation:} Compare the ByT5-Small baseline from~\citet{fsboard} with a new T5-Small baseline.
    \item \textbf{RQ2:} \textit{What happens if we train a sign language translation model with a character-level vocabulary?} \\
    \textbf{Ablation:} Compare T5-Base vs. ByT5-Base trained on noisy YouTube-ASL.
    \item \textbf{RQ3:} \textit{What happens if we train a sign language translation model on fingerspelling recognition data too?} \\
    \textbf{Ablation:} Compare ByT5-Base trained on noisy YouTube-ASL vs. ByT5-Base trained on a mixture of noisy YouTube-ASL and FSboard (weighted by number of examples). After all, fingerspelling recognition is a special case of translation.
    \item \textbf{RQ4:} \textit{How well do fingerspelling capabilities learned from full translation data transfer back to fingerspelling recognition?} \\
    \textbf{Ablation:} Finetune the above pretrained translation models on FSboard and compare performance to baselines.
\end{enumerate}

We pretrain the translation models for 1m steps (continuing from the pretrained text checkpoints) at batch size 128 and learning rate 0.001 with Adafactor~\citep{shazeer2018adafactor} and validation BLEU as a checkpoint selection criterion. Each 10k steps took about 64 TPUv3-hours. For fingerspelling recognition models, we train for 100k steps (without YouTube pretraining) or finetune for 10k steps (with YouTube pretraining) at batch size 64 and learning rate 0.001 with Adafactor and validation CER as a checkpoint selection criterion. Each 10k steps took about 24 TPUv4-hours. These hyperparameters were chosen based on prior work without additional optimization.

\begin{table*}
\begin{minipage}{0.45\textwidth}
    \centering
    \setlength{\tabcolsep}{5pt} 
    \begin{tabular}{lccc}
    \toprule
    \bf model & \bf CER ($\downarrow$) & \bf 1-acc ($\uparrow$) \\
    \midrule
    T5 (FS) & 45.9\% & 14.6\%\\
    ByT5 (FS) & \bf 11.3\% & \bf 53.1\%\\
    \midrule
    T5 (YT$\rightarrow$FS) & 14.4\% & 39.2\% \\
    ByT5 (YT$\rightarrow$FS) & \bf 8.9\% & \bf 59.3\% \\
    ByT5 (YT+FS$\rightarrow$FS) & 9.0\% & \bf 59.3\% \\
    \bottomrule
    \end{tabular}
    \caption{\textbf{Quantitative results for ASL fingerspelling recognition on FSboard} measured in Character Error Rate (CER, $\downarrow$) and phrase-level top-1 accuracy ($\uparrow$). There are two sets of ablations: T5-Small vs. ByT5-Small trained only on FSboard, vs. finetuned on FSboard after YouTube pretraining.}
    \label{tab:fs-results}
\end{minipage}\hfill%
\begin{minipage}{0.51\textwidth}
\centering
    \begin{tabular}{lcc}
    \toprule
    \bf model & \bf overall (FS / no-FS) (CER) \\
    \midrule
    T5 (YT) & 38.8 (37.3 / 42.5) (76.6\%) \\
    ByT5 (YT) & \textbf{45.4} (\textbf{44.8} / 46.5) (\textbf{41.6\%}) \\
    ByT5 (YT+FS) & 45.1 (44.3 / \textbf{46.6}) (42.1\%) \\
    \bottomrule
    \end{tabular}
    \caption{\textbf{Quantitative results for zero-shot sentence-level ASL to English translation on FLEURS-ASL.} We report overall BLEURT ($\uparrow$), (in parentheses) BLEURT on the subset of sentences with / without fingerspelling, and then and (in parentheses) CER ($\downarrow)$ for the spans of the translation that are identified as corresponding to the FLEURS-ASL-FS fingerspelling annotations by an LLM.}
    \label{tab:trans-results}
\end{minipage}
\end{table*}

\begin{table*}
    \centering
    %\small
    \fontsize{8.4}{10}\selectfont
    \begin{tabular}{ccccc}
    \toprule
    \bf Reference & \bf T5 (FS) & \bf ByT5 (FS) & \bf T5 (YT$\rightarrow$FS) & \bf ByT5 (YT$\rightarrow$FS) \\
    sparks nv & paula novak & aparkan2 & aparta nv & sparks nv \\
    kristopher mahoney & kristopher marsh & kerista horney & kiris-permaney & kristen mahoney \\
    +54-1778-53-128-8661 & +54-1778-53-1268-867 & +54-1778-55-1268-8661 & +54-1778-53-127-866 & +54-1778-53-126-8661 \\
    leroy atkinson & jerry adkins & leroya turner & terry atkinson & leroy atkinson \\
    fort lauderdale florida & fort worth tx & 90 lauderdale florida & fort lauderdale florida & fort lauderdale florida \\
    joliet illinois & jolie elliott & joliet illins & joliet llinois & jolliet illinois \\
    \bottomrule
    \end{tabular}
    \caption{\textbf{Qualitative examples of fingerspelling recognition from the FSboard test set.} Examples are selected randomly from those where at least one ByT5 model does not predict a perfect output in order to show the gradation across models; note that this biases the selections against ByT5. We also exclude clips where the ground truth is wrong, e.g. one example has the target ``798 lumber jack ln'' but ``798'' is missing from the video clip, so all models fail and we omit it.}
    \label{tab:qual-examples-fs}
\end{table*}

\begin{table*}
    \centering
    %\small
    \fontsize{8.4}{10}\selectfont
    \begin{tabular}{lp{13.3cm}}
    \toprule
    \bf Reference & Bird \textit{flu}, or more formally \textit{avian influenza}, can infect both birds and \textit{mammals}. \\
    T5 (YT) & The Fever LORD is a formal name. It can be spelled. \\
    ByT5 (YT) & 20 Former name Avian Influenza can affect both birds and mamals. \\
    ByT5 (YT+FS) & Twitter FLU, formally known as Avian Influenza, can affect both birds and mammals. \\     
    Human & The bird flu, formally known as the Avian influenza, affects both birds and mammals. \\
    \midrule
    \bf Reference & Tournament top seeds South Africa started on the right note when they had a comfortable 26 - 00 win against 5th seeded Zambia. \\
    T5 (YT) & In South Africa, there is a perfect chance of winning the Legislative Awareness Day (LAD)! \\
    ByT5 (YT) &  The tournaments in South Africa were perfect for the winning tournament against the 26th and the 5th place. \\
    ByT5 (YT+FS) & The South African tournament was perfect for winning the 26th East African-American basketball tournament.\\     
    Human & In the tournament bracket, South Africa moved forward strong and won 26 to 0 against Zambia, who was 5th. \\
    \midrule
    \bf Reference & During the 1980s he worked on shows such as \textit{Taxi}, \textit{Cheers}, and \textit{The Tracy Ullman} Show. \\
    T5 (YT) & During the 1980s, she worked on a show called Axi-Cheers and The Tray Royal Man Theater. \\
    ByT5 (YT) &  During the 1980s, they worked in theatre such as Taxi, Chern, The Tracy Ullman Show. \\
    ByT5 (YT+FS) & During the 1980's, there were shows such as Taxi, Cheers, and The Tracy Ullman Show. \\     
    Human & During the 1980s, they worked in shows like Taxi, Cheers, and The Tracy Ullman show. \\
    \bottomrule
    \end{tabular}
    \caption{\textbf{Qualitative examples of sentence-level ASL to English translation on FLEURS-ASL.} Fingerspelled spans are italicized. Examples are randomly selected from FLEURS-ASL's human baseline. Fingerspelling recognition improves dramatically from T5 to ByT5, bringing the adequacy of the translations much closer to human-level, but there are still significant gaps. For example, in the first example the sign for ``bird'' and ``Twitter'' are the same (and almost ``20'', though the location is not the same), but the choice of word should be clear from context. In the second example, the model also does not recognize uncommon signs like for ``Zambia'', and struggles to understand the grammatical structure of the comparison between teams.}
    \label{tab:qual-examples-trans}
\end{table*}

\subsection{Results}
\label{subsec:results}

\paragraph*{RQ1: Subword-tokenized fingerspelling.}

See Table~\ref{tab:fs-results} for quantitative results for fingerspelling recognition on FSboard, comparing our reproduction of~\citet{fsboard}'s ByT5-Small baseline to the same method using T5-Small instead. See Table~\ref{tab:qual-examples-fs} for a qualitative sample of outputs. Changing T5 to ByT5 massively improves results, achieving 45.9\% CER $\rightarrow$ 11.3\% and 14.6\% $\rightarrow$ 53.1\% top-1 accuracy.

\paragraph*{RQ2: Character-level translation.}

See Table~\ref{tab:trans-results} for quantitative results for ASL to English translation on FLEURS-ASL, Table~\ref{tab:qual-examples-trans} for a qualitative sample of outputs. See Appendix~\ref{app:disagg} for results disaggregated for different signers and Appendix~\ref{app:how2sign} for results on How2Sign~\citep{how2sign}, including BLEU scores. Again, changing T5 to ByT5 majorly improves results, achieving 38.8 $\rightarrow$ 45.4 BLEURT on zero-shot sentence-level translation and 76.6\% CER $\rightarrow$ 41.6\% CER on transcriptions of fingerspelling within the translations. Qualitatively, the fingerspelling recognition quality goes from being completely unusable to generally correct; most of the remaining error seems to be from terms that were omitted entirely in the translation or that were not detected by the autorater. ByT5 also converges in many fewer steps than T5, taking about 350k steps vs. 900k steps.

Like OpenASL, we find that translation quality is initially worse on sentences with fingerspelling than without (37.3 vs. 42.5 BLEURT), though this gap is smaller when measured with BLEU (4.1 vs. 4.2). More of the BLEURT gain from ByT5 comes from sentences with fingerspelling (37.3 $\rightarrow$ 44.8 BLEURT) than without (42.5 $\rightarrow$ 46.5 BLEURT). The difference is even more stark when measured with BLEU, (4.1 $\rightarrow$ 6.3 vs. 4.1 $\rightarrow$ 4.4), where sentences with fingerspelling actually score higher post-ByT5 than those without. It is not clear why there is a discrepancy between the metrics, but we speculate that BLEU is more sensitive to exact matches in spelling because it uses discrete n-gram matching, whereas BLEURT is a neural metric using a language model that itself uses subword-level tokenization. We speculate that the improvement in scores for sentences without fingerspelling could be attributed to factors such as improved translation of numbers (due to T5's lack of per-digit tokenization), improved ability to learn non-fingerspelling aspects of training data with fingerspelling due to better understanding of context, or improved ability to learn from data (such as YouTube captions) with noisy formatting, as seen in~\citet{xue2022byt5}.

\paragraph*{RQ3: Cotraining with fingerspelling recognition.}

Adding FSboard into the translation training mixture gives mixed results on FLEURS-ASL, regressing from 45.4 $\rightarrow$ 45.1 BLEURT on zero-shot sentence-level translation and 41.6\% $\rightarrow$ 42.1\% CER, though the models trade off the lead in slices of the data. There is no unambiguous benefit to integrating the fingerspelling recognition data, unlike using ByT5. We speculate that this could be due to a domain mismatch between FSboard's smartphone fingerspelling and fingerspelling within full signing, or because even 250 additional hours is a relatively small fraction of the YouTube data.

\paragraph*{RQ4: Transfer from translation back to fingerspelling.} When finetuning on FSboard after noisy YouTube pretraining, moving from T5 to ByT5 again dramatically improves results (14.4\% $\rightarrow$ 8.9\% CER and 39.2\% $\rightarrow$ 59.3\% top-1 accuracy). This represents a 21\% relative CER reduction and 13\% top-1 error reduction vs. the 11.3\% CER and 53.1\% accuracy of baselines trained only on FSboard. The gap between T5 and ByT5 is much narrower with YouTube pretraining than it was for \textbf{RQ1} when training on FSboard only, though ByT5 trained on FSboard alone still outperforms T5 pretrained on noisy YouTube-ASL. There is no appreciable difference when including FSboard in pretraining vs. finetuning only.

\section{Conclusion}
In this paper we examined the role of fingerspelling within sign language machine translation, using FSboard, noisy YouTube-ASL, and FLEURS-ASL with new fingerspelling annotations. We found that ByT5 (character-level tokenization) substantially improves upon T5 (subword-level tokenization) on continuous ASL fingerspelling recognition, ASL fingerspelling recognition within translation, and ASL to English translation overall. Cotraining translation models on fingerspelling recognition data, on the other hand, provides mixed gains. We hope that these results will encourage researchers to use and improve upon character-level tokenization for sign language translation models, and that the community will continue to reevaluate sign language modeling decisions that were inherited from mainstream machine translation, where the tradeoffs may no longer be relevant or have different balance in this domain~\citep{desai2024systemic}.

\section*{Limitations}

While the main distinction between T5 and ByT5 is the token vocabulary, there are a variety of other differences between the models (multilingual pretraining, parameter allocation between the encoder vs. decoder, number of embedding vs. dense model parameters, number of parameters at each model size), many optimized to better serve character-level modeling. These factors are difficult to ablate without pretraining new ByT5 variants, which is expensive. We leave more comprehensive experiments across scales to future work.

We evaluate only American Sign Language, which has relatively high usage of fingerspelling compared to other sign languages~\citep{fspct,fs-langs}. We expect that character-level modeling would benefit fingerspelling in other sign languages, but it would have a reduced effect on overall translation quality to the extent that fingerspelling is less frequent.

We evaluate only sign language understanding, not sign language generation (which is less mature). We expect that results there would be similar, since the model needs to know how English words are spelled in order to produce fingerspelling for them; this is even more analogous to~\citet{liu2023characteraware}'s results for text-conditioned image generation than the results in this paper.

\section*{Ethics Statement}
We use MediaPipe Holistic landmarks as a form of anonymization, but more work is needed to establish rigorous best practices for private sign language pretraining. Research on machine learning for sign languages has the potential to improve access to technology for Deaf/Hard of Hearing signers, but with equal access comes equal exposure to potential negative consequences such as automated bias and misinformation. On balance, we believe that sign language technology will eventually have a large positive social impact; in the meantime, it is important not to overstate the quality of the models that currently exist. This paper significantly improves the quality of fingerspelling within sign language translation, but many limitations remain with translation quality overall.

\section*{Acknowledgements}
We thank Manfred Georg, David Uthus, Biao Zhang, and the T5X team for general help with infrastructure and Chris Dyer and Caroline Pantofaru for giving feedback on drafts of this paper.

\bibliography{custom}
\bibliographystyle{acl_natbib}

\appendix

\newpage

\section{FLEURS-ASL-FS Evaluation Protocol}
\label{app:prompt}

The process we used to automatically evaluate fingerspelling recognition within sign language translation using the FLEURS-ASL-FS annotations is as follows.

First, we filtered out annotations for fingerspelled terms from the following denylist. These were selected because we do not expect a correct translation output to include these terms verbatim, or it is ambiguous whether the translation should include this term vs. an expanded version of the abbreviation, so they would introduce noise into the automatic scoring. These tend to be short and relatively closed-vocabulary terms, which makes them poorly suited to measuring character error rate of fingerspelling in generality.

\begin{verbatim}

of, or, at, by, if, did, the, so, up, mps,
mph, kmph, km/h, km, cm, mm, inch, ft, kg,
oz, per oz, sec, apt, ref, dx, co, ed, lit,
supint, supt, div, fs, de, Cs, I, Dr., Th,
Jan, Feb, Aug, Sept, Oct, Nov, Dec, PM, MP,
TV, MD, ID, HK, MS, CL, BW, ABC, HW, SW,
PO, TB, AC, FY, XC, CNS, UN, WW1, WW2, LA,
SF, Albuq, Nev, NC, PA, SC, US, EU, Tas,

\end{verbatim}

Next, we used an LLM, Gemini 1.5 Pro (version \texttt{gemini-1.5-pro-001}), to automatically extract spans for each FLEURS-ASL example that correspond to the filtered FLEURS-ASL-FS annotations. We give the complete prompt below. The model gets the reference translation, predicted translation, and reference fingerspelled terms as input and predicts a list of spans in the prediction corresponding to these fingerspelled terms. The in-context examples were selected iteratively in response to errors that the model made, primarily to improve performance on the T5 outputs, where the correspondence between fingerspelled terms and the predicted translation was very weak.

\begin{verbatim}

I want you to help me analyze some errors
in sign language translation. I'm going to
give you a reference translation for a
clip, then the model's predicted
translation for that clip. Then I'm going
to give you a numbered list of terms that
were fingerspelled in the source video
(usually also the reference translation),
and I want you to find the strings you
think correspond to them in the predicted
translation. Make the output a numbered
list with each entry in double quotes. If
you can't find a correspondence, just
output "" for that term. Output the list
and no other text.


Examples:


Reference: As with respiratory problems
in colder climates, intestinal problems
in hot climates are fairly common and in
most cases are distinctly annoying but
not really dangerous.
Prediction: Respiratory problems in cold
weather include insidious problems, hot
weather, and common situations.

Fingerspelled terms:
1. intestinal

Output:
1. "insidious" 


Reference: Two songs from the movie,
Audition (The Fools Who Dream) and City
of Stars, received nominations for best
original song. Lionsgate studio received
26 nominations — more than any other
studio.
Prediction: Two daughters of Adrian have
been tested for salmonella and the city
of St. Louis' soccer star have received

Fingerspelled terms:
1. Audition (The Fools Who Dream)
2. City of Stars
3. Lionsgate studio
4. studio

Output:
1. Adrian
2. "city of St. Louis' soccer star"
3. ""
4. ""


Reference: Persian has a relatively
easy and mostly regular grammar.
Prediction: Persyan’s language is
easy to pick up and pick up.

Fingerspelled terms:
1. Persian

Output:
1. "Persyan" 


Task:


Reference: {reference}
Prediction: {prediction}

Fingerspelled terms:
1. {term 1}
...
N. {term n}
\end{verbatim}

Luckily, Gemini 1.5 Pro always produced outputs in the correct format, so it was straightforward to parse them into a structured representation. Finally, we scored the character error rate (CER) of the predicted terms against the references.

Qualitatively, the failure cases of the end-to-end framework include:
\begin{itemize}
    \item The autorater hallucinates spans that were not in the predicted translation.
    \item The autorater predicts an incomplete span.
    \item The autorater detects no span corresponding to a given phrase, but human judgment suggests that there is an appropriate span (typically one with high character error rate, but better than 100\%).
    \item The predicted translation includes a correct rendition of a fingerspelled term in an unexpected format, despite our best attempts with the denylist.
\end{itemize}

These errors make it difficult to interpret the final CER scores output from the framework in absolute, but relative comparison between experimental settings is still valid. There are clear opportunities to improve this kind of rubric-based autorater with more work, such as many more in-context examples, or the autorater may automatically improve with future language models. However, the accuracy of the autorater is more than sufficient to draw out the distinctions between models relevant to this paper.
%We audit this autorater by comparing with the first author as a human rater on a sample of 25 sentences, and the resulting CER is within \todotext{X}\% of the LLM-derived estimate across all settings.

\section{Disaggregated FLEURS-ASL Analysis}
\label{app:disagg}

See Table~\ref{tab:statistics-disagg} for summary statistics of FLEURS-ASL-FS, including disaggregation across signers. See Table~\ref{tab:trans-results-disagg} for zero-shot sentence-level translation results on FLEURS-ASL, again disaggregated across signers.

We see some variation across signers in the fingerspelling annotations in Table~\ref{tab:statistics-disagg}, in line with the scale of variation seen in the original FLEURS-ASL description~\citep{fleursasl}, still much less than in datasets like How2Sign~\citep{reconsideringsentence}. Signers \#1 and \#2 fingerspell markedly less often than the others, which matches qualitative impressions from watching the videos. Signer \#2 and Signer \#4 use fingerspelled functional words like ``at'' or ``by'' more often than the other signers, reflected in the lower average fingerspelled term length.

In the experiment results in Table~\ref{tab:trans-results-disagg}, the general trend (especially when looking at BLEU rather than BLEURT) is that gains are larger for signers who fingerspell more. We also see interesting variation in the within-translation CER across signers; Signer \#1 and \#3 have markedly higher CER than the others. Qualitatively, these signers fingerspell faster than the others, especially Signer \#3, and often in orientations that aren't strictly front-facing. The speed and orientation may be less in-domain in YouTube pretrainig data, or fundamentally more challenging of a task. It is possible that performing better on these signers requires modeling the full 30 Hz frame rate, rather than 15 Hz, and perhaps some of the information is just irrecoverable, even for a perfect model at full frame rate. (The human baseline from FLEURS-ASL does not include enough examples to robustly measure this, and it was not performed with frame-by-frame analysis but rather normal viewing.)

\begin{table*}[!t]
    \centering
    \begin{tabular}{lcccc}
    \toprule
         & frac of sents w/ fs & avg fs terms / sent & avg fs chars / sent&  avg fs term len \\
        \midrule
        FLEURS-ASL & 0.73 & 1.63 & 11.56 & 7.11 \vspace{1mm} \\
        \it signers \\ 
        \hspace{4mm}\#0 &  0.77 & 1.82 & 14.11 & 7.74 \\
        \hspace{4mm}\#1 &  0.67 & 1.21 & 8.80 & 7.25 \\
        \hspace{4mm}\#2 &  0.63 & 1.50 & 9.92 & 6.60 \\
        \hspace{4mm}\#3 &  0.78 & 1.81 & 12.86 & 7.12 \\
        \hspace{4mm}\#4 &  0.81 & 1.63 & 13.51 & 6.12 \\
    \bottomrule
    \end{tabular}
    \caption{\textbf{Summary statistics for FLEURS-ASL-FS annotations}, disaggregated by signer. We report the fraction of sentences with any fingerspelled terms, the average number of fingerspelled terms per sentence (including those with none), the average number of fingerspelled characters per sentence, and the average fingerspelled term length in characters.}
    \label{tab:statistics-disagg}
\end{table*}

\begin{table*}[!t]
    \centering
    \begin{tabular}{lccc}
    \toprule
         & \bf T5 (YT) & \bf ByT5 (YT) & \bf ByT5 (YT+FS) \\
        \midrule
        & \multicolumn{3}{c}{BLEU: overall (with FS / without FS)} \\
        \midrule 
        FLEURS-ASL & 4.1 (4.1 / 4.2) & \textbf{5.8} (\textbf{6.3} / 4.4) & 5.6 (6.0 / \textbf{4.5}) \vspace{1mm} \\
        \it signers \\ 
        \hspace{4mm}\#0 &  5.4 (5.4 / 5.1) & \textbf{8.3} (\textbf{8.9} / 5.4) & 7.7 (8.3 / \textbf{5.6}) \\
        \hspace{4mm}\#1 &  3.0 (2.9 / 3.3) & \textbf{3.9} (\textbf{4.1} / \textbf{3.4}) & 3.6 (3.7 / 3.3) \\
        \hspace{4mm}\#2 &  4.6 (4.3 / 5.1) & 6.3 (6.7 / 5.3) & \textbf{6.6} (\textbf{7.1} / \textbf{5.7}) \\
        \hspace{4mm}\#3 &  6.0 (5.9 / 6.4) & \textbf{8.6} (\textbf{8.3} / \textbf{9.9}) & 8.3 (8.3 / 8.4) \\
        \hspace{4mm}\#4 &  6.0 (6.6 / 4.2) & \textbf{9.3} (\textbf{10.3} / 4.7) & 8.9 (9.7 / \textbf{5.9}) \\
    \midrule

        & \multicolumn{3}{c}{BLEURT: overall (with FS / without FS)} \\
        \midrule
        FLEURS-ASL & 38.8 (37.3 / 42.5) & \textbf{45.4} (\textbf{44.8} / 46.5) & 45.1 (44.3 / \textbf{46.6}) \vspace{1mm} \\
        \it signers \\ 
        \hspace{4mm}\#0 &  39.1 (38.0 / 42.9) & \textbf{48.2} (\textbf{47.9} / \textbf{49.3}) & 47.6 (47.3 / 48.7) \\
        \hspace{4mm}\#1 &  36.3 (34.6 / 39.7) & \textbf{41.6} (40.8 / \textbf{43.3}) & 41.4 (40.4 / 43.3) \\
        \hspace{4mm}\#2 &  41.6 (41.6 / 47.3) & \textbf{48.9} (\textbf{48.3} / 50.1) & 48.9 (47.7 / \textbf{51.1}) \\
        \hspace{4mm}\#3 &  41.6 (39.5 / 49.3) & \textbf{48.3} (\textbf{47.0} / 53.1) & 48.0 (46.6 / \textbf{53.1}) \\
        \hspace{4mm}\#4 &  40.6 (39.6 / 44.5) & 48.4 (48.0 / \textbf{50.3}) & \textbf{48.4} (\textbf{48.3} / 49.1) \\
    \midrule
        & \multicolumn{3}{c}{CER / \% no terms with no match found} \\
        \midrule
        FLEURS-ASL & 76.6\% / 53.1\% & \textbf{41.6\%} / \textbf{26.6\%} & 42.1\% / 28.0\% \vspace{1mm} \\
        \it signers \\ 
        \hspace{4mm}\#0 & 74.4\% / 47.3\%  & \textbf{31.7\%} / \textbf{20.8\%}  & 34.5\% / 22.4\%  \\
        \hspace{4mm}\#1 & 79.6\% / 60.7\%  & 55.0\% / 33.0\%  & \textbf{53.6}\% / \textbf{32.9}\%  \\
        \hspace{4mm}\#2 & 75.0\% / 49.3\%  & 35.7\% / \textbf{24.7}\%  & \textbf{35.3}\% / 28.7\%  \\
        \hspace{4mm}\#3 & 82.6\% / 56.8\%  & 55.8\% / 28.5\%  & \textbf{54.7\%} / \textbf{28.2\%}  \\
        \hspace{4mm}\#4 & 77.4\% / 54.5\%  & 34.5\% / 22.1\%  & \textbf{32.6\%} / \textbf{19.7\%}  \\
    \bottomrule
    \end{tabular}
    \vspace{2mm}
    \caption{\textbf{Quantitative results for zero-shot sentence-level ASL to English translation on FLEURS-ASL}, both on the test set and disaggregated across unique signers. Top: BLEU ($\uparrow$) (with sacreBLEU~\citep{post-2018-call} with \texttt{intl} tokenization), and subsets of the sentences with and without fingerspelling in parentheses. Middle: BLEURT ($\uparrow$), and again with/without fingerspelling subsets. Bottom: CER for the spans of the translation that are identified as corresponding to the FLEURS-ASL-FS fingerspelling annotations by an LLM / the percentage of reference fingerspelled terms where the LLM finds no match in the predicted translation.}
    \label{tab:trans-results-disagg}
\end{table*}

\section{How2Sign Results}
\label{app:how2sign}

In order to facilitate comparisons with prior works that report only on How2Sign~\citep{how2sign} (prior to the creation of FLEURS-ASL), in this section we report zero-shot and finetuned scores on How2Sign; see Table~\ref{tab:how2sign-results} for scores. Our best finetuned model surpasses the previous SOTA from~\citet{rust2024privacyaware}, reaching 18.1 vs. 15.5 BLEU and 50.8 vs. 49.6 BLEURT. Our use of more data and character-level modeling compensate for their use of self-supervised vision encodings rather than MediaPipe Holistic landmarks. We expect that combining both approaches would yield even better results, but that is outside the scope of this paper, which aims to analyze fingerspelling within sign language translation rather than strictly pursue state-of-the-art results.

See Table~\ref{tab:how2sign-qual} for qualitative examples and analysis. There are clear, interpretable examples where our ByT5-based model properly transcribes fingerspelled phrases that frustrate the other models. However, at this point these qualitative examples are almost saturated due to issues with How2Sign translation quality described in~\citet{reconsideringsentence}. Looking at the translated sign language clips rather than the English references:

\begin{itemize}
    \item[(1)] The~\citet{rust2024privacyaware} translation and our ByT5 translation are effectively perfect; there is no ``vital'' in the ASL source.
    \item[(2)] Our ByT5 translation translates ``name'' because the interpreter signs ``label'' while mouthing what looks more like ``name'' than ``tape'', and the followup is not very clear. It is somewhat miraculous that the models understand ``cable'' at all, because it is fingerspelled something like ``pcile''. There is no taping \textit{down} per se. The natural way to translate this sentence would be to depict taping down cables with a classifier predicate, but the video does not do this, probably because it is being interpreted live.
    \item[(3)] This example has the most room for improvement, but still not much. ``Feed'' in~\citet{rust2024privacyaware}'s translation is more accurate than ``install''. ``Distillation'' is fingerspelled as what looks like ``ditdiag'', then followed by the sign for ``nice'' or ``clean'' (presumably intended to mean ``clean/purify'' as a verb, as an explanation of ``distilling''). ``Essential'' is absent.
    \item[(4)] The subject of ``dance'' is unstated. ``Veil'' is represented with a sign that does not seem to be the standard form. When introduced there is some mouthing, but it is essentially a classifier without a referent in context.
    \item[(5)] Our ByT5 translation is essentially perfect here. ``Mound'' is spelled like ``cond''. ``Trash'' is seemingly not translated, and ``can'' is indeed spelled ``pan'' in the clip.
    \item[(6)] ``Important'' is right on the start boundary of the clip, so it could be interpreted either as a vestige of the previous sentence or part of the new one due to artifacts of sentence-level clipping. \citet{rust2024privacyaware}'s translation and our ByT5 translation are both essentially perfect.
\end{itemize}

We finetuned each of the 3 models with 32 TPUv3s with batch size 64 and Adafactor~\citep{shazeer2018adafactor} with base learning rate 0.001, for up to 10k steps. Each 1k steps takes about 20 minutes due to frequent checkpointing. We select the finetuned checkpoint based val set BLEU and manually stop training early once the model has clearly passed convergence.

\begin{table}[t]
    \centering
    \begin{tabular}{lcc}
    \toprule
    \bf model & \bf zero-shot & \bf finetuned \\
    \midrule
    \citet{rust2024privacyaware} & {\bf 7.1} (41.8) & 15.5 (49.6) \\
    \midrule
    T5 (YT) & 5.7 (38.1) & 16.8 (49.0) \\
    ByT5 (YT) & 6.8 {\bf (42.2)} & \bf 18.1 (50.8) \\
    ByT5 (YT+FS) & 6.3 (42.1) & 17.8 (50.6) \\
    \bottomrule
    \end{tabular}
    \caption{\textbf{Quantitative results for sentence-level ASL to English translation on How2Sign} measured in BLEU (sacreBLEU~\citep{post-2018-call} with default tokenization), and (in parentheses) BLEURT.}
    \label{tab:how2sign-results}
\end{table}

\begin{table*}[t]
    \centering
    \begin{tabular}{clp{11.5cm}}
    \toprule
    \multirow{4}{*}{(1)} & \bf reference & And that's a great vital point technique for women's self defense. \\
    &\citet{rust2024privacyaware} & This is a really great point for women’s self defense. \\
    &T5 (YT) & It's a really great point for women's self defenders. \\
    &ByT5 (YT) & It's a really great point for women's self defense. \\
    \midrule
    \multirow{4}{*}{(2)} & \bf reference & In this clip I'm going to show you how to tape your cables down. \\
    &\citet{rust2024privacyaware} & In this clip I’m going to show you how to clip the cable, the cable. \\
    &T5 (YT) & In this clip I'm going to show you how to cut a piece of sticky wire. \\
    &ByT5 (YT) & In this segment I'm going to show you how to draw a name of the cable. \\
    \midrule
    \multirow{8}{*}{(3)} & \bf reference & In this segment we're going to talk about how to load your still for distillation of lavender essential oil. \\
    &\citet{rust2024privacyaware} &  In this clip we’re going to talk about how to feed the trail for draining clean for laborer
oil. \\
    &T5 (YT) & In this clip we're going to talk about how to install a steel wool cleanser for a laser oil. \\
    &ByT5 (YT) & In this clip we're going to talk about how to install a still for ditching nice for a lavender oil. \\
    \midrule
    \multirow{8}{*}{(4)} & \bf reference & You are dancing, and now you are going to need the veil and you are going to just grab the veil as far as possible. \\
    &\citet{rust2024privacyaware} & So that she’s going to get her hips up as far as she can, and now she’s going to lift her
head up as far as possible. \\
    &T5 (YT) & Hi, I'm dancing now so we're going to need a brush to pull up and stretch our back as far as possible. \\
    &ByT5 (YT) & Now we're going to have her dance, and now we're going to have her braided her hair up as far as possible. \\
    \midrule
    \multirow{15}{*}{(5)} & \bf reference & But if you have to setup a new campfire, there's two ways to do it in a very low impact; one is with a mound fire, which we should in the campfire segment earlier and the other way to setup a low impact campfire is to have a fire pan, which is just a steel pan like the top of a trash can. \\
    &\citet{rust2024privacyaware} & But if you have to set up a new campfire, this is one way to do it in a low impact. One is a monk fire. One is a campfire. The other one is to set a campfire in a campfire. That’s just a post like the top of the post. \\
    &T5 (YT) & But if you have to set up a new campfire, there are two ways to do it in a low impact. One is a band fire which we should do in a campfire early, and the other one is to set up a campfire in a fire plan, which is just a patty like the top of the pan. \\
    &ByT5 (YT) & But if you have to set up a new campfire, there's two ways to do it in a low impact, one is a bond fire, which we should do in a campfire stack early and the other one is to set up a campfire in a fire pan, that's just a steel pan like the top of the pan. \\
    \midrule
    \multirow{4}{*}{(6)} & \bf reference & So, this is a very important part of the process. \\
    &\citet{rust2024privacyaware} & It’s an important part of the process. \\
    &T5 (YT) & That's the proper part of the process. \\
    &ByT5 (YT) & This is a part of the process. \\
    \bottomrule
    \end{tabular}
    \caption{\textbf{Qualitative results for finetuned sentence-level ASL to English translation on How2Sign}, originally selected by~\citet{tarres2023sign} and carried through~\citet{youtubeasl},~\citet{rust2024privacyaware}, and~\citet{youtubesl25}. Note the fingerspelled words that our ByT5 model recognizes correctly where all others fail, like ``lavender'' in (3) and ``pan'' in (5). The improvement on How2Sign is more muted than FLEURS-ASL because of issues with the translation quality; see Appendix~\ref{app:how2sign} for elaboration.}
    \label{tab:how2sign-qual}
\end{table*}

\section{Complete FLEURS-ASL-FS Annotations}
\label{app:full-fleurs-asl-fs}

See Table~\ref{tab:annotations0} for the complete set of FLEURS-ASL-FS annotations, also available at \datalink. We hope this will serve as a more convenient, glanceable reference than the raw dataset.

\clearpage
\begin{table*}[]
    \caption{\textbf{Complete set of FLEURS-ASL-FS annotations}. Each reference sentence is paired with a multiset of human-annotated fingerspelled phrases.}
    \centering
    \fontsize{8.4}{10}\selectfont
    \begin{tabular}{cp{10.5cm}p{3.5cm}}
        \toprule
        \textbf{signer} & \textbf{reference} & \textbf{fingerspelled terms} \\
        \midrule
0 & "We now have 4-month-old mice that are non-diabetic that used to be diabetic," he added. & \{{ \it  }\} \\ \midrule 
 0 & Dr. Ehud Ur, professor of medicine at Dalhousie University in Halifax, Nova Scotia and chair of the clinical and scientific division of the Canadian Diabetes Association cautioned that the research is still in its early days. & \{{ \it Dr. Ehud Ur, Dalhousie, Halifax, Nova Scotia, div }\} \\ \midrule 
 0 & Like some other experts, he is skeptical about whether diabetes can be cured, noting that these findings have no relevance to people who already have Type 1 diabetes. & \{{ \it  }\} \\ \midrule 
 0 & On Monday, Sara Danius, permanent secretary of the Nobel Committee for Literature at the Swedish Academy, publicly announced during a radio program on Sveriges Radio in Sweden the committee, unable to reach Bob Dylan directly about winning the 2016 Nobel Prize in Literature, had abandoned its efforts to reach him. & \{{ \it Sara Danius, Nobel, Lit, Sveriges, Bob Dylan, Nobel, Lit }\} \\ \midrule 
 0 & Danius said, "Right now we are doing nothing. I have called and sent emails to his closest collaborator and received very friendly replies. For now, that is certainly enough." & \{{ \it Danius }\} \\ \midrule 
 0 & Previously, Ring's CEO, Jamie Siminoff, remarked the company started when his doorbell wasn't audible from his shop in his garage. & \{{ \it Ring, CEO, Jamie Siminoff }\} \\ \midrule 
 0 & He built a WiFi door bell, he said. & \{{ \it WiFi }\} \\ \midrule 
 0 & Siminoff said sales boosted after his 2013 appearance in a Shark Tank episode where the show panel declined funding the startup. & \{{ \it Siminoff, TV, Shark Tank }\} \\ \midrule 
 0 & In late 2017, Siminoff appeared on shopping television channel QVC. & \{{ \it Siminoff, TV, QVC }\} \\ \midrule 
 0 & Ring also settled a lawsuit with competing security company, the ADT Corporation. & \{{ \it Ring, ADT Corporation }\} \\ \midrule 
 0 & While one experimental vaccine appears able to reduce Ebola mortality, up until now, no drugs have been clearly demonstrated suitable for treating existing infection. & \{{ \it Ebola, drugs }\} \\ \midrule 
 0 & One antibody cocktail, ZMapp, initially showed promise in the field, but formal studies indicated it had less benefit than sought in preventing death. & \{{ \it antibody, ZMapp }\} \\ \midrule 
 0 & In the PALM trial, ZMapp served as a control, meaning scientists used it as a baseline and compared the three other treatments to it. & \{{ \it PALM, control, baseline }\} \\ \midrule 
 0 & USA Gymnastics supports the United States Olympic Committee's letter and accepts the absolute need of the Olympic family to promote a safe environment for all of our athletes. & \{{ \it USA, US }\} \\ \midrule 
 0 & We agree with the USOC's statement that the interests of our athletes and clubs, and their sport, may be better served by moving forward with meaningful change within our organization, rather than decertification. & \{{ \it USOC, clubs, decertification }\} \\ \midrule 
 0 & USA Gymnastics supports an independent investigation that may shine light on how abuse of the proportion described so courageously by the survivors of Larry Nassar could have gone undetected for so long and embraces any necessary and appropriate changes. & \{{ \it USA, Larry Nassar }\} \\ \midrule 
 0 & USA Gymnastics and the USOC have the same goal — making the sport of gymnastics, and others, as safe as possible for athletes to follow their dreams in a safe, positive and empowered environment. & \{{ \it USA, USOC }\} \\ \midrule 
 0 & Throughout 1960s, Brzezinski worked for John F. Kennedy as his advisor and then the Lyndon B. Johnson administration. & \{{ \it Brzezinski, John F. Kennedy, Lyndon B. Johnson }\} \\ \midrule 
 0 & During the 1976 selections he advised Carter on foreign policy, then served as National Security Advisor (NSA) from 1977 to 1981, succeeding Henry Kissinger. & \{{ \it Carter, NSA, Henry Kissinger }\} \\ \midrule 
 0 & As NSA, he assisted Carter in diplomatically handling world affairs, such as the Camp David Accords, 1978; normalizing US–China relations thought the late 1970s; the Iranian Revolution, which led to the Iran hostage crisis, 1979; and the Soviet invasion in Afghanistan, 1979. & \{{ \it NSA, Carter, Camp David Accords, US, crisis, Soviet, Afghanistan }\} \\ \midrule 
 0 & The movie, featuring Ryan Gosling and Emma Stone, received nominations in all major categories. & \{{ \it Ryan Gosling, Emma Stone }\} \\ \midrule 
 0 & Gosling and Stone received nominations for Best Actor and Actress respectively. & \{{ \it Gosling, Stone }\} \\ \midrule 
 0 & The other nominations include Best Picture, Director, Cinematography, Costume Design, Film-editing, Original Score, Production Design, Sound Editing, Sound Mixing and Original Screenplay. & \{{ \it  }\} \\ \midrule 
 0 & Two songs from the movie, Audition (The Fools Who Dream) and City of Stars, received nominations for best original song. Lionsgate studio received 26 nominations — more than any other studio. & \{{ \it Audition (The Fools Who Dream),  City of Stars, Lionsgate studio, studio }\} \\

        \bottomrule
    \end{tabular}
    \label{tab:annotations0}
\end{table*}

\begin{table*}[]
    \centering
    \fontsize{8.4}{10}\selectfont
    \begin{tabular}{cp{10.5cm}p{3.5cm}}
        \toprule
        \textbf{signer} & \textbf{reference} & \textbf{fingerspelled terms} \\
        \midrule

0 & Late on Sunday, the United States President Donald Trump, in a statement delivered via the press secretary, announced US troops would be leaving Syria. & \{{ \it US, Donald Trump, press, US troops, Syria }\} \\ \midrule 
 0 & The announcement was made after Trump had a phone conversation with Turkish President Recep Tayyip Erdoğan. & \{{ \it Trump, Recep Tayyip Erdoğan }\} \\ \midrule 
 0 & Turkey would also take over guarding captured ISIS fighters which, the statement said, European nations have refused to repatriate. & \{{ \it ISIS }\} \\ \midrule 
 0 & This not only confirms that at least some dinosaurs had feathers, a theory already widespread, but provides details fossils generally cannot, such as color and three-dimensional arrangement. & \{{ \it fossils, 3D }\} \\ \midrule 
 0 & . Scientists say this animal's plumage was chestnut-brown on top with a pale or carotenoid-colored underside. & \{{ \it chestnut, carotenoid }\} \\ \midrule 
 0 & The find also grants insight into the evolution of feathers in birds. & \{{ \it evolution }\} \\ \midrule 
 0 & Because the dinosaur feathers do not have a well-developed shaft, called a rachis, but do have other features of feathers — barbs and barbules — the researchers inferred the rachis was likely a later evolutionary development that these other features. & \{{ \it rachis, barbs, barbules }\} \\ \midrule 
 0 & The feathers' structure suggests that they were not used in flight but rather for temperature regulation or display. & \{{ \it  }\} \\ \midrule 
 0 & The researchers suggested that, even though this is the tail of a young dinosaur, the sample shows adult plumage and not a chick's down. & \{{ \it  }\} \\ \midrule 
 0 & A car bomb detonated at police headquarters in Gaziantep, Turkey yesterday morning killed two police officers and injured more than twenty other people. & \{{ \it Gaziantep }\} \\ \midrule 
 0 & The governor's office said nineteen of the injured were police officers. & \{{ \it  }\} \\ \midrule 
 0 & Police said they suspect an alleged Daesh (ISIL) militant of responsibility for the attack. & \{{ \it Daesh, ISIL }\} \\ \midrule 
 0 & They found the Sun operated on the same basic principles as other stars: The activity of all stars in the system was found to be driven by their luminosity, their rotation, and nothing else. & \{{ \it  }\} \\ \midrule 
 0 & The luminosity and rotation are used together to determine a star's Rossby number, which is related to plasma flow. & \{{ \it Rossby, plasma }\} \\ \midrule 
 0 & The smaller the Rossby number, the less active the star with respect to magnetic reversals. & \{{ \it Rossby, magnetic reversal }\} \\ \midrule 
 0 & During his trip, Iwasaki ran into trouble on many occasions. & \{{ \it Iwasaki }\} \\ \midrule 
 0 & He was robbed by pirates, attacked in Tibet by a rabid dog, escaped marriage in Nepal and was arrested in India. & \{{ \it Tibet, rabid, Nepal }\} \\ \midrule 
 0 & The 802.11n standard operates on both the 2.4Ghz and 5.0Ghz frequencies. & \{{ \it 802.11n, 2.4Ghz, 5.0Ghz }\} \\ \midrule 
 0 & This will allow it to be backwards compatible with 802.11a, 802.11b and 802.11g, provided that the base station has dual radios. & \{{ \it 802.11a, 802.11b, 802.11g, base, radios }\} \\ \midrule 
 0 & The speeds of 802.11n are substantially faster than that of its predecessors with a maximum theoretical throughput of 600Mbit/s. & \{{ \it 802.11n, Mbit/s }\} \\ \midrule 
 0 & Duvall, who is married with two adult children, did not leave a big impression on Miller, to whom the story was related. & \{{ \it Duvall, Miller }\} \\ \midrule 
 0 & When asked for comment, Miller said, "Mike talks a lot during the hearing...I was getting ready so I wasn't really hearing what he was saying." & \{{ \it Miller, Mike }\} \\ \midrule 
 0 & "We will endeavour to cut carbon dioxide emissions per unit of GDP by a notable margin by 2020 from the 2005 level," Hu said. & \{{ \it Hu, carbon dioxide, CO2, unit, GDP }\} \\ \midrule 
 0 & He did not set a figure for the cuts, saying they will be made based on China's economic output. & \{{ \it  }\} \\ \midrule 
 0 & Hu encouraged developing countries "to avoid the old path of polluting first and cleaning up later." & \{{ \it Hu }\} \\ \midrule 
 0 & He added that "they should not, however, be asked to take on obligations that go beyond their development stage, responsibility and capabilities." & \{{ \it Hu }\} \\ \midrule 
 0 & The Iraq Study Group presented its report at 12.00 GMT today. & \{{ \it GMT }\} \\ \midrule 
 0 & It warns No one can guarantee that any course of action in Iraq at this point will stop sectarian warfare, growing violence, or a slide toward chaos. & \{{ \it chaos }\} \\ \midrule 
 0 & The Report opens with plea for open debate and the formation of a consensus in the United States about the policy towards the Middle East. & \{{ \it consensus, US }\} \\ 

        \bottomrule
    \end{tabular}
    \label{tab:annotations1}
\end{table*}

\begin{table*}[]
    \centering
    \fontsize{8.4}{10}\selectfont
    \begin{tabular}{cp{10.5cm}p{3.5cm}}
        \toprule
        \textbf{signer} & \textbf{reference} & \textbf{fingerspelled terms} \\
        \midrule
0 & The Report is highly critical of almost every aspect of the present policy of the Executive towards Iraq and it urges an immediate change of direction. & \{{ \it  }\} \\ \midrule 
 0 & First among its 78 recommendations is that a new diplomatic initiative should be taken before the end of this year to secure Iraq’s borders against hostile interventions and to re-establish diplomatic relations with its neighbors. & \{{ \it  }\} \\ \midrule 
 0 & Current senator and Argentine First Lady Cristina Fernandez de Kirchner announced her presidential candidacy yesterday evening in La Plata, a city 50 kilometers (31 miles) away from Buenos Aires. & \{{ \it Cristina Fernandez de Kirchner, La Plata, km,  miles, Buenos Aires }\} \\ \midrule 
 0 & Mrs. Kirchner announced her intention to run for president at the Argentine Theatre, the same location she used to start her 2005 campaign for the Senate as member of the Buenos Aires province delegation. & \{{ \it  }\} \\ \midrule 
 0 & The debate was sparked by controversy over spending on relief and reconstruction in the wake Hurricane Katrina; which some fiscal conservatives have humorously labeled "Bush's New Orleans Deal." & \{{ \it Katrina, Bush, deal }\} \\ \midrule 
 0 & Liberal criticism of the reconstruction effort has focused on the awarding of reconstruction contracts to perceived Washington insiders. & \{{ \it  }\} \\ \midrule 
 0 & Over four million people went to Rome to attend the funeral. & \{{ \it  }\} \\ \midrule 
 0 & The number of people present was so large that it was not possible for everybody to gain access to the funeral in St. Peter's Square. & \{{ \it St. Peter's Square }\} \\ \midrule 
 0 & Several large television screens were installed in various places in Rome to let the people watch the ceremony. & \{{ \it TV }\} \\ \midrule 
 0 & In many other cities of Italy and in the rest of the world, particularly in Poland, similar setups were made, which were viewed by a great number of people. & \{{ \it Poland, TV }\} \\ \midrule 
 0 & Historians have criticized past FBI policies for focusing resources on cases which are easy to solve, especially stolen car cases, with the intent of boosting the agency's success rate. & \{{ \it FBI, cases, cases, FBI }\} \\ \midrule 
 0 & Congress began funding the obscenity initiative in fiscal 2005 and specified that the FBI must devote 10 agents to adult pornography. & \{{ \it FY, FBI, FBI, porn }\} \\ \midrule 
 0 & Robin Uthappa made the innings highest score, 70 runs in just 41 balls by hitting 11 fours and 2 sixes. & \{{ \it Robin Uthappa }\} \\ \midrule 
 0 & Middle order batsmen, Sachin Tendulkar and Rahul Dravid, performed well and made a hundred-run partnership. & \{{ \it Sachin Tendulkar, Rahul Dravid }\} \\ \midrule 
 0 & But, after losing the captain's wicket India only made 36 runs loosing 7 wickets to end the innings. & \{{ \it  }\} \\ \midrule 
 0 & U.S. President George W. Bush arrived in Singapore the morning of November 16, beginning a week-long tour of Asia. & \{{ \it US, George W. Bush, Singapore, Nov }\} \\ \midrule 
 0 & He was greeted by Singapore's Deputy Prime Minister Wong Kan Seng and discussed trade and terrorism issues with the Singapore Prime Minister Lee Hsien Loong. & \{{ \it PM, Wong Kan Seng, terrorism, PM, Lee Hsien Loong }\} \\ \midrule 
 0 & After a week of losses in the midterm election, Bush told an audience about the expansion of trade in Asia. & \{{ \it midterm, Bush }\} \\ \midrule 
 0 & Prime Minister Stephen Harper has agreed to send the government's 'Clean Air Act' to an all-party committee for review, before its second reading, after Tuesday's 25 minute meeting with NDP leader Jack Layton at the PMO. & \{{ \it PM, Stephen Harper, Clean Air Act, NDP, Jack Layton, PMO }\} \\ \midrule 
 0 & Layton had asked for changes to the conservatives' environmental bill during the meeting with the PM, asking for a "thorough and complete rewriting" of the Conservative party's environmental bill. & \{{ \it Layton, PM }\} \\ \midrule 
 0 & Ever since the Federal Government stepped in to take over funding of the Mersey hospital in Devonport, Tasmania, the state government and some federal MPs have criticised this act as a stunt in the prelude to the federal election to be called by November. & \{{ \it Mersey, Devonport, Tasmania, MP, stunt, Nov }\} \\ \midrule 
 0 & But Prime Minister John Howard has said the act was only to safeguard the facilities of the hospital from being downgraded by the Tasmanian government, in giving an extra AUD\$45 million. & \{{ \it PM, John Howard, Tas }\} \\ \midrule 
 0 & According to the latest bulletin, sea level readings indicated a tsunami was generated. There was some definite tsunami activity recorded near Pago Pago and Niue. & \{{ \it sea, tsunami, Pago Pago, Niue }\} \\ \midrule
 0 & No major damage or injuries have been reported in Tonga, but power was temporarily lost, which reportedly prevented Tongan authorities from receiving the tsunami warning issued by the PTWC. & \{{ \it Tonga, PTWC, Tonga }\} \\
        \bottomrule
    \end{tabular}
    \label{tab:annotations2}
\end{table*}

\begin{table*}[]
    \centering
    \fontsize{8.4}{10}\selectfont
    \begin{tabular}{cp{10.5cm}p{3.5cm}}
        \toprule
        \textbf{signer} & \textbf{reference} & \textbf{fingerspelled terms} \\
        \midrule
 0 & Fourteen schools in Hawaii located on or near coastlines were closed all of Wednesday despite the warnings being lifted. & \{{ \it  }\} \\ \midrule 
 0 & U.S. President George W. Bush welcomed the announcement. & \{{ \it US, George W. Bush }\} \\ \midrule 
 0 & Bush spokesman Gordon Johndroe called North Korea's pledge "a major step towards the goal of achieving the verifiable denuclearization of the Korean peninsula." & \{{ \it Bush, Gordon Johndroe, nuclear weapons }\} \\ \midrule 
 0 & The tenth named storm of the Atlantic Hurricane season, Subtropical Storm Jerry, formed in the Atlantic Ocean today. & \{{ \it Atlantic, Jerry, Subtropical, Atlantic }\} \\ \midrule 
 0 & The National Hurricane Center (NHC) says that at this point Jerry poses no threat to land. & \{{ \it NHC, Jerry }\} \\ \midrule 
 0 & The U.S. Corps of Engineers estimated that 6 inches of rainfall could breach the previously damaged levees. & \{{ \it US Corps, inches, levee }\} \\ \midrule 
 0 & The Ninth Ward, which saw flooding as high as 20 feet during Hurricane Katrina, is currently in waist-high water as the nearby levee was overtopped. & \{{ \it Ninth Ward, Katrina, levee }\} \\ \midrule 
 0 & Water is spilling over the levee in a section 100 feet wide. & \{{ \it feet }\} \\ \midrule 
 0 & Commons Administrator Adam Cuerden expressed his frustration over the deletions when he spoke to Wikinews last month. & \{{ \it Commons, Adam Cuerden, Wikinews }\} \\ \midrule 
 0 & "He [Wales] basically lied to us from the start. First, by acting as if this was for legal reasons. Second, by pretending he was listening to us, right up to his art deletion." & \{{ \it Wales }\} \\ \midrule 
 0 & The community irritation led to current efforts to draft a policy regarding sexual content for the site which hosts millions of openly-licensed media. & \{{ \it sexual, media, gifs }\} \\ \midrule 
 0 & The work done was mostly theoretical, but the program was written to simulate observations made of the Sagittarius galaxy. & \{{ \it Sagittarius, galaxy }\} \\ \midrule 
 0 & The effect the team was looking for would be caused by tidal forces between the galaxy's dark matter and the Milky Way's dark matter. & \{{ \it effect, force, dark matter, Milky Way, dark matter }\} \\ \midrule 
 0 & Just like the moon exerts a pull on the earth, causing tides, so does the Milky Way exert a force on the Sagittarius galaxy. & \{{ \it Milky Way, Sagittarius, galaxy }\} \\ \midrule 
 0 & The scientists were able to conclude that the dark matter affect other dark matter in the same way regular matter does. & \{{ \it dark matter, dark matter, matter }\} \\ \midrule 
 0 & This theory says that most dark matter around a galaxy is located around a galaxy in a kind of halo, and is made of lots of small particles. & \{{ \it dark matter, galaxy, particles }\} \\ \midrule 
 0 & Television reports show white smoke coming from the plant. & \{{ \it TV, plant }\} \\ \midrule 
 0 & Local authorities are warning residents in the vicinity of the plant to stay indoors, turn off air-conditioners and not to drink tap water. & \{{ \it AC }\} \\ \midrule 
 0 & According to Japan's nuclear agency, radioactive caesium and iodine has been identified at the plant. & \{{ \it nuclear, Cs, caesium, I, iodine, radioactive }\} \\ \midrule 
 0 & Authorities speculate that this indicates that containers holding uranium fuel at the site may have ruptured and are leaking. & \{{ \it uranium }\} \\ \midrule 
 0 & Dr. Tony Moll discovered the Extremely Drug Resistant Tuberculosis (XDR-TB) in the South African region KwaZulu-Natal. & \{{ \it Dr., Tony Moll, Tuberculosis, XDR-TB, KwaZulu-Natal }\} \\ \midrule 
 0 & In an interview, he said the new variant was "very highly troubling and alarming because of the very high fatality rate." & \{{ \it  }\} \\ \midrule 
 0 & Some patients might have contracted the bug in the hospital, Dr. Moll thinks, and at least two were hospital health workers. & \{{ \it  }\} \\ \midrule 
 0 & In one year's time, an infected person may infect 10 to 15 close contacts. & \{{ \it  }\} \\ \midrule 
 0 & However, the percentage of XDR-TB in the entire group of people with tuberculosis still seems to be low; 6,000 of the total 330,000 people infected at any particular moment in South Africa. & \{{ \it XDR-TB, TB }\} \\ \midrule 
 0 & The satellites, both of which weighed in excess of 1,000 pounds, and traveling at approximately 17,500 miles per hour, collided 491 miles above the Earth. & \{{ \it satellites, pounds, miles, miles }\} \\ \midrule 
 0 & Scientists say the explosion caused by the collision was massive. & \{{ \it  }\} \\ \midrule 
 0 & They are still trying to determine just how large the crash was and how the Earth will be affected. & \{{ \it  }\} \\ \midrule 
 0 & The United States Strategic Command of the U.S. Department of Defense office is tracking the debris. & \{{ \it US, US }\} \\ \midrule 
 0 & The result of plotting analysis will be posted to a public website. & \{{ \it plotting }\} \\

        \bottomrule
    \end{tabular}
    \label{tab:annotations3}
\end{table*}

\begin{table*}[]
    \centering
    \fontsize{8.4}{10}\selectfont
    \begin{tabular}{cp{10.5cm}p{3.5cm}}
        \toprule
        \textbf{signer} & \textbf{reference} & \textbf{fingerspelled terms} \\
        \midrule
0 & A doctor who worked at Children's Hospital of Pittsburgh, Pennsylvania will be charged with aggravated murder after her mother was found dead in the trunk of her car Wednesday, authorities in Ohio say. & \{{ \it Pittsburgh, PA, aggravated murder, Ohio }\} \\ \midrule 
 0 & Dr. Malar Balasubramanian, 29, was found in Blue Ash, Ohio, a suburb approximately 15 miles north of Cincinnati lying on the ground beside the road in a T-shirt and underwear in an apparently heavily medicated state. & \{{ \it Dr. Malar Balasubramanian, Blue Ash, Ohio, suburb, miles, Cincinnati }\} \\ \midrule 
 0 & She directed officers to her black Oldsmobile Intrigue which was 500 feet away. & \{{ \it Oldsmobile }\} \\ \midrule 
 0 & There, they found the body of Saroja Balasubramanian, 53, covered with blood-stained blankets. & \{{ \it Saroja Balasubramanian }\} \\ \midrule 
 0 & Police said that the body appeared to have been there for about a day. & \{{ \it  }\} \\ \midrule 
 0 & The first cases of the disease this season were reported in late July. & \{{ \it July }\} \\ \midrule 
 0 & The disease is carried by pigs, which then migrates to humans through mosquitos. & \{{ \it  }\} \\ \midrule 
 0 & The outbreak has prompted the Indian government to undertake such measures as deployment of pig catchers in seriously affected areas, distributing thousands of mosquito curtains and spraying pesticides. & \{{ \it  }\} \\ \midrule 
 0 & Several million vials of encephalitis vaccine have also been promised by the government, which will help prepare health agencies for next year. & \{{ \it encephalitis }\} \\ \midrule 
 0 & Plans for vaccines to be delivered to the historically most affected areas this year were delayed due to lack of funds and low prioritisation relative to other diseases. & \{{ \it  }\} \\ \midrule 
 0 & In 1956 Słania moved to Sweden, where three years later he began work for the Swedish Post Office and became their chief engraver. & \{{ \it Słania, PO }\} \\ \midrule 
 0 & He produced over 1,000 stamps for Sweden and 28 other countries. & \{{ \it  }\} \\ \midrule 
 0 & His work is of such recognized quality and detail that he is one of the very few "household names" among philatelists. Some specialize in collecting his work alone. & \{{ \it  }\} \\ \midrule 
 0 & His 1,000th stamp was the magnificent "Great Deeds by Swedish Kings" by David Klöcker Ehrenstrahl in 2000, which is listed in the Guinness Book of World Records. & \{{ \it Great Deeds by Swedish Kings, David Klöcker Ehrenstrahl, Guinness }\} \\ \midrule 
 0 & He was also engaged in engraving banknotes for many countries, recent examples of his work including the Prime Ministerial portraits on the front of the new Canadian \$5 and \$100 bills. & \{{ \it PM }\} \\ \midrule 
 0 & After the accident occurred, Gibson was transported to a hospital but died shortly afterwards. & \{{ \it Gibson }\} \\ \midrule 
 0 & The truck driver, who is aged 64, was not injured in the crash. & \{{ \it  }\} \\ \midrule 
 0 & The vehicle itself was taken away from the scene of the accident at approximately 1200 GMT on the same day. & \{{ \it GMT }\} \\ \midrule 
 0 & A person working in a garage near where the accident occurred said: "There were children waiting to cross the road and they were all screaming and crying." & \{{ \it  }\} \\ \midrule 
 0 & They all ran back from where the accident had happened. & \{{ \it  }\} \\ \midrule 
 0 & Other subjects on the agenda in Bali include saving the world's remaining forests, and sharing technologies to help developing nations grow in less-polluting ways. & \{{ \it Bali }\} \\ \midrule 
 0 & The U.N. also hopes to finalize a fund to help countries affected by global warming to cope with the impacts. & \{{ \it UN }\} \\ \midrule 
 0 & The money could go toward flood-proof houses, better water management, and crop diversification. & \{{ \it  }\} \\ \midrule 
 0 & Fluke wrote that the efforts by some to drown out women from speaking out about women’s health were unsuccessful. & \{{ \it Fluke }\} \\ \midrule 
 0 & She came to this conclusion due to the multitude of positive comments and encouragement sent to her by both female and male individuals urging that contraception medication be considered a medical necessity. & \{{ \it Fluke }\} \\ \midrule 
 0 & When the fighting ceased after the wounded were transported to the hospital, about 40 of the other remaining inmates stayed in the yard and refused to return to their cells. & \{{ \it yard }\} \\ \midrule 
 0 & Negotiators tried to rectify the situation, but the prisoners' demands are not clear. & \{{ \it  }\} \\ \midrule 
 0 & Between 10:00-11:00 pm MDT, a fire was started by the inmates in the yard. & \{{ \it MDT, yard }\} \\ \midrule 
 0 & Soon, officers equipped with riot gear entered the yard and cornered the inmates with tear gas. & \{{ \it tear gas }\} \\

        \bottomrule
    \end{tabular}
    \label{tab:annotations4}
\end{table*}

\begin{table*}[]
    \centering
    \fontsize{8.4}{10}\selectfont
    \begin{tabular}{cp{10.5cm}p{3.5cm}}
        \toprule
        \textbf{signer} & \textbf{reference} & \textbf{fingerspelled terms} \\
        \midrule
0 & Fire rescue crews eventually doused the fire by 11:35 pm. & \{{ \it  }\} \\ \midrule 
 0 & After the dam was built in 1963, the seasonal floods that would spread sediment throughout the river were halted. & \{{ \it dam, sediment, mineral }\} \\ \midrule 
 0 & This sediment was necessary for creating sandbars and beaches, which served as wildlife habitats. & \{{ \it sediment, sandbar }\} \\ \midrule 
 0 & As a result, two fish species have become extinct, and two others have become endangered, including the humpback chub. & \{{ \it humpback chub }\} \\ \midrule 
 0 & Although the water level will only rise a few feet after the flood, officials are hoping it will be enough to restore eroded sandbars downstream. & \{{ \it feet, sandbar }\} \\ \midrule 
 0 & No tsunami warning has been issued, and according to the Jakarta geophysics agency, no tsunami warning will be issued because the quake did not meet the magnitude 6.5 requirement. & \{{ \it tsunami, Jakarta, magnitude }\} \\ \midrule 
 0 & Despite there being no tsunami threat, residents started to panic and began to leave their businesses and homes. & \{{ \it  }\} \\ \midrule 
 0 & Although Winfrey was tearful in her farewell, she made it clear to her fans she will be back. & \{{ \it Winfrey, fans }\} \\ \midrule 
 0 & "This is not going to be goodbye. This is the closing of one chapter and the opening of a new one." & \{{ \it  }\} \\ \midrule 
 0 & Final results from Namibian presidential and parliamentary elections have indicated that the incumbent president, Hifikepunye Pohamba, has been reelected by a large margin. & \{{ \it Namibia, parliamentary, Hifikepunye Pohamba }\} \\ \midrule 
 0 & The ruling party, South West Africa People's Organisation (SWAPO), also retained a majority in the parliamentary elections. & \{{ \it SWAPO, parliamentary }\} \\ \midrule 
 0 & Coalition and Afghan troops moved into the area to secure the site and other coalition aircraft have been sent to assist. & \{{ \it coalition, Afghan }\} \\ \midrule 
 0 & The crash occurred high up in mountainous terrain, and is believed to have been the result of hostile fire. & \{{ \it crash }\} \\ \midrule 
 0 & Efforts to search for the crash site are being met by bad weather and harsh terrain. & \{{ \it  }\} \\ \midrule 
 0 & The medical charity Mangola, Medecines Sans Frontieres and the World Health Organisation say it is the worst outbreak recorded in the country. & \{{ \it Mangola, Medecines Sans Frontieres }\} \\ \midrule 
 0 & Spokesman for Medecines Sans Frontiere Richard Veerman said: "Angola is heading for its worst ever outbreak and the situation remains very bad in Angola," he said. & \{{ \it Medecines Sans Frontiere, Richard Veerman, Angola }\} \\ \midrule 
 0 & The games kicked off at 10:00am with great weather and apart from mid morning drizzle which quickly cleared up, it was a perfect day for 7's rugby. & \{{ \it  }\} \\ \midrule 
 0 & Tournament top seeds South Africa started on the right note when they had a comfortable 26 - 00 win against 5th seeded Zambia. & \{{ \it seeded, Zambia }\} \\ \midrule 
 0 & Looking decidedly rusty in the game against their southern sisters, South Africa however steadily improved as the tournament progressed. & \{{ \it  }\} \\ \midrule 
 0 & Their disciplined defence, ball handling skills and excellent team work made them stand out and it was clear that this was the team to beat. & \{{ \it  }\} \\ \midrule 
 0 & Officials for the city of Amsterdam and the Anne Frank Museum state that the tree is infected with a fungus and poses a public health hazard as they argue that it was in imminent danger of falling over. & \{{ \it Amsterdam, Anne Frank, fungus }\} \\ \midrule 
 0 & It had been scheduled to be cut down on Tuesday, but was saved after an emergency court ruling. & \{{ \it  }\} \\ \midrule 
 0 & All of the cave entrances, which were named "The Seven Sisters", are at least 100 to 250 meters (328 to 820 feet) in diameter. & \{{ \it cave, ft }\} \\ \midrule 
 0 & Infrared images show that the temperature variations from night and day show that they are likely caves. & \{{ \it IR, cave }\} \\ \midrule 
 0 & "They are cooler than the surrounding surface in the day and warmer at night. & \{{ \it  }\} \\ \midrule 
 0 & Their thermal behavior is not as steady as large caves on Earth that often maintain a fairly constant temperature, but it is consistent with these being deep holes in the ground," said Glen Cushing of the United States Geological Survey (USGS) Astrogeology Team and of Northern Arizona University located in Flagstaff, Arizona. & \{{ \it Glen Cushing, US, USGS, Astrogeology }\} \\ \midrule 
 0 & In France, voting has traditionally been a low-tech experience: voters isolate themselves in a booth, put a pre-printed sheet of paper indicating their candidate of choice into an envelope. & \{{ \it booth }\} \\ 
        \bottomrule
    \end{tabular}
    \label{tab:annotations5}
\end{table*}

\begin{table*}[]
    \centering
    \fontsize{8.4}{10}\selectfont
    \begin{tabular}{cp{10.5cm}p{3.5cm}}
        \toprule
        \textbf{signer} & \textbf{reference} & \textbf{fingerspelled terms} \\
        \midrule
0 & After officials verify the voter's identity, the voter drops the envelope into the ballot box and signs the voting roll. & \{{ \it ID, ballot }\} \\ \midrule 
 0 & French electoral law rather strictly codifies the proceedings. & \{{ \it  }\} \\ \midrule 
 0 & Since 1988, ballot boxes must be transparent so that voters and observers can witness that no envelopes are present at the start of the vote and that no envelopes are added except those of the duly counted and authorized voters. & \{{ \it  }\} \\ \midrule 
 0 & Candidates can send representatives to witness every part of the process. In the evening, votes are counted by volunteers under heavy supervision, following specific procedures. & \{{ \it  }\} \\ \midrule 
 0 & ASUS Eee PC, earlier launched world-wide for cost-saving and functionality factors, became a hot topic in 2007 Taipei IT Month. & \{{ \it ASUS Eee PC, Taipei, IT }\} \\ \midrule 
 0 & But the consumer market on laptop computer will be radically varied and changed after ASUS was awarded in the 2007 Taiwan Sustainable Award by Executive Yuan of the Republic of China. & \{{ \it Yuan, ASUS, radically }\} \\ \midrule 
 0 & The station's web site describes the show as "old school radio theater with a new and outrageous geeky spin!" & \{{ \it  }\} \\ \midrule 
 0 & In its early days, the show was featured solely at the long-running internet radio site TogiNet Radio, a site focused on talk radio. & \{{ \it TogiNet Radio }\} \\ \midrule 
 0 & In late 2015, TogiNet established AstroNet Radio as a subsidiary station. & \{{ \it TogiNet, AstroNet }\} \\ \midrule 
 0 & The show originally featured amateur voice actors, local to East Texas. & \{{ \it  }\} \\ \midrule 
 0 & Widespread looting reportedly continued overnight, as law enforcement officers were not present on Bishkek's streets. & \{{ \it Bishkek }\} \\ \midrule 
 0 & Bishkek was described as sinking into a state of "anarchy" by one observer, as gangs of people roamed the streets and plundered stores of consumer goods. & \{{ \it Bishkek, anarchy }\} \\ \midrule 
 0 & Several Bishkek residents blamed protesters from the south for the lawlessness. & \{{ \it Bishkek }\} \\ \midrule 
 0 & South Africa have defeated the All Blacks (New Zealand) in a rugby union Tri Nations match at the Royal Bafokeng Stadium in Rustenburg, South Africa. & \{{ \it All Blacks, Tri Nations, Royal Bafokeng, Rustenburg }\} \\ \midrule 
 0 & The final score was a one-point victory, 21 to 20, ending the All Blacks' 15 game winning streak. & \{{ \it All Blacks }\} \\ \midrule 
 0 & For the Springboks, it ended a five-match losing streak. & \{{ \it Springboks }\} \\ \midrule 
 0 & It was the final match for the All Blacks, who had already won the trophy two weeks ago. & \{{ \it All Blacks }\} \\ \midrule 
 0 & The final match of the series will take place at Ellis Park in Johannesburg next week, when the Springboks play Australia. & \{{ \it Ellis Park, Johannesburg, Springboks }\} \\ \midrule 
 0 & A moderate earthquake shook western Montana at 10:08 p.m. on Monday. & \{{ \it  }\} \\ \midrule 
 0 & No immediate reports of damage have been received by the United States Geological Survey (USGS) and its National Earthquake Information Center. & \{{ \it USGS, US, USGS }\} \\ \midrule 
 0 & The earthquake was centered about 20 km (15 miles) north-northeast of Dillon, and about 65 km (40 miles) south of Butte. & \{{ \it Dillon, km, miles, Butte, km, miles }\} \\ \midrule 
 0 & The strain of bird flu lethal to humans, H5N1, has been confirmed to have infected a dead wild duck, found on Monday, in marshland near Lyon in the east of France. & \{{ \it H5N1, flu, Lyon }\} \\ \midrule 
 0 & France is the seventh country in the European Union to suffer this virus; following Austria, Germany, Slovenia, Bulgaria, Greece and Italy. & \{{ \it  }\} \\ \midrule 
 0 & Suspected cases of H5N1 in Croatia and Denmark remain unconfirmed. & \{{ \it H5N1 }\} \\ \midrule 
 0 & Chambers had sued God for "widespread death, destruction and terrorization of millions upon millions of the Earth's inhabitants." & \{{ \it Chambers }\} \\ \midrule 
 0 & Chambers, an agnostic, argues that his lawsuit is "frivolous" and "anybody can sue anybody." & \{{ \it Chambers, frivolous }\} \\ \midrule 
 0 & The story presented in the French opera, by Camille Saint-Saens, is of an artist "whose life is dictated by a love for drugs and Japan." & \{{ \it opera, Camille Saint-Saens }\} \\ \midrule 
 0 & As a result, the performers smoke cannabis joints on stage, and the theatre itself is encouraging the audience to join in. & \{{ \it cannabis }\} \\ \midrule 
 0 & Former House Speaker Newt Gingrich, Texas governor Rick Perry, and Congresswoman Michele Bachmann finished in fourth, fifth, and sixth place, respectively. & \{{ \it Newt Gingrich, Rick Perry, Michele Bachmann }\} \\
        \bottomrule
    \end{tabular}
    \label{tab:annotations6}
\end{table*}

\begin{table*}[]
    \centering
    \fontsize{8.4}{10}\selectfont
    \begin{tabular}{cp{10.5cm}p{3.5cm}}
        \toprule
        \textbf{signer} & \textbf{reference} & \textbf{fingerspelled terms} \\
        \midrule
0 & After the results came in, Gingrich lauded Santorum, but had tough words for Romney, on whose behalf negative campaign advertisements were aired in Iowa against Gingrich. & \{{ \it Gingrich, Santorum, Gingrich, Romney, Iowa, Gingrich }\} \\ \midrule 
 0 & Perry stated that he would "return to Texas to assess the results of tonight's caucus, determine whether there is a path forward for myself in this race", but later said that he would remain in the race and compete in the January 21 South Carolina primary. & \{{ \it Perry, January, SC, primary }\} \\ \midrule 
 0 & Bachmann, who won the Ames Straw Poll in August, decided to end her campaign. & \{{ \it Bachmann, Ames Straw Poll, Aug }\} \\ \midrule 
 0 & The photographer was transported to Ronald Reagan UCLA Medical Center, where he subsequently died. & \{{ \it Ronald Reagan UCLA }\} \\ \midrule 
 0 & He was reportedly aged in his 20s. In a statement, Bieber said "[w]hile I was not present nor directly involved with this tragic accident, my thoughts and prayers are with the family of the victim." & \{{ \it Bieber }\} \\ \midrule 
 0 & Entertainment news website TMZ understands the photographer stopped his vehicle on the other side of Sepulveda Boulevard and attempted to take pictures of the police stop before crossing the road and continuing, prompting the California Highway Patrol police officer conducting the traffic stop to order him back across, twice. & \{{ \it TMZ, Sepulveda Blvd }\} \\ \midrule 
 0 & According to police, the driver of the vehicle that hit the photographer is unlikely to face criminal charges. & \{{ \it criminal }\} \\ \midrule 
 0 & With only eighteen medals available a day, a number of countries have failed to make the medal podium. & \{{ \it  }\} \\ \midrule 
 0 & They include the Netherlands, with Anna Jochemsen finishing ninth in the women's standing class in the Super-G yesterday, and Finland with Katja Saarinen finishing tenth in the same event. & \{{ \it Anna Jochemsen, Super-G, Katja Saarinen }\} \\ \midrule 
 0 & Australia's Mitchell Gourley finished eleventh in the men's standing Super-G. Czech competitor Oldrich Jelinek finished sixteenth in the men's sitting Super-G. & \{{ \it Mitchell Gourley, Super-G, Oldrich Jelinek, Super-G }\} \\ \midrule 
 0 & Arly Velasquez of Mexico finished fifteenth in the men's sitting Super-G. New Zealand's Adam Hall finished ninth in the men's standing Super-G. & \{{ \it Arly Velasquez, Super-G, Adam Hall, Super-G }\} \\ \midrule 
 0 & Poland's men's visually impaired skier Maciej Krezel and guide Anna Ogarzynska finished thirteenth in the Super-G. South Korea's Jong Seork Park finished twenty-fourth in the men's sitting Super-G. & \{{ \it Maciej Krezel, Anna Ogarzynska, Super-G, Jong Seork Park, Super-G }\} \\ \midrule 
 0 & UN peacekeepers, whom arrived in Haiti after the 2010 earthquake, are being blamed for the spread of the disease which started near the troop's encampment. & \{{ \it UN }\} \\ \midrule 
 0 & According to the lawsuit, waste from the UN camp was not properly sanitized, causing bacteria to enter the tributary of the Artibonite River, one of Haiti's largest. & \{{ \it UN, Artibonite }\} \\ \midrule 
 0 & Prior to the arrival of troops, Haiti had not encountered problems related to the disease since the 1800s. & \{{ \it  }\} \\ \midrule 
 0 & The Haitian Institute for Justice and Democracy has referenced independent studies that suggest the Nepalese UN peacekeeping battalion unknowingly brought the disease to Haiti. & \{{ \it UN }\} \\ \midrule 
 0 & Danielle Lantagne, a UN expert on the disease, stated the outbreak was likely caused by the peacekeepers. & \{{ \it UN, Danielle Lantagne, UN }\} \\ \midrule 
 0 & Hamilton confirmed Howard University Hospital admitted the patient in stable condition. & \{{ \it Hamilton, Howard }\} \\ \midrule 
 0 & The patient had been to Nigeria, where some cases of the Ebola virus have occurred. & \{{ \it Nigeria, Ebola }\} \\ \midrule 
 0 & The hospital has followed protocol for infection control, including separating the patient from others to prevent possible infection of others. & \{{ \it  }\} \\ \midrule 
 0 & Before The Simpsons Simon had worked on several shows in various positions. & \{{ \it The Simpsons, Simon }\} \\ \midrule 
 0 & During the 1980s he worked on shows such as Taxi, Cheers, and The Tracy Ullman Show. & \{{ \it Taxi, Cheers, The Tracy Ullman }\} \\ \midrule 
 0 & In 1989 he helped create The Simpsons with Brooks and Groening, and was responsible for hiring the show's first writing team. & \{{ \it The Simpsons, Brooks, Groening }\} \\ \midrule 
 0 & Despite leaving the show in 1993 he kept the title of executive producer, and continued to receive tens of millions of dollars every season in royalties. & \{{ \it  }\} \\ \midrule 
 0 & Earlier the Chinese news agency Xinhua reported a plane to be hijacked. & \{{ \it Xinhua }\} \\ \midrule 
 0 & Later reports then stated the plane received a bomb threat and was diverted back to Afghanistan, landing in Kandahar. & \{{ \it Afghanistan, Kandahar }\} \\
        \bottomrule
    \end{tabular}
    \label{tab:annotations7}
\end{table*}

\begin{table*}[]
    \centering
    \fontsize{8.4}{10}\selectfont
    \begin{tabular}{cp{10.5cm}p{3.5cm}}
        \toprule
        \textbf{signer} & \textbf{reference} & \textbf{fingerspelled terms} \\
        \midrule
 0 & The early reports say the plane was diverted back to Afghanistan after being denied an emergency landing in Ürümqi. & \{{ \it Urumqi }\} \\ \midrule 
 0 & Air accidents are common in Iran, which has an aging fleet that is poorly maintained both for civil and military operations. & \{{ \it  }\} \\ \midrule 
 0 & International sanctions have meant that new aircraft cannot be purchased. & \{{ \it  }\} \\ \midrule 
 0 & Earlier this week, a police helicopter crash killed three people and wounded three more. & \{{ \it  }\} \\ \midrule 
 0 & Last month Iran saw its worst air disaster in years when an airliner heading to Armenia crashed, killing the 168 on board. & \{{ \it Armenia }\} \\ \midrule 
 0 & The same month saw another airliner overrun a runway at Mashhad and strike a wall, killing seventeen. & \{{ \it Mashhad }\} \\ \midrule 
 0 & Aerosmith have cancelled their remaining concerts on their tour. & \{{ \it Aerosmith }\} \\ \midrule 
 0 & The rock band was due to tour the United States and Canada until September 16. & \{{ \it Sept }\} \\ \midrule 
 0 & They have cancelled the tour after lead singer Steven Tyler was injured after he fell off stage while performing on August 5. & \{{ \it Steven Tyler, Aug }\} \\ \midrule 
 0 & Murray lost the first set in a tie break after both men held each and every serve in the set. & \{{ \it Murray }\} \\ \midrule 
 0 & Del Potro had the early advantage in the second set, but this too required a tie break after reaching 6-6. & \{{ \it Del Potro }\} \\ \midrule 
 0 & Potro received treatment to his shoulder at this point but managed to return to the game. & \{{ \it Potro }\} \\ \midrule 
 0 & The program started at 8:30 p.m. local time (15.00 UTC). & \{{ \it UTC }\} \\ \midrule 
 0 & Famous singers across the country presented bhajans, or devotional songs, to Shri Shyam's feet. & \{{ \it bhajans, Shri Shyam }\} \\ \midrule 
 0 & Singer Sanju Sharma started the evening, followed by Jai Shankar Choudhary. esented the chhappan bhog bhajan as well. Singer, Raju Khandelwal was accompanying him. & \{{ \it Sanju Sharma, Jai Shankar Choudhary, chhappan bhog bhajan, Raju Khandelwal }\} \\ \midrule 
 0 & Then, Lakkha Singh took the lead in singing the bhajans. & \{{ \it Lakkha Singh, bhajans }\} \\ \midrule 
 0 & 108 plates of Chhappan Bhog (in Hinduism, 56 different edible items, like, sweets, fruits, nuts, dishes etc. which are offered to deity) were served to Baba Shyam. & \{{ \it chhappan bhog, Hinduism, Baba Shyam }\} \\ \midrule 
 0 & Lakkha Singh presented the chhappan bhog bhajan as well. Singer, Raju Khandelwal was accompanying him. & \{{ \it Lakkha Singh, chhappan bhog bhajan, Raju Khandelwal }\} \\ \midrule 
 0 & At Thursday's keynote presentation of the Tokyo Game Show, Nintendo president Satoru Iwata unveiled the controller design for the company's new Nintendo Revolution console. & \{{ \it keynote, Nintendo, Satoru Iwata, Nintendo, Revolution }\} \\ \midrule 
 0 & Resembling a television remote, the controller uses two sensors placed near the user's television to triangulate its position in three-dimensional space. & \{{ \it TV, 3D }\} \\ \midrule 
 0 & This will allow players to control actions and movements in video games by moving the device through the air. & \{{ \it air }\} \\ \midrule 
 0 & Giancarlo Fisichella lost control of his car and ended the race very soon after the start. & \{{ \it Giancarlo Fisichella }\} \\ \midrule 
 0 & His teammate Fernando Alonso was in the lead for most of the race, but ended it right after his pit-stop, probably because a badly tucked right front wheel. & \{{ \it Fernando Alonso }\} \\ \midrule 
 0 & Michael Schumacher ended his race not long after Alonso, because of the suspension damage in the numerous battles during the race. & \{{ \it Michael Schumacher, Alonso, suspension }\} \\ \midrule 
 0 & "She’s very cute and sings quite well, too," he said according to a transcript of the news conference. & \{{ \it  }\} \\ \midrule 
 0 & "I was moved every time we did a rehearsal on this, from the bottom of my heart." & \{{ \it  }\} \\ \midrule 
 0 & Around 3 minutes into the launch, an on-board camera showed numerous pieces of insulation foam break away from the fuel tank. & \{{ \it insulation foam, fuel tank }\} \\ \midrule 
 0 & However, they are not thought to have caused any damage to the shuttle. & \{{ \it  }\} \\ \midrule 
 0 & NASA's shuttle program chief N. Wayne Hale Jr. said the foam had fallen "after the time we are concerned about." & \{{ \it NASA, N. Wayne Hale Jr. }\} \\ \midrule 
 0 & Five minutes into the display a wind starts rolling in, about a minute later, the wind is reaching 70km/h... then the rain comes, but so hard and so large that it slaps your skin like a needle, then hail fell from the sky, people panicking and screaming and running over each other. & \{{ \it km, needle, ice hail }\} \\ 
        \bottomrule
    \end{tabular}
    \label{tab:annotations8}
\end{table*}

\begin{table*}[]
    \centering
    \fontsize{8.4}{10}\selectfont
    \begin{tabular}{cp{10.5cm}p{3.5cm}}
        \toprule
        \textbf{signer} & \textbf{reference} & \textbf{fingerspelled terms} \\
        \midrule
 0 & I lost my sister and her friend, and on my way there were two disabled people in wheelchairs, people just jumping over and pushing them," Armand Versace said. & \{{ \it Armand Versace }\} \\ \midrule 
 0 & NHK also reported that the Kashiwazaki Kariwa nuclear power plant in Niigata prefecture was operating normally. & \{{ \it NHK, nuclear, Kashiwazaki Kariwa, Niigata }\} \\ \midrule 
 0 & Hokuriku Electric Power Co. reported no effects from the earthquake and that the Number 1 and 2 reactors at its Shika nuclear power plant were shut down. & \{{ \it Hokuriku, Co }\} \\ \midrule 
 0 & It is reported that some 9400 homes in the region are without water and approximately 100 without electricity. & \{{ \it  }\} \\ \midrule 
 0 & Some roads have been damaged, railway service interrupted in the affected areas, and the Noto Airport in Ishikawa prefecture remains closed. & \{{ \it bus, Noto, Ishikawa }\} \\ \midrule 
 0 & One bomb exploded outside the governor general's office. & \{{ \it  }\} \\ \midrule 
 0 & Three more bombs exploded near government buildings in a period of two hours. & \{{ \it  }\} \\ \midrule 
 0 & Some reports put the official death toll at eight, and official reports confirm that up to 30 were injured; but final numbers are not yet known. & \{{ \it  }\} \\ \midrule 
 0 & Both cyanuric acid and melamine were found in urine samples from pets that died after consuming contaminated pet food. & \{{ \it cyanuric acid, melamine }\} \\ \midrule 
 0 & The two compounds react with one another to form crystals that may block kidney function, researchers at the university said. & \{{ \it crystal }\} \\ \midrule 
 0 & The researchers observed crystals formed in cat urine by the addition of melamine and cyanuric acid. & \{{ \it melamine, cyanuric acid }\} \\ \midrule 
 0 & The composition of these crystals matches those found in the urine of affected pets when compared by infrared spectroscopy (FTIR). & \{{ \it IR spectroscopy, FTIR }\} \\ \midrule 
 0 & I don't know if you realize it or not, but most of the goods from Central America came into this country duty-free. & \{{ \it if, tax }\} \\ \midrule 
 0 & Yet eighty percent of our goods were taxed through tariffs in Central American countries. we treat you. & \{{ \it tax }\} \\ \midrule 
 0 & That didn't seem to make sense to me; it certainly wasn't fair. & \{{ \it  }\} \\ \midrule 
 0 & All I say to people is you treat us the way we treat you. & \{{ \it  }\} \\ \midrule 
 0 & California Governor Arnold Schwarzenegger signed into law a bill that bans the sale or rental of violent video games to minors. & \{{ \it Arnold Schwarzenegger }\} \\ \midrule 
 0 & The bill requires violent video games sold in the state of California to be labeled with a decal reading "18" and makes their sale to a minor punishable by a fine of \$1000 per offense. & \{{ \it  }\} \\ \midrule 
 0 & The Director of Public Prosecutions, Kier Starmer QC, gave a statement this morning announcing the prosecution of both Huhne and Pryce. & \{{ \it Kier Starmer QC, Huhne, Pryce }\} \\ \midrule 
 0 & Huhne has resigned and he will be replaced in the Cabinet by Ed Davey MP. Norman Lamb MP is expected to take the Business Minister job Davey is vacating. & \{{ \it Huhne, Cabinet, Ed Davey Mp, Norman Lamb MP, minister, Davey }\} \\ \midrule 
 0 & Huhne and Pryce are scheduled to appear at the Westminster Magistrates Court on February 16. & \{{ \it Huhne, Pryce, Westminster Magistrates, Feb }\} \\ \midrule 
 0 & The fatalities were Nicholas Alden, 25, and Zachary Cuddeback, 21. Cuddeback had been the driver. & \{{ \it Nicholas Alden, Zachary Cuddeback, Cuddeback }\} \\ \midrule 
 0 & Edgar Veguilla received arm and jaw wounds while Kristoffer Schneider was left requiring reconstructive surgery for his face. & \{{ \it Edgar Veguilla, Kristoffer Schneider }\} \\ \midrule 
 0 & Uka's weapon failed whilst pointed at a fifth man's head. Schneider has ongoing pain, blindness in one eye, a missing section of skull and a face rebuilt from titanium. & \{{ \it Uka, weapon, Schneider, skull, titanium }\} \\ \midrule 
 0 & Schneider testified via videolink from a USAF base in his homeland. & \{{ \it Schneider, USAF }\} \\ \midrule 
 0 & Beyond Wednesday's event, Carpanedo competed in two individual races at the Championships. & \{{ \it Carpanedo }\} \\ \midrule 
 0 & Her first was the Slalom, where she earned a Did Not Finish in her first run. 36 of the 116 competitors had the same result in that race. & \{{ \it Slalom, Did Not Finish, Did Not Finish }\} \\ \midrule 
 0 & Her other race, the Giant Slalom, saw her finish in tenth in the women's sitting group with a combined run time of 4:41.30, 2:11.60 minutes slower than first place finisher Austrian Claudia Loesch and 1:09.02 minutes slower than the ninth place finisher Gyöngyi Dani of Hungary. & \{{ \it Giant Slalom, sec, Claudia Loesch, sec, Gyöngyi Dani, sec }\} \\ 
        \bottomrule
    \end{tabular}
    \label{tab:annotations9}
\end{table*}

\begin{table*}[]
    \centering
    \fontsize{8.4}{10}\selectfont
    \begin{tabular}{cp{10.5cm}p{3.5cm}}
        \toprule
        \textbf{signer} & \textbf{reference} & \textbf{fingerspelled terms} \\
        \midrule
 0 & Four skiers in the women's sitting group failed to finish their runs, and 45 of the 117 total skiers in the Giant Slalom failed to rank in the race. & \{{ \it Giant Slalom }\} \\ \midrule 
 0 & The Madhya Pradesh Police recovered the stolen laptop and mobile phone. & \{{ \it Madhya Pradesh }\} \\ \midrule 
 0 & Deputy Inspector General D K Arya said, "We have arrested five persons who raped the Swiss woman and recovered her mobile and laptop". & \{{ \it D. K. Arya }\} \\ \midrule 
 0 & The accused are named as Baba Kanjar, Bhutha Kanjar, Rampro Kanjar, Gaza Kanjar and Vishnu Kanjar. & \{{ \it Baba Kanjar, Bhutha Kajar, Rampro Kanjar, Gaza Kanjar, Vishnu Kanjar }\} \\ \midrule 
 0 & Police superintendent Chandra Shekhar Solanki said the accused appeared in court with covered faces. & \{{ \it supint, Chandra Shekhar Solanki }\} \\ \midrule 
 0 & Although three people were inside the house when the car impacted it, none of them were hurt. & \{{ \it  }\} \\ \midrule 
 0 & However, the driver sustained serious injuries to the head. & \{{ \it  }\} \\ \midrule 
 0 & The road where the crash happened was temporarily closed while emergency services freed the driver from the red Audi TT. & \{{ \it Audi TT }\} \\ \midrule 
 0 & He was initially hospitalised in the James Paget Hospital in Great Yarmouth. & \{{ \it James Paget, Great Yarmouth }\} \\ \midrule 
 0 & He was subsequently relocated to Addenbrooke's Hospital in Cambridge. & \{{ \it Addenbrooke, Cambridge }\} \\ \midrule 
 0 & Adekoya has since been in Edinburgh Sheriff Court charged with murdering her son. & \{{ \it Adekoya, Edinburgh }\} \\ \midrule 
 0 & She is in custody pending indictment and trial, but any eyewitness evidence may be tainted because her image has been widely published. & \{{ \it indictment }\} \\ \midrule 
 0 & This is common practice elsewhere in the UK but Scottish justice works differently and courts have viewed publication of photos as potentially prejudicial. & \{{ \it UK }\} \\ \midrule 
 0 & Professor Pamela Ferguson of the University of Dundee notes "journalists do seem to be walking a dangerous line if publishing photos etc of suspects." & \{{ \it Pamela Ferguson, Dundee }\} \\ \midrule 
 0 & Crown Office, which is in overall charge of prosecutions, has indicated to journalists that no further comment will be made at least until indictment. & \{{ \it Crown, news, indictment }\} \\ \midrule 
 0 & The document, according to the leak, will refer to the borders dispute, which Palestine wants based on the borders before the 1967 Mideast War. & \{{ \it Palestine }\} \\ \midrule 
 0 & Other topics covered reportedly include the future state of Jerusalem which is sacred to both nations and the Jordan Valley issue. & \{{ \it Jordan }\} \\ \midrule 
 0 & Israel demands an ongoing military presence in the valley for ten years once an agreement is signed while the PA agrees to leave such presence only for five years. & \{{ \it PA }\} \\ \midrule 
 0 & Shooters in the supplementary pest control trial were to be closely supervised by rangers, as the trial was monitored and its effectiveness evaluated. & \{{ \it  }\} \\ \midrule 
 0 & In a partnership of NPWS and the Sporting Shooters Association of Australia (NSW) Inc, qualified volunteers were recruited, under the Sporting Shooters Association's hunting program. & \{{ \it NPWS, NSW, Inc }\} \\ \midrule 
 0 & According to Mick O'Flynn, the Acting Director Park Conservation and Heritage with the NPWS, the four shooters selected for the first shooting operation received comprehensive safety and training instruction. & \{{ \it Mick O'Flynn, NPWS }\} \\ \midrule 
 0 & Martelly swore in a new Provisional Electoral Council (CEP) of nine members yesterday. & \{{ \it Martelly, CEP }\} \\ \midrule 
 0 & It is Martelly's fifth CEP in four years. & \{{ \it Martelly, CEP }\} \\ \midrule 
 0 & Last month a presidential commission recommended the prior CEP's resignation as part of a package of measures to move the country towards new elections. & \{{ \it CEP }\} \\ \midrule 
 0 & The commission was Martelly's response to widespread anti-regime protests that started in October. & \{{ \it Martelly, Oct }\} \\ \midrule 
 0 & The sometimes-violent protests were triggered by failure to hold elections, some due since 2011. & \{{ \it  }\} \\ \midrule 
 0 & Around 60 cases of malfunctioning iPods overheating have been reported, causing a total of six fires and leaving four people with minor burns. & \{{ \it iPod, burn, burn }\} \\ \midrule 
 0 & Japan's Ministry of Economy, Trade and Industry (METI) said that it had been aware of 27 accidents related to the devices. & \{{ \it METI, iPod }\} \\ \midrule 
 0 & Last week, METI announced that Apple had informed it of 34 additional overheating incidents, which the company called "non-serious." & \{{ \it METI, non-serious }\} \\
        \bottomrule
    \end{tabular}
    \label{tab:annotations10}
\end{table*}

\begin{table*}[]
    \centering
    \fontsize{8.4}{10}\selectfont
    \begin{tabular}{cp{10.5cm}p{3.5cm}}
        \toprule
        \textbf{signer} & \textbf{reference} & \textbf{fingerspelled terms} \\
        \midrule
 0 & The ministry responded by calling Apple's postponement of the report "truly regrettable." & \{{ \it  }\} \\ \midrule 
 0 & The eathquake struck Mariana at 07:19 a.m. local time (09:19 p.m. GMT Friday). & \{{ \it Mariana, GMT }\} \\ \midrule 
 0 & The Northern Marianas emergency management office said that there were no damages reported in the nation. & \{{ \it Marianas }\} \\ \midrule 
 0 & Also the Pacific Tsunami Warning Center said that there was no Tsunami indication. & \{{ \it Pacific }\} \\ \midrule 
 0 & A former Filipino policeman has kept Hong Kong tourists hostage by hijacking their bus in Manila, the capital of the Philippines. & \{{ \it Hong Kong, bus, Manila }\} \\ \midrule 
 0 & Rolando Mendoza fired his M16 rifle at the tourists. & \{{ \it Rolando Mendoza, M16 }\} \\ \midrule 
 0 & Several hostages have been rescued and least six have been confirmed dead so far. & \{{ \it  }\} \\ \midrule 
 0 & Six hostages, including the children and elderly, were released early, as were the Filipino photographers. & \{{ \it  }\} \\ \midrule 
 0 & The photographers later took the place of an aged lady as she needed the lavatory. Mendoza was gunned down. & \{{ \it Mendoza }\} \\ \midrule 
 0 & Liggins followed in his father’s footsteps and entered a career in medicine. & \{{ \it Liggins }\} \\ \midrule 
 0 & He trained as an obstetrician and began to work at the Auckland's National Women's Hospital in 1959. & \{{ \it obstetrician, Auckland }\} \\ \midrule 
 0 & While he was working at the hospital Liggins began to investigate premature labor during his spare time. & \{{ \it  }\} \\ \midrule 
 0 & His research showed that if a hormone was administered it would speed up the baby's foetal lung maturation. & \{{ \it hormone }\} \\ \midrule 
 0 & Xinhua reported that government investigators recovered two 'black box' flight recorders on Wednesday. & \{{ \it Xinhua, black box }\} \\ \midrule 
 0 & Fellow wrestlers also paid tribute to Luna. & \{{ \it Luna }\} \\ \midrule 
 0 & Tommy Dreamer said "Luna was the first Queen of Extreme. My first manager. Luna passed away on the night of two moons. Pretty unique just like her. Strong woman." & \{{ \it Tommy Dreamer, Luna, Luna }\} \\ \midrule 
 0 & Dustin "Goldust" Runnels commented that "Luna was as freaky as me...maybe even more...love her and will miss her...hopefully she's in a better place." & \{{ \it Dustin Runnels, Goldust, Luna, freaky }\} \\ \midrule 
 0 & Out of 1,400 people polled prior to the 2010 federal election, those who oppose Australia becoming a republic grew by 8 per cent since 2008. & \{{ \it republic }\} \\ \midrule 
 0 & Caretaker Prime Minister Julia Gillard claimed during the campaign of the 2010 federal election that she believed Australia should become a republic at the end of Queen Elizabeth II's reign. & \{{ \it PM Julia Gillard, Elizabeth }\} \\ \midrule 
 0 & 34 per cent of those in the poll share this view, wanting Queen Elizabeth II to be Australia's last monarch. & \{{ \it Elizabeth }\} \\ \midrule 
 0 & At the extremes of the poll, 29 per cent of those surveyed believe Australia should become a republic as soon as possible, while 31 per cent believe Australia should never become a republic. & \{{ \it  }\} \\ \midrule 
 0 & The Olympic gold medalist was due to swim in the 100m and 200m freestyle and in three relays at the Commonwealth Games, but due to his complaints his fitness has been in doubt. & \{{ \it 100m, 200m, fs, Commonwealth }\} \\ \midrule 
 0 & He has been unable to take the drugs needed to overcome his pain as they are banned from the Games. & \{{ \it  }\} \\ \midrule 
 0 & Curtis Cooper, a mathematician and computer science professor at the University of Central Missouri, has discovered the largest known prime number to date on January 25. & \{{ \it Curtis Cooper, Central, Miss, prime, Jan }\} \\ \midrule 
 0 & Several people verified the discovery using different hardware and software by the beginning of February and it was announced on Tuesday. & \{{ \it HW, SW, Feb }\} \\ \midrule 
 0 & Comets may possibly have been a source of water delivery to the earth along with organic matter that can form proteins and support life. & \{{ \it comet, organic matter, protein }\} \\ \midrule 
 0 & Scientists hope to understand how planets form, especially how the Earth formed, since comets collided with the Earth long ago. & \{{ \it  }\} \\ \midrule 
 0 & Cuomo, 53, began his governorship earlier this year and signed a bill last month legalizing same-sex marriage. & \{{ \it Cuomo, bill, same sex }\} \\ \midrule 
 0 & He referred to the rumors as "political chatter and silliness". & \{{ \it  }\} \\ \midrule 
 0 & He is speculated to make a run for president in 2016. & \{{ \it  }\} \\
        \bottomrule
    \end{tabular}
    \label{tab:annotations11}
\end{table*}

\begin{table*}[]
    \centering
    \fontsize{8.4}{10}\selectfont
    \begin{tabular}{cp{10.5cm}p{3.5cm}}
        \toprule
        \textbf{signer} & \textbf{reference} & \textbf{fingerspelled terms} \\
        \midrule
 0 & NextGen is a system the FAA claims would allow aircraft to fly shorter routes and save millions of gallons of fuel each year and cut carbon emissions. & \{{ \it NextGen, FAA, gallon, carbon emission }\} \\ \midrule 
 0 & It uses satellite-based technology as opposed to older ground-radar-based technology to allow air traffic controllers to pinpoint aircraft with greater precision and give pilots more accurate information. & \{{ \it  }\} \\ \midrule 
 0 & No extra transport is being put on and overground trains will not stop at Wembley, and car parking and park-and-ride facilities are unavailable at the ground. & \{{ \it Wembley }\} \\ \midrule 
 0 & Fears of lack of transportation raised the possibility that the game would be forced to play behind closed doors without the team's supporters. & \{{ \it  }\} \\ \midrule 
 0 & A study published on Thursday in the journal Science reported on formation of a new bird species on the Ecuadorean Galápagos Islands. & \{{ \it Th, Galapagos }\} \\ \midrule 
 0 & Researchers from Princeton University in the United States and Uppsala University in Sweden reported the new species evolved in just two generations, though this process had been believed to take much longer, due to breeding between an endemic Darwin finch, Geospiza fortes, and the immigrant cactus finch, Geospiza conirostris. & \{{ \it Princeton, Uppsala, endemic Darwin finch, Geospiza fortes, immigrant cactus finch, Geospiza conirostris }\} \\ \midrule 
 0 & Gold may be worked into all sorts of shapes. It can be rolled into tiny shapes. & \{{ \it  }\} \\ \midrule 
 0 & It can be pulled into thin wire, which can be twisted and plaited. It can be hammered or rolled into sheets. & \{{ \it or }\} \\ \midrule 
 0 & It can be made very thin, and stuck onto other metal. It can be made so thin that it was sometimes used to decorate the hand-painted pictures in books called "illuminated manuscripts". & \{{ \it illuminated manuscripts }\} \\ \midrule 
 0 & This is called a chemical's pH. You can make an indicator using red cabbage juice. & \{{ \it pH, pH, cabbage }\} \\ \midrule 
 0 & The cabbage juice changes color depending on how acidic or basic (alkaline) the chemical is. & \{{ \it cabbage, acidic, basic, alkaine }\} \\ \midrule 
 0 & The pH level is indicated by the amount of Hydrogen (the H in pH) ions in the tested chemical. & \{{ \it pH, hydrogen, H, ph, ions }\} \\ \midrule 
 0 & Hydrogen ions are protons that had their electrons stripped off them (since Hydrogen atoms consist of one proton and one electron). & \{{ \it ions, protons, electrons, proton, electron }\} \\ \midrule 
 0 & Swirl the two dry powders together and then, with clean wet hands, squeeze them into a ball. & \{{ \it powder }\} \\ \midrule 
 0 & The moisture on your hands will react with the outer layers, which will feel funny and form a sort of shell. & \{{ \it shell }\} \\
1 & This not only confirms that at least some dinosaurs had feathers, a theory already widespread, but provides details fossils generally cannot, such as color and three-dimensional arrangement. & \{{ \it Feather, Fossil }\} \\ \midrule 
 1 & . Scientists say this animal's plumage was chestnut-brown on top with a pale or carotenoid-colored underside. & \{{ \it chestnut, carotenoid }\} \\ \midrule 
 1 & The find also grants insight into the evolution of feathers in birds. & \{{ \it feather }\} \\ \midrule 
 1 & Because the dinosaur feathers do not have a well-developed shaft, called a rachis, but do have other features of feathers — barbs and barbules — the researchers inferred the rachis was likely a later evolutionary development that these other features. & \{{ \it rachis, barbs, barbules }\} \\ \midrule 
 1 & The feathers' structure suggests that they were not used in flight but rather for temperature regulation or display. & \{{ \it  }\} \\ \midrule 
 1 & The researchers suggested that, even though this is the tail of a young dinosaur, the sample shows adult plumage and not a chick's down. & \{{ \it tail, plumage, down }\} \\ \midrule 
 1 & ASUS Eee PC, earlier launched world-wide for cost-saving and functionality factors, became a hot topic in 2007 Taipei IT Month. & \{{ \it Taipei, IT, ASUS Eee PC }\} \\ \midrule 
 1 & But the consumer market on laptop computer will be radically varied and changed after ASUS was awarded in the 2007 Taiwan Sustainable Award by Executive Yuan of the Republic of China. & \{{ \it sustainable, Yuan, Republic }\} \\ \midrule 
 1 & In the warm climate of the Middle East, the house was not so important. & \{{ \it  }\} \\ \midrule 
 1 & Most of the life of the Hebrew family happened in the open air. & \{{ \it  }\} \\ \midrule 
 1 & Women did the cooking in the yard; stores were just open counters looking into the street. Stone was used for building houses. & \{{ \it  }\} \\
        \bottomrule
    \end{tabular}
    \label{tab:annotations12}
\end{table*}

\begin{table*}[]
    \centering
    \fontsize{8.4}{10}\selectfont
    \begin{tabular}{cp{10.5cm}p{3.5cm}}
        \toprule
        \textbf{signer} & \textbf{reference} & \textbf{fingerspelled terms} \\
        \midrule
1 & There were no large forests in the land of Canaan, so wood was extremely expensive. & \{{ \it Canaan }\} \\ \midrule 
 1 & Ottawa is Canada's charming, bilingual capital and features an array of art galleries and museums that showcase Canada's past and present. & \{{ \it Ottawa }\} \\ \midrule 
 1 & Farther south is Niagara Falls and the north is home to the untapped natural beauty of the Muskoka and beyond. & \{{ \it Niagara Falls, Muskoka }\} \\ \midrule 
 1 & All these things and more highlight Ontario as what is considered quintessentially Canadian by outsiders. & \{{ \it  }\} \\ \midrule 
 1 & Large areas further north are quite sparsely populated and some is nearly uninhabited wilderness. & \{{ \it  }\} \\ \midrule 
 1 & For a comparison of population that surprises many: There are more African Americans living in the US than there are Canadian citizens. & \{{ \it  }\} \\ \midrule 
 1 & Needless to say, if you know a Romance language, it will be easier for you to learn Portuguese. & \{{ \it Romance }\} \\ \midrule 
 1 & However, people who know a little Spanish may hastily conclude that Portuguese is close enough that it need not be studied separately. & \{{ \it  }\} \\ \midrule 
 1 & Widespread looting reportedly continued overnight, as law enforcement officers were not present on Bishkek's streets. & \{{ \it  }\} \\ \midrule 
 1 & Bishkek was described as sinking into a state of "anarchy" by one observer, as gangs of people roamed the streets and plundered stores of consumer goods. & \{{ \it Bishkek }\} \\ \midrule 
 1 & Several Bishkek residents blamed protesters from the south for the lawlessness. & \{{ \it  }\} \\ \midrule 
 1 & Other subjects on the agenda in Bali include saving the world's remaining forests, and sharing technologies to help developing nations grow in less-polluting ways. & \{{ \it Bali, agenda }\} \\ \midrule 
 1 & The U.N. also hopes to finalize a fund to help countries affected by global warming to cope with the impacts. & \{{ \it UN, global warming }\} \\ \midrule 
 1 & The money could go toward flood-proof houses, better water management, and crop diversification. & \{{ \it crop,  diversification }\} \\ \midrule 
 1 & Thus, the pencil was a good friend to many people when it came out. & \{{ \it  }\} \\ \midrule 
 1 & Sadly, as newer methods of writing have emerged, the pencil has been relegated to lesser status and uses. & \{{ \it  }\} \\ \midrule 
 1 & People now write messages on computer screens, never having to come close to a sharpener. & \{{ \it sharpener }\} \\ \midrule 
 1 & One can only wonder what the keyboard will become when something newer comes along. & \{{ \it  }\} \\ \midrule 
 1 & The movie, featuring Ryan Gosling and Emma Stone, received nominations in all major categories. & \{{ \it Ryan Gosling, Emma Stone }\} \\ \midrule 
 1 & Gosling and Stone received nominations for Best Actor and Actress respectively. & \{{ \it  }\} \\ \midrule 
 1 & The other nominations include Best Picture, Director, Cinematography, Costume Design, Film-editing, Original Score, Production Design, Sound Editing, Sound Mixing and Original Screenplay. & \{{ \it  }\} \\ \midrule 
 1 & Two songs from the movie, Audition (The Fools Who Dream) and City of Stars, received nominations for best original song. Lionsgate studio received 26 nominations — more than any other studio. & \{{ \it Audition (The Fools Who Dream), City of Stars, Lionsgate, studio, studio }\} \\ \midrule 
 1 & For example, the most common still image photography format in the world is 35mm, which was the dominant film size at the close of the analog film era. & \{{ \it still image, 35mm, analog film era }\} \\ \midrule 
 1 & It is still produced today, but more importantly its aspect ratio has been inherited by digital camera image sensor formats. & \{{ \it 35mm, aspect ratio, digital, digital image, sensor }\} \\ \midrule 
 1 & The 35mm format is actually, somewhat confusingly, 36mm in width by 24mm in height. & \{{ \it 35mm, 36mm, 36mm, 24mm }\} \\ \midrule 
 1 & The aspect ratio of this format (dividing by twelve to obtain the simplest whole-number ratio) is therefore said to be 3:2. & \{{ \it aspect ratio }\} \\ \midrule 
 1 & Many common formats (APS family of formats, for example) are equal to or closely approximate this aspect ratio. & \{{ \it APS format }\} \\ \midrule 
 1 & The Madhya Pradesh Police recovered the stolen laptop and mobile phone. & \{{ \it Madhya Pradesh }\} \\ \midrule 
 1 & Deputy Inspector General D K Arya said, "We have arrested five persons who raped the Swiss woman and recovered her mobile and laptop". & \{{ \it D. K. Arya }\} \\
        \bottomrule
    \end{tabular}
    \label{tab:annotations13}
\end{table*}

\begin{table*}[]
    \centering
    \fontsize{8.4}{10}\selectfont
    \begin{tabular}{cp{10.5cm}p{3.5cm}}
        \toprule
        \textbf{signer} & \textbf{reference} & \textbf{fingerspelled terms} \\
        \midrule
 1 & The accused are named as Baba Kanjar, Bhutha Kanjar, Rampro Kanjar, Gaza Kanjar and Vishnu Kanjar. & \{{ \it Baba Kanjar, Bhutha Kanjar, Rampro Kanjar, Gaza Kanjar, Vishnu Kanjar }\} \\ \midrule 
 1 & Police superintendent Chandra Shekhar Solanki said the accused appeared in court with covered faces. & \{{ \it Chandra Shekhar Solanki }\} \\ \midrule 
 1 & There are also a few buses going north to Hebron, the traditional burial place of the Biblical patriarchs Abraham, Isaac, Jacob, and their wives. & \{{ \it bus, Hebron, patriarch, Abraham, Isaac, Jacob }\} \\ \midrule 
 1 & Check that the bus you are thinking of taking goes into Hebron and not just to the nearby Jewish settlement of Kiryat Arba. & \{{ \it bus, Hebron, Kiryat Arba }\} \\ \midrule 
 1 & This is called a chemical's pH. You can make an indicator using red cabbage juice. & \{{ \it pH, juice }\} \\ \midrule 
 1 & The cabbage juice changes color depending on how acidic or basic (alkaline) the chemical is. & \{{ \it juice, acidic, basic, basic, alkaline }\} \\ \midrule 
 1 & The pH level is indicated by the amount of Hydrogen (the H in pH) ions in the tested chemical. & \{{ \it pH, ion }\} \\ \midrule 
 1 & Hydrogen ions are protons that had their electrons stripped off them (since Hydrogen atoms consist of one proton and one electron). & \{{ \it proton, electron, atom }\} \\ \midrule 
 1 & The games kicked off at 10:00am with great weather and apart from mid morning drizzle which quickly cleared up, it was a perfect day for 7's rugby. & \{{ \it  }\} \\ \midrule 
 1 & Tournament top seeds South Africa started on the right note when they had a comfortable 26 - 00 win against 5th seeded Zambia. & \{{ \it  }\} \\ \midrule 
 1 & Looking decidedly rusty in the game against their southern sisters, South Africa however steadily improved as the tournament progressed. & \{{ \it  }\} \\ \midrule 
 1 & Their disciplined defence, ball handling skills and excellent team work made them stand out and it was clear that this was the team to beat. & \{{ \it ball }\} \\ \midrule 
 1 & Although AI has a strong connotation of science fiction, AI forms a very important branch of computer science, dealing with behavior, learning and intelligent adaptation in a machine. & \{{ \it AI, scifi, AI }\} \\ \midrule 
 1 & Research in AI involves making machines to automate tasks that require intelligent behavior. & \{{ \it AI }\} \\ \midrule 
 1 & Examples include control, planning and scheduling, the ability to answer customer diagnoses and questions, as well as handwriting recognition, voice and face. & \{{ \it dx }\} \\ \midrule 
 1 & Such things have become separate disciplines, which focus on providing solutions to real life problems. & \{{ \it  }\} \\ \midrule 
 1 & The AI system is now often used in the fields of economics, medicine, engineering and the military, as has been built in several home computer and video game software applications. & \{{ \it AI, AI }\} \\ \midrule 
 1 & Backpacking by ski: This activity is also called backcountry ski, ski touring or ski hiking. & \{{ \it backcountry, touring, hiking }\} \\ \midrule 
 1 & It is related to but usually not involving alpine style ski touring or mountaineering, the latter ones done in steep terrain and requiring much stiffer skis and boots. & \{{ \it alpine, stiff }\} \\ \midrule 
 1 & Think of the skiing route as of a similar hiking route. & \{{ \it  }\} \\ \midrule 
 1 & In good conditions you will be able to cover somewhat greater distances than walking – but only very seldom you will get the speeds of cross country skiing without a heavy backpack in groomed tracks. & \{{ \it XC, groomed }\} \\ \midrule 
 1 & Italy's national football, along with German national football team is the second most successful team in the world and were the FIFA World Cup champions in 2006. & \{{ \it FIFA World Cup }\} \\ \midrule 
 1 & Popular sports include football, basketball, volleyball, water-polo, fencing, rugby, cycling, ice hockey, roller hockey and F1 motor racing. & \{{ \it ice, F1 }\} \\ \midrule 
 1 & Winter sports are most popular in the Northern regions, with Italians competing in international games and Olympic events. & \{{ \it  }\} \\ \midrule 
 1 & The Olympic gold medalist was due to swim in the 100m and 200m freestyle and in three relays at the Commonwealth Games, but due to his complaints his fitness has been in doubt. & \{{ \it Commonwealth Games, freestyle }\} \\ \midrule
 1 & He has been unable to take the drugs needed to overcome his pain as they are banned from the Games. & \{{ \it  }\} \\
        \bottomrule
    \end{tabular}
    \label{tab:annotations14}
\end{table*}

\begin{table*}[]
    \centering
    \fontsize{8.4}{10}\selectfont
    \begin{tabular}{cp{10.5cm}p{3.5cm}}
        \toprule
        \textbf{signer} & \textbf{reference} & \textbf{fingerspelled terms} \\
        \midrule
 1 & Rip currents are the returning flow from waves breaking off the beach, often at a reef or similar. & \{{ \it beach, rip current }\} \\ \midrule 
 1 & Due to the underwater topology the return flow is concentrated at a few deeper sections, and a fast current to deep water may form there. & \{{ \it topology }\} \\ \midrule 
 1 & Most deaths happen as result of fatigue trying to swim back against the current, which may be impossible. & \{{ \it  }\} \\ \midrule 
 1 & As soon as you get out of the current, swimming back is no more difficult than normally. & \{{ \it  }\} \\ \midrule 
 1 & Try aiming somewhere where you are not caught again or, depending on your skills and on whether you have been noticed, you might want to wait for rescue. & \{{ \it  }\} \\ \midrule
 1 & Previously, Ring's CEO, Jamie Siminoff, remarked the company started when his doorbell wasn't audible from his shop in his garage. & \{{ \it Ring, Jamie Siminoff, CEO }\} \\ \midrule 
 1 & He built a WiFi door bell, he said. & \{{ \it WiFi }\} \\ \midrule 
 1 & Siminoff said sales boosted after his 2013 appearance in a Shark Tank episode where the show panel declined funding the startup. & \{{ \it Shark Tank, sales }\} \\ \midrule 
 1 & In late 2017, Siminoff appeared on shopping television channel QVC. & \{{ \it QVC }\} \\ \midrule 
 1 & Ring also settled a lawsuit with competing security company, the ADT Corporation. & \{{ \it ADT }\} \\ \midrule 
 1 & Combined with its relative inaccessibility, "Timbuktu" has come to be used as a metaphor for exotic, distant lands. & \{{ \it Bali, Timbuktu }\} \\ \midrule 
 1 & Today, Timbuktu is an impoverished town, although its reputation makes it a tourist attraction, and it has an airport. & \{{ \it  }\} \\ \midrule 
 1 & In 1990, it was added to the list of world heritage sites in danger, due to the threat of desert sands. & \{{ \it sand, heritage site }\} \\ \midrule 
 1 & It was one of the major stops during Henry Louis Gates' PBS special Wonders of the African World. & \{{ \it Henry Louis Gates, PBS }\} \\ \midrule 
 1 & The city is in stark contrast to the rest of the country's cities, because it has more of an Arabic flair than of an African. & \{{ \it Timbuktu, Arabic }\} \\ \midrule 
 1 & Though it is widely used, especially among non-Romani, the word "Gypsy" is often considered offensive because of its associations with negative stereotypes and inaccurate perceptions of Romani people. & \{{ \it gypsy, Romani, gypsy, stereotypes, Romani }\} \\ \midrule 
 1 & The satellites, both of which weighed in excess of 1,000 pounds, and traveling at approximately 17,500 miles per hour, collided 491 miles above the Earth. & \{{ \it satellites, mph, miles }\} \\ \midrule 
 1 & Scientists say the explosion caused by the collision was massive. & \{{ \it  }\} \\ \midrule 
 1 & They are still trying to determine just how large the crash was and how the Earth will be affected. & \{{ \it  }\} \\ \midrule 
 1 & The United States Strategic Command of the U.S. Department of Defense office is tracking the debris. & \{{ \it US, US, debris }\} \\ \midrule 
 1 & The result of plotting analysis will be posted to a public website. & \{{ \it  }\} \\ \midrule 
 1 & Apia is the capital of Samoa. The town is on the island of Upolu and has a population of just under 40,000. & \{{ \it Samoa, Upolu, Apia }\} \\ \midrule 
 1 & Apia was founded in the 1850s and has been the official capital of Samoa since 1959. & \{{ \it Apia }\} \\ \midrule 
 1 & The harbor was the site of an infamous naval standoff in 1889 when seven ships from Germany, the US, and Britain refused to leave the harbor. & \{{ \it US, Mau }\} \\ \midrule 
 1 & All the ships were sunk, except for one British cruiser. Nearly 200 American and German lives were lost. & \{{ \it  }\} \\ \midrule 
 1 & During the struggle for independence organised by the Mau movement, a peaceful gathering in the town resulted in the killing of the paramount chief Tupua Tamasese Lealofi III. & \{{ \it Tupua Tamasese Lealofi }\} \\ \midrule 
 1 & Since Montevideo is south of the Equator, it is summer there when it's winter in the Northern Hemisphere and vice versa. & \{{ \it Montevideo }\} \\ \midrule 
 1 & Montevideo is in the subtropics; in the summer months, temperatures above +30°C are common. & \{{ \it subtropics }\} \\ \midrule 
 1 & The winter can be deceptively chilly: temperatures rarely go below freezing, but the wind and humidity combine to make it feel colder than what the thermometer says. & \{{ \it  }\} \\
        \bottomrule
    \end{tabular}
    \label{tab:annotations15}
\end{table*}

\begin{table*}[]
    \centering
    \fontsize{8.4}{10}\selectfont
    \begin{tabular}{cp{10.5cm}p{3.5cm}}
        \toprule
        \textbf{signer} & \textbf{reference} & \textbf{fingerspelled terms} \\
        \midrule
1 & There are no particular "rainy" and "dry" seasons: the amount of rain stays roughly the same throughout the year. & \{{ \it  }\} \\ \midrule 
 1 & All animals that originally arrived in the islands came here either by swimming, flying or floating. & \{{ \it Galapagos }\} \\ \midrule 
 1 & Due to the long distance from the continent mammals were unable to make the journey making the giant tortoise the primary grazing animal in the Galapagos. & \{{ \it tortoise }\} \\ \midrule 
 1 & Since the arrival of man to the Galapagos, many mammals have been introduced including goats, horses, cows, rats, cats and dogs. & \{{ \it Galapagos }\} \\ \midrule 
 1 & Every morning, people leave small country towns in cars to go their workplace and are passed by others whose work destination is the place they have just left. & \{{ \it car }\} \\ \midrule 
 1 & In this dynamic transport shuttle everyone is somehow connected with, and supporting, a transport system based on private cars. & \{{ \it car }\} \\ \midrule 
 1 & Science now indicates that this massive carbon economy has dislodged the biosphere from one of its stable states that has supported human evolution for the past two million years. & \{{ \it biosphere }\} \\ \midrule 
 1 & Sleep interruption is the process of purposefully awakening during your normal sleep period and falling asleep a short time later (10–60 minutes). & \{{ \it biorhythm }\} \\ \midrule 
 1 & This can be easily done by using a relatively quiet alarm clock to bring you to consciousness without fully waking you. & \{{ \it  }\} \\ \midrule 
 1 & If you find yourself resetting the clock in your sleep, it can be placed on the other side of the room, forcing you to get out of bed to turn it off. & \{{ \it  }\} \\ \midrule 
 1 & Other biorhythm-based options involve drinking lots of fluid (particularly water or tea, a known diuretic) prior to sleep, forcing one to get up to urinate. & \{{ \it biorhythm }\} \\ \midrule 
 1 & Buses depart the inter-district bus station (across the river) throughout the day, though most, especially those heading to the east and Jakar/Bumthang leave between 06:30 and 07:30. & \{{ \it bus, bus, Jakar Bumthang }\} \\ \midrule
  1 & As the inter-district buses are often full, it is advisable to purchase a ticket a few days in advance. & \{{ \it bus }\} \\ \midrule 
 1 & Most districts are served by small Japanese Coaster Buses, which are comfortable and sturdy. & \{{ \it by, coaster buses }\} \\ \midrule 
 1 & Shared taxis are a quick and comfortable means to travel to nearby places, such as Paro (Nu 150) and Punakha (Nu 200). & \{{ \it taxi, paro nu punakha nu }\} \\ \midrule 
 1 & Science’s main goal is to figure out the way the world works through the scientific method. This method in fact guides most scientific research. & \{{ \it  }\} \\ \midrule 
 1 & It isn’t alone though, experimentation, and an experiment is a test that is used to eliminate one or more of the possible hypotheses, asking questions, and making observations also guide scientific research. & \{{ \it hypotheses }\} \\ \midrule 
 1 & Blogs can also help improve student writing. While students often begin their blog experience with sloppy grammar and spelling, the presence of an audience generally changes that. & \{{ \it log, tool }\} \\ \midrule 
 1 & Since students are often the most critical audience, the blog writer begins to strive to improve writing to avoid criticism. & \{{ \it  }\} \\ \midrule 
 1 & Also blogging "forces students to become more savvy about the world around them." The need to feed the interest of the audience inspires students to be clever and interesting (Toto, 2004). & \{{ \it Toto, savvy }\} \\ \midrule 
 1 & Blogging is a tool that inspires collaboration, and encourages students to extend learning well beyond the traditional school day. & \{{ \it tool }\} \\ \midrule 
 1 & Appropriate use of blogs "can empower students to become more analytical and critical; through actively responding to Internet materials, students can define their positions in the context of others' writings as well as outline their own perspectives on particular issues (Oravec, 2002). & \{{ \it Oravec, blogging }\} \\ \midrule 
 1 & One of the most common methods used to illustrate the importance of socialization is to draw upon the few unfortunate cases of children who were, through neglect, misfortune, or wilful abuse, not socialized by adults while they were growing up. & \{{ \it abuse }\} \\ \midrule 
 1 & Such children are called "feral" or wild. Some feral children have been confined by people (usually their own parents); in some cases this child abandonment was due to the parents' rejection of a child's severe intellectual or physical impairment. & \{{ \it feral, wild, own }\} \\
        \bottomrule
    \end{tabular}
    \label{tab:annotations16}
\end{table*}

\begin{table*}[]
    \centering
    \fontsize{8.4}{10}\selectfont
    \begin{tabular}{cp{10.5cm}p{3.5cm}}
        \toprule
        \textbf{signer} & \textbf{reference} & \textbf{fingerspelled terms} \\
        \midrule
1 & Feral children may have experienced severe child abuse or trauma before being abandoned or running away. & \{{ \it abuse }\} \\ \midrule 
 1 & Others are alleged to have been brought up by animals; some are said to have lived in the wild on their own. & \{{ \it  }\} \\ \midrule 
 1 & When completely brought up by non-human animals, the feral child exhibits behaviors (within physical limits) almost entirely like those of the particular care-animal, such as its fear of or indifference to humans. & \{{ \it  }\} \\ \midrule 
 1 & The stretch between Point Marion and Fairmont presents the most challenging driving conditions on the Buffalo-Pittsburgh Highway, passing frequently through isolated backwoods terrain. & \{{ \it Point Marion, Fairmont }\} \\ \midrule 
 1 & If you're not used to driving on country roads, keep your wits about you: steep grades, narrow lanes, and sharp curves predominate. & \{{ \it  }\} \\ \midrule 
 1 & Posted speed limits are noticeably lower than in previous and subsequent sections — commonly 35-40 mph (56-64 km/h) — and strict obedience to them is even more important than otherwise. & \{{ \it low, mph, km/h }\} \\ \midrule 
 1 & Curiously, though, mobile phone service is much stronger here than along many other stretches of the route, e.g. the Pennsylvania Wilds. & \{{ \it Pennsylvania Wilds, park }\} \\ \midrule 
 1 & The 802.11n standard operates on both the 2.4Ghz and 5.0Ghz frequencies. & \{{ \it 802.11n, 2.4Ghz, 5.0Ghz }\} \\ \midrule 
 1 & This will allow it to be backwards compatible with 802.11a, 802.11b and 802.11g, provided that the base station has dual radios. & \{{ \it 802.11a, 802.11b, 802.11g, radio }\} \\ \midrule 
 1 & The speeds of 802.11n are substantially faster than that of its predecessors with a maximum theoretical throughput of 600Mbit/s. & \{{ \it 802.11n, 600Mbit/s }\} \\ \midrule 
 1 & Curtis Cooper, a mathematician and computer science professor at the University of Central Missouri, has discovered the largest known prime number to date on January 25. & \{{ \it Curtis Cooper, prime, January }\} \\ \midrule 
 1 & Several people verified the discovery using different hardware and software by the beginning of February and it was announced on Tuesday. & \{{ \it February }\} \\ \midrule 
 1 & The amount of inner peace a person possesses correlates oppositely to the amount of tension in one’s body and spirit. & \{{ \it  }\} \\ \midrule 
 1 & The lower the tension, the more positive the life force present. Every person has the potential to find absolute peace and contentment. & \{{ \it  }\} \\ \midrule 
 1 & Everyone can achieve enlightenment. The only thing standing in the way of this goal is our own tension and negativity. & \{{ \it  }\} \\ \midrule 
 1 & South Africa have defeated the All Blacks (New Zealand) in a rugby union Tri Nations match at the Royal Bafokeng Stadium in Rustenburg, South Africa. & \{{ \it Tri Nations match }\} \\ \midrule 
 1 & The final score was a one-point victory, 21 to 20, ending the All Blacks' 15 game winning streak. & \{{ \it  }\} \\ \midrule 
 1 & For the Springboks, it ended a five-match losing streak. & \{{ \it Springboks }\} \\ \midrule 
 1 & It was the final match for the All Blacks, who had already won the trophy two weeks ago. & \{{ \it  }\} \\ \midrule 
 1 & The final match of the series will take place at Ellis Park in Johannesburg next week, when the Springboks play Australia. & \{{ \it Ellis Park }\} \\ \midrule 
 1 & Re-entry shock comes on sooner than culture shock (there's less of a honeymoon phase), lasts longer, and can be more severe. & \{{ \it  }\} \\ \midrule 
 1 & Travellers who had an easy time adjusting to the new culture sometimes have a particularly hard time readjusting to their native culture. & \{{ \it  }\} \\ \midrule 
 1 & When returning home after living abroad, you've adapted to the new culture and lost some of your habits from your home culture. & \{{ \it  }\} \\ \midrule 
 1 & When you went abroad at first, people were probably patient and understanding, knowing that travellers in a new country need to adapt. & \{{ \it  }\} \\ \midrule 
 1 & People may not anticipate that patience and understanding are also necessary for travellers returning home. & \{{ \it  }\} \\ \midrule 
 1 & Muhammad was deeply interested in matters beyond this mundane life. He used to frequent a cave that became known as “Hira‘” on the Mountain of “Noor” (light) for contemplation. & \{{ \it Muhammad, cave, Noor, Hira }\} \\ \midrule 
 1 & he cave itself, which survived the times, gives a very vivid image of Muhammad’s spiritual inclinations. & \{{ \it  }\} \\
        \bottomrule
    \end{tabular}
    \label{tab:annotations17}
\end{table*}

\begin{table*}[]
    \centering
    \fontsize{8.4}{10}\selectfont
    \begin{tabular}{cp{10.5cm}p{3.5cm}}
        \toprule
        \textbf{signer} & \textbf{reference} & \textbf{fingerspelled terms} \\
        \midrule
1 & Resting on the top of one of the mountains north of Mecca, the cave is completely isolated from the rest of the world. & \{{ \it Mecca, cave }\} \\ \midrule 
 1 & In fact, it is not easy to find at all even if one knew it existed. Once inside the cave, it is a total isolation. & \{{ \it cave }\} \\ \midrule 
 1 & Nothing can be seen other than the clear, beautiful sky above and the many surrounding mountains. Very little of this world can be seen or heard from inside the cave. & \{{ \it  }\} \\ \midrule 
 1 & The story presented in the French opera, by Camille Saint-Saens, is of an artist "whose life is dictated by a love for drugs and Japan." & \{{ \it Camille Saint-Saens }\} \\ \midrule 
 1 & As a result, the performers smoke cannabis joints on stage, and the theatre itself is encouraging the audience to join in. & \{{ \it  }\} \\ \midrule 
 1 & Insects were the first animals to take to the air. Their ability to fly helped them evade enemies more easily and find food and mates more efficiently. & \{{ \it air }\} \\ \midrule 
 1 & Most insects have the advantage of being able to fold their wings back along the body. & \{{ \it  }\} \\ \midrule 
 1 & This gives them a wider range of small places to hide from predators. & \{{ \it  }\} \\ \midrule 
 1 & Today, the only insects that cannot fold back their wings are dragon flies and mayflies. & \{{ \it mayflies }\} \\ \midrule 
 1 & California Governor Arnold Schwarzenegger signed into law a bill that bans the sale or rental of violent video games to minors. & \{{ \it Arnold Schwarzenegger, bill }\} \\ \midrule 
 1 & The bill requires violent video games sold in the state of California to be labeled with a decal reading "18" and makes their sale to a minor punishable by a fine of \$1000 per offense. & \{{ \it if }\} \\ \midrule 
 1 & Giancarlo Fisichella lost control of his car and ended the race very soon after the start. & \{{ \it Giancarlo Fisichella }\} \\ \midrule 
 1 & His teammate Fernando Alonso was in the lead for most of the race, but ended it right after his pit-stop, probably because a badly tucked right front wheel. & \{{ \it Fernando Alonso, pit stop }\} \\ \midrule 
 1 & Michael Schumacher ended his race not long after Alonso, because of the suspension damage in the numerous battles during the race. & \{{ \it Michael Schumacher }\} \\ \midrule 
 1 & Taiwan beginning start way back in 15th century where European sailors passing by record the island’s name as Ilha Formosa, or beautiful island. & \{{ \it Ilha Formosa }\} \\ \midrule 
 1 & In 1624,Dutch East India Company establishes a base in southwestern Taiwan, initiating a transformation in aboriginal grain production practices and employing Chinese laborers to work on its rice and sugar plantations. & \{{ \it Dutch East India, grain, rice, plantation }\} \\ \midrule 
 1 & In 1683, Qing dynasty (1644-1912) forces take control of Taiwan’s western and northern coastal areas and declared Taiwan as a province of the Qing Empire in 1885. & \{{ \it Qing dynasty, Qing empire }\} \\ \midrule 
 1 & In 1895, after defeat in the First Sino-Japanese War (1894-1895), the Qing government signs the Treaty of Shimonoseki, by which it cedes sovereignty over Taiwan to Japan, which rules the island until 1945. & \{{ \it Sino, War, Qing, Shimonoseki }\} \\ \midrule 
 1 & Earlier the Chinese news agency Xinhua reported a plane to be hijacked. & \{{ \it Xinhua, news }\} \\ \midrule 
 1 & Later reports then stated the plane received a bomb threat and was diverted back to Afghanistan, landing in Kandahar. & \{{ \it Kandahar }\} \\ \midrule 
 1 & The early reports say the plane was diverted back to Afghanistan after being denied an emergency landing in Ürümqi. & \{{ \it Urumqi }\} \\ \midrule 
 1 & Around 60 cases of malfunctioning iPods overheating have been reported, causing a total of six fires and leaving four people with minor burns. & \{{ \it iPod, burn, minor }\} \\ \midrule 
 1 & Japan's Ministry of Economy, Trade and Industry (METI) said that it had been aware of 27 accidents related to the devices. & \{{ \it METI, iPod }\} \\ \midrule 
 1 & Last week, METI announced that Apple had informed it of 34 additional overheating incidents, which the company called "non-serious." & \{{ \it METI }\} \\ \midrule 
 1 & The ministry responded by calling Apple's postponement of the report "truly regrettable." & \{{ \it  }\} \\ \midrule 
 1 & Reindeer husbandry is an important livelihood among the Sámi and the culture surrounding the trade is important also for many with other professions. & \{{ \it reindeer, Sami }\} \\ \midrule 
 1 & Even traditionally, though, not all Sámi have been involved in big scale reindeer husbandry, but lived from fishing, hunting and similar, having reindeer mostly as draft animals. & \{{ \it Sami }\} \\ \midrule 
 1 & Today many Sámi work in modern trades. Tourism is an important income in Sápmi, the Sámi area. & \{{ \it Sami, Sapmi }\} \\
        \bottomrule
    \end{tabular}
    \label{tab:annotations18}
\end{table*}

\begin{table*}[]
    \centering
    \fontsize{8.4}{10}\selectfont
    \begin{tabular}{cp{10.5cm}p{3.5cm}}
        \toprule
        \textbf{signer} & \textbf{reference} & \textbf{fingerspelled terms} \\
        \midrule
1 & Hong Kong Island gives the territory of Hong Kong its name and is the place that many tourists regard as the main focus. & \{{ \it  }\} \\ \midrule 
 1 & The parade of buildings that make the Hong Kong skyline has been likened to a glittering bar chart that is made apparent by the presence of the waters of Victoria Harbour. & \{{ \it Victoria Harbor, bar chart }\} \\ \midrule 
 1 & To get the best views of Hong Kong, leave the island and head for the Kowloon waterfront opposite. & \{{ \it Kowloon }\} \\ \midrule 
 1 & The great majority of Hong Kong Island's urban development is densely packed on reclaimed land along the northern shore. & \{{ \it shore }\} \\ \midrule 
 1 & This is the place the British colonisers took as their own and so if you are looking for evidence of the territory's colonial past, this is a good place to start. & \{{ \it  }\} \\ \midrule 
 1 & The capital of Moldova is Chişinău. The local language is Romanian, but Russian is widely used. & \{{ \it Moldova, Chisinau }\} \\ \midrule 
 1 & Moldova is a multi-ethnic republic that has suffered from ethnic conflict. & \{{ \it Moldova }\} \\ \midrule 
 1 & In 1994, this conflict led to the creation of the self-proclaimed Transnistria Republic in eastern Moldova, which has its own government and currency but is not recognised by any UN member country. & \{{ \it Transnistria Republic, Moldova, UN }\} \\ \midrule 
 1 & Economic links have been re-established between these two parts of Moldova despite the failure in political negotiations. & \{{ \it  }\} \\ \midrule 
 1 & The major religion in Moldova is Orthodox Christian. & \{{ \it  }\} \\ \midrule 
 1 & Under them are more medium sized cats that eat medium sized prey ranging from rabbits to antelopes and deer. & \{{ \it prey }\} \\ \midrule 
 1 & Finally, there are many small cats (including loose pet cats) that eat the far more numerous small prey like insects, rodents, lizards, and birds. & \{{ \it prey }\} \\ \midrule 
 1 & The secret to their success is the concept of the niche, a special job each cat holds that keeps it from competing with others. & \{{ \it niche }\} \\ \midrule 
 1 & Both cyanuric acid and melamine were found in urine samples from pets that died after consuming contaminated pet food. & \{{ \it cyanuric acid, melamine }\} \\ \midrule 
 1 & The two compounds react with one another to form crystals that may block kidney function, researchers at the university said. & \{{ \it crystals, kidney }\} \\ \midrule 
 1 & The researchers observed crystals formed in cat urine by the addition of melamine and cyanuric acid. & \{{ \it crystals }\} \\ \midrule 
 1 & The composition of these crystals matches those found in the urine of affected pets when compared by infrared spectroscopy (FTIR). & \{{ \it infrared spectroscopy (FTIR) }\} \\ \midrule 
 1 & Content theories are centered on finding what makes people tick or appeals to them. & \{{ \it  }\} \\ \midrule 
 1 & These theories suggest that people have certain needs and/or desires which have been internalized as they mature to adulthood. & \{{ \it  }\} \\ \midrule 
 1 & These theories look at what it is about certain people that make them want the things that they do and what things in their environment will make them do or not do certain things. & \{{ \it  }\} \\ \midrule 
 1 & Two popular content theories are Maslow's Hierarchy of Needs Theory and Hertzberg's Two Factor Theory. & \{{ \it Maslow's, Hertzberg's }\} \\ \midrule 
 1 & Duvall, who is married with two adult children, did not leave a big impression on Miller, to whom the story was related. & \{{ \it Duvall, Miller }\} \\ \midrule 
 1 & When asked for comment, Miller said, "Mike talks a lot during the hearing...I was getting ready so I wasn't really hearing what he was saying." & \{{ \it Miller, Mike }\} \\
 1 & The eathquake struck Mariana at 07:19 a.m. local time (09:19 p.m. GMT Friday). & \{{ \it Mariana, GMT }\} \\ \midrule 
 1 & The Northern Marianas emergency management office said that there were no damages reported in the nation. & \{{ \it Marianas }\} \\ \midrule 
 1 & Also the Pacific Tsunami Warning Center said that there was no Tsunami indication. & \{{ \it Pacific, tsunami }\} \\ \midrule 
 1 & Wagonways were built in England as early as the 16th Century. & \{{ \it wagonways }\} \\ \midrule 
 1 & Although wagonways merely consisted of parallel planks of wood, they allowed horses pulling them to achieve greater speeds and pull larger loads than on the slightly more rough roads of the day. & \{{ \it  }\} \\ \midrule 
 1 & Crossties were introduced fairly early to hold the tracks in place. Gradually, however, it was realised that tracks would be more efficient if they had a stip of iron on the top. & \{{ \it crossties, if iron }\} \\
        \bottomrule
    \end{tabular}
    \label{tab:annotations19}
\end{table*}

\begin{table*}[]
    \centering
    \fontsize{8.4}{10}\selectfont
    \begin{tabular}{cp{10.5cm}p{3.5cm}}
        \toprule
        \textbf{signer} & \textbf{reference} & \textbf{fingerspelled terms} \\
        \midrule
 1 & This became common practice, but the iron caused more wear on the wooden wheels of the wagons. & \{{ \it iron }\} \\ \midrule 
 1 & Eventually, wooden wheels were replaced by iron wheels. In 1767, the first full-iron rails were introduced. & \{{ \it iron, iron }\} \\ \midrule 
 1 & The East African Islands are in the Indian Ocean off the eastern coast of Africa. & \{{ \it  }\} \\ \midrule 
 1 & Madagascar is by far the biggest, and a continent on its own when it comes to wildlife. & \{{ \it Madagascar, wildlife }\} \\ \midrule 
 1 & Most of the smaller islands are independent nations, or associated with France, and known as luxury beach resorts. & \{{ \it nations }\} \\ \midrule 
 1 & The Arabs also brought Islam to the lands, and it took in a big way in the Comoros and Mayotte. & \{{ \it Arabs, Comoros, Mayotte }\} \\ \midrule 
 1 & European influence and colonialism began in the 15th century, as Portuguese explorer Vasco da Gama found the Cape Route from Europe to India. & \{{ \it Vasco da Gama }\} \\ \midrule 
 1 & Although Winfrey was tearful in her farewell, she made it clear to her fans she will be back. & \{{ \it Winfrey }\} \\ \midrule 
 1 & "This is not going to be goodbye. This is the closing of one chapter and the opening of a new one." & \{{ \it  }\} \\ \midrule 
 1 & Adekoya has since been in Edinburgh Sheriff Court charged with murdering her son. & \{{ \it Adekoya, Edinburgh }\} \\ \midrule 
 1 & She is in custody pending indictment and trial, but any eyewitness evidence may be tainted because her image has been widely published. & \{{ \it  }\} \\ \midrule 
 1 & This is common practice elsewhere in the UK but Scottish justice works differently and courts have viewed publication of photos as potentially prejudicial. & \{{ \it UK }\} \\ \midrule 
 1 & Professor Pamela Ferguson of the University of Dundee notes "journalists do seem to be walking a dangerous line if publishing photos etc of suspects." & \{{ \it Pamela Ferguson, Dundee }\} \\ \midrule 
 1 & Crown Office, which is in overall charge of prosecutions, has indicated to journalists that no further comment will be made at least until indictment. & \{{ \it Crown Office }\} \\ \midrule 
 1 & Shooters in the supplementary pest control trial were to be closely supervised by rangers, as the trial was monitored and its effectiveness evaluated. & \{{ \it rangers }\} \\ \midrule 
 1 & In a partnership of NPWS and the Sporting Shooters Association of Australia (NSW) Inc, qualified volunteers were recruited, under the Sporting Shooters Association's hunting program. & \{{ \it NPWS, NSW }\} \\ \midrule 
 1 & According to Mick O'Flynn, the Acting Director Park Conservation and Heritage with the NPWS, the four shooters selected for the first shooting operation received comprehensive safety and training instruction. & \{{ \it Mick O'Flynn, Park }\} \\ \midrule 
 1 & Virtually all computers in use today are based on the manipulation of information which is coded in the form of binary numbers. & \{{ \it code, binary }\} \\ \midrule 
 1 & A binary number can have only one of two values, i.e. 0 or 1, and these numbers are referred to as binary digits - or bits, to use computer jargon. & \{{ \it binary, binary digits, bits }\} \\ \midrule 
 1 & Germany was a common enemy in World War 2, leading to cooperation between the USSR and USA. With the end of the war the clashes of system, process and culture led to the countries falling out. & \{{ \it WW2, USSR, US }\} \\ \midrule 
 1 & With two years of the end of the war, the former allies were now enemies and the Cold War began. & \{{ \it Cold War }\} \\ \midrule 
 1 & It was to last for the next 40 years and would be fought for real, by proxy armies, on battlefields from Africa to Asia, in Afghanistan, Cuba and many other places. & \{{ \it  }\} \\ \midrule 
 1 & Swirl the two dry powders together and then, with clean wet hands, squeeze them into a ball. & \{{ \it  }\} \\ \midrule 
 1 & The moisture on your hands will react with the outer layers, which will feel funny and form a sort of shell. & \{{ \it shell }\} \\ \midrule 
 1 & Historians have criticized past FBI policies for focusing resources on cases which are easy to solve, especially stolen car cases, with the intent of boosting the agency's success rate. & \{{ \it FBI }\} \\ \midrule 
 1 & Congress began funding the obscenity initiative in fiscal 2005 and specified that the FBI must devote 10 agents to adult pornography. & \{{ \it obscenity initiative, FBI, pornography }\} \\ \midrule 
 1 & If crossing the Northern Baltic in winter, check the cabin location, as going through ice causes quite horrible noise for those most affected. & \{{ \it Baltic, ice, cabin }\} \\
        \bottomrule
    \end{tabular}
    \label{tab:annotations20}
\end{table*}

\begin{table*}[]
    \centering
    \fontsize{8.4}{10}\selectfont
    \begin{tabular}{cp{10.5cm}p{3.5cm}}
        \toprule
        \textbf{signer} & \textbf{reference} & \textbf{fingerspelled terms} \\
        \midrule
 1 & Saint Petersburg cruises include time in town. Cruise passengers are exempted from visa requirements (check the terms). & \{{ \it Saint Petersburg, passengers, terms }\} \\ \midrule 
 1 & Anyone who's going to drive at high latitudes or over mountain passes should consider the possibility of snow, ice, or freezing temperatures. & \{{ \it ice }\} \\ \midrule 
 1 & On icy and snowy roadways, friction is low and you cannot drive as if you were on bare asphalt. & \{{ \it  }\} \\ \midrule 
 1 & During blizzards, enough snow to get you stuck can fall in very little time. & \{{ \it  }\} \\ \midrule 
 1 & Visibility may also be restricted by falling or blowing snow or by condensation or ice on vehicle windows. & \{{ \it condensation }\} \\ \midrule 
 1 & On the other hand, icy and snowy conditions are normal in many countries, and traffic goes on mostly uninterrupted all year round. & \{{ \it icy }\} \\ \midrule 
 1 & Dr. Tony Moll discovered the Extremely Drug Resistant Tuberculosis (XDR-TB) in the South African region KwaZulu-Natal. & \{{ \it Dr. Tony Moll, tuberculosis, XDR-TB, KwaZulu-Natal }\} \\ \midrule 
 1 & In an interview, he said the new variant was "very highly troubling and alarming because of the very high fatality rate." & \{{ \it  }\} \\ \midrule 
 1 & Some patients might have contracted the bug in the hospital, Dr. Moll thinks, and at least two were hospital health workers. & \{{ \it  }\} \\ \midrule 
 1 & In one year's time, an infected person may infect 10 to 15 close contacts. & \{{ \it  }\} \\ \midrule 
 1 & However, the percentage of XDR-TB in the entire group of people with tuberculosis still seems to be low; 6,000 of the total 330,000 people infected at any particular moment in South Africa. & \{{ \it XDR-TB, TB }\} \\ \midrule 
 1 & For some festivals, the vast majority of the attendants to music festivals decide to camp on site, and most attendants consider it a vital part of the experience. & \{{ \it  }\} \\ \midrule 
 1 & If you want to be close to the action you're going to have to get in early to get a camping site close to the music. & \{{ \it  }\} \\ \midrule 
 1 & Remember that even though music on the main stages may have finished, there may be sections of the festival that will keep playing music until late into the night. & \{{ \it  }\} \\ \midrule 
 1 & Some festivals have special camping areas for families with young children. & \{{ \it  }\} \\ \midrule 
 1 & Martelly swore in a new Provisional Electoral Council (CEP) of nine members yesterday. & \{{ \it Martelly }\} \\ \midrule 
 1 & It is Martelly's fifth CEP in four years. & \{{ \it  }\} \\ \midrule 
 1 & Last month a presidential commission recommended the prior CEP's resignation as part of a package of measures to move the country towards new elections. & \{{ \it CEP }\} \\ \midrule 
 1 & The commission was Martelly's response to widespread anti-regime protests that started in October. & \{{ \it Martelly }\} \\ \midrule 
 1 & The sometimes-violent protests were triggered by failure to hold elections, some due since 2011. & \{{ \it  }\} \\ \midrule 
 1 & Goats seem to have been first domesticated roughly 10,000 years ago in the Zagros Mountains of Iran. & \{{ \it domesticated, Zagros }\} \\ \midrule 
 1 & Ancient cultures and tribes began to keep them for easy access to milk, hair, meat, and skins. & \{{ \it tribes, hair, skin }\} \\ \midrule 
 1 & Domestic goats were generally kept in herds that wandered on hills or other grazing areas, often tended by goatherds who were frequently children or adolescents, similar to the more widely known shepherd. These methods of herding are still used today. & \{{ \it goatherd, shepherd }\} \\ \midrule 
 1 & Many observed rhythms in physiology and behavior often crucially depend on the presence of endogenous cycles and their production through biological clocks. & \{{ \it physiology, endogenous }\} \\ \midrule 
 1 & Periodic rhythms, which are not simply responses to external periodic cues, have been documented for most living beings, including bacteria, fungi, plants, and animals. & \{{ \it fungi }\} \\ \midrule 
 1 & Biological clocks are self sustaining oscillators which will continue a period of free-running cycling even in the absence of external cues. & \{{ \it  }\} \\ \midrule 
 1 & A couple may decide it is not in their best interest, or in the interest of their child, to raise a baby. & \{{ \it  }\} \\ \midrule 
 1 & These couples may choose to make an adoption plan for their baby. & \{{ \it  }\} \\ \midrule 
 1 & In an adoption, the birth parents terminate their parental rights so that another couple may parent the child. & \{{ \it  }\} \\
        \bottomrule
    \end{tabular}
    \label{tab:annotations21}
\end{table*}

\begin{table*}[]
    \centering
    \fontsize{8.4}{10}\selectfont
    \begin{tabular}{cp{10.5cm}p{3.5cm}}
        \toprule
        \textbf{signer} & \textbf{reference} & \textbf{fingerspelled terms} \\
        \midrule
 1 & The Amazon River is the second longest and the biggest river on Earth. It carries more than 8 times as much water as the second biggest river. & \{{ \it Amazon }\} \\ \midrule 
 1 & The Amazon is also the widest river on Earth, at times six miles wide. & \{{ \it miles }\} \\ \midrule 
 1 & A full 20 percent of the water that pours out of the planet's rivers into the oceans comes from the Amazon. & \{{ \it Amazon }\} \\ \midrule 
 1 & The main Amazon River is 6,387 km (3,980 miles). It collects water from thousands of smaller rivers. & \{{ \it rivers }\} \\ \midrule 
 1 & The obvious way of flying in first or business class is to fork out a thick wad of money for the privilege (or, better yet, get your company to do it for you). & \{{ \it  }\} \\ \midrule 
 1 & However, this does not come cheap: as rough rules of thumb, you can expect to pay up to four times the normal economy fare for business, and eleven times for first class! & \{{ \it  }\} \\ \midrule 
 1 & Generally speaking, there is no point in even looking for discounts for business or first-class seats on direct flights from A to B. & \{{ \it  }\} \\ \midrule 
 1 & Airlines know well that there is a certain core group of flyers who are willing to pay top dollar for the privilege of getting somewhere fast and in comfort, and charge accordingly. & \{{ \it  }\} \\ \midrule 
 1 & The photographer was transported to Ronald Reagan UCLA Medical Center, where he subsequently died. & \{{ \it Ronald Reagan UCLA }\} \\ \midrule 
 1 & He was reportedly aged in his 20s. In a statement, Bieber said "[w]hile I was not present nor directly involved with this tragic accident, my thoughts and prayers are with the family of the victim." & \{{ \it Bieber }\} \\ \midrule 
 1 & Entertainment news website TMZ understands the photographer stopped his vehicle on the other side of Sepulveda Boulevard and attempted to take pictures of the police stop before crossing the road and continuing, prompting the California Highway Patrol police officer conducting the traffic stop to order him back across, twice. & \{{ \it news, TMZ, Sepulveda Boulevard }\} \\ \midrule 
 1 & According to police, the driver of the vehicle that hit the photographer is unlikely to face criminal charges. & \{{ \it  }\} \\ \midrule 
 1 & Prime Minister Stephen Harper has agreed to send the government's 'Clean Air Act' to an all-party committee for review, before its second reading, after Tuesday's 25 minute meeting with NDP leader Jack Layton at the PMO. & \{{ \it Stephen Harper, Clean Air Act, all party, NDP, Jack Layton }\} \\ \midrule 
 1 & Layton had asked for changes to the conservatives' environmental bill during the meeting with the PM, asking for a "thorough and complete rewriting" of the Conservative party's environmental bill. & \{{ \it Layton, act }\} \\ \midrule 
 1 & Ever since the Federal Government stepped in to take over funding of the Mersey hospital in Devonport, Tasmania, the state government and some federal MPs have criticised this act as a stunt in the prelude to the federal election to be called by November. & \{{ \it Mersey, Devonport, Tasmania, MP, November }\} \\ \midrule 
 1 & But Prime Minister John Howard has said the act was only to safeguard the facilities of the hospital from being downgraded by the Tasmanian government, in giving an extra AUD\$45 million. & \{{ \it John Howard, act, AUD }\} \\ \midrule 
 1 & While one experimental vaccine appears able to reduce Ebola mortality, up until now, no drugs have been clearly demonstrated suitable for treating existing infection. & \{{ \it  }\} \\ \midrule 
 1 & One antibody cocktail, ZMapp, initially showed promise in the field, but formal studies indicated it had less benefit than sought in preventing death. & \{{ \it ZMapp, antibody }\} \\ \midrule 
 1 & In the PALM trial, ZMapp served as a control, meaning scientists used it as a baseline and compared the three other treatments to it. & \{{ \it PALM trial, ZMapp, baseline }\} \\ \midrule 
 1 & NextGen is a system the FAA claims would allow aircraft to fly shorter routes and save millions of gallons of fuel each year and cut carbon emissions. & \{{ \it NextGen, FAA, gallons, fuel, CO2 }\} \\ \midrule 
 1 & It uses satellite-based technology as opposed to older ground-radar-based technology to allow air traffic controllers to pinpoint aircraft with greater precision and give pilots more accurate information. & \{{ \it NextGen, satellite, radar }\} \\ \midrule 
 1 & Aerosmith have cancelled their remaining concerts on their tour. & \{{ \it Aerosmith }\} \\ \midrule 
 1 & The rock band was due to tour the United States and Canada until September 16. & \{{ \it September }\} \\ \midrule 
 1 & They have cancelled the tour after lead singer Steven Tyler was injured after he fell off stage while performing on August 5. & \{{ \it Steven Tyler, August }\} \\ \midrule 
 1 & İzmir is the third largest city in Turkey with a population of around 3.7 million, the second biggest port after Istanbul, and a very good transport hub. & \{{ \it Izmir, hub }\} \\ \midrule 
 1 & Once the ancient city of Smyrna, it is now a modern, developed, and busy commercial center, set around a huge bay and surrounded by mountains. & \{{ \it Smyrna, bay }\} \\
        \bottomrule
    \end{tabular}
    \label{tab:annotations22}
\end{table*}

\begin{table*}[]
    \centering
    \fontsize{8.4}{10}\selectfont
    \begin{tabular}{cp{10.5cm}p{3.5cm}}
        \toprule
        \textbf{signer} & \textbf{reference} & \textbf{fingerspelled terms} \\
        \midrule
 1 & The broad boulevards, glass-fronted buildings and modern shopping centers are dotted with traditional red-tiled roofs, the 18th century market, and old mosques and churches, although the city has an atmosphere more of Mediterranean Europe than traditional Turkey. & \{{ \it roof }\} \\ \midrule 
 1 & The Kruger National Park (KNP) lies in the north-east of South Africa and runs along the border of Mozambique in the east, Zimbabwe in the north, and the southern border is the Crocodile River. & \{{ \it Kruger, Park, KNP }\} \\ \midrule 
 1 & The park covers 19,500 km² and is divided in 14 different ecozones, each supporting different wildlife. & \{{ \it km2, ecozones, wildlife }\} \\ \midrule 
 1 & It is one of the main attractions of South Africa and it is considered the flagship of South African National Parks (SANParks). & \{{ \it SANParks, park }\} \\ \midrule 
 1 & As with all South African National Parks, there are daily conservation and entry fees for the park. & \{{ \it park, fee }\} \\ \midrule 
 1 & It may also be beneficial for one to buy a Wild Card, which provides entry to either selections of parks in South Africa or all of the South African National Parks. & \{{ \it Wild Card, SANPark }\} \\ \midrule 
 1 & Many trips to the reef are made all year around, and injuries due to any of these causes on the reef are rare. & \{{ \it reef, rare }\} \\ \midrule 
 1 & Still, take advice from authorities, obey all signs, and pay close attention to safety warnings. & \{{ \it  }\} \\ \midrule 
 1 & Box jellyfish occur near beaches and near river estuaries from October to April north of 1770. They can occasionally be found outside these times. & \{{ \it box jellyfish, estuaries, October, April, October, April }\} \\ \midrule 
 1 & Sharks do exist, however they rarely attack humans. Most sharks are scared of humans and would swim away. & \{{ \it rare }\} \\ \midrule 
 1 & Saltwater Crocodiles do not actively live in the ocean, their primary habitat is in river estuaries north from Rockhampton. & \{{ \it saltwater, estuary, Rockhampton }\} \\ \midrule 
 1 & "We now have 4-month-old mice that are non-diabetic that used to be diabetic," he added. & \{{ \it  }\} \\ \midrule 
 1 & Dr. Ehud Ur, professor of medicine at Dalhousie University in Halifax, Nova Scotia and chair of the clinical and scientific division of the Canadian Diabetes Association cautioned that the research is still in its early days. & \{{ \it Dr. Ehud Ur, Dalhousie }\} \\ \midrule 
 1 & Like some other experts, he is skeptical about whether diabetes can be cured, noting that these findings have no relevance to people who already have Type 1 diabetes. & \{{ \it  }\} \\ \midrule 
 1 & Ironing damp clothes can help them dry. Many hotels have an iron and ironing board available for loan, even if one is not present in the room. & \{{ \it  }\} \\ \midrule 
 1 & If an iron isn't available, or if you don't fancy wearing ironed socks, then you can try using a hairdryer, if available. & \{{ \it  }\} \\ \midrule 
 1 & Be careful not to allow fabric to become too hot (which can cause shrinkage, or in extreme cases, scorch). & \{{ \it scorch }\} \\ \midrule 
 1 & Parisians have a reputation for being egocentric, rude and arrogant. & \{{ \it  }\} \\ \midrule 
 1 & While this is often only an inaccurate stereotype, the best way to get along in Paris still is to be on your best behavior, acting like someone who is "bien élevé" (well brought up). It will make getting about considerably easier. & \{{ \it bien élevé }\} \\ \midrule 
 1 & Parisians' abrupt exteriors will rapidly evaporate if you display some basic courtesies. & \{{ \it  }\} \\ \midrule 
 1 & Naturalists and philosophers focused on classical texts and, in particular, on the Bible in Latin. & \{{ \it text }\} \\ \midrule 
 1 & Accepted were Aristotle's views on all matters of science, including psychology. & \{{ \it Aristotle }\} \\ \midrule 
 1 & As knowledge of Greek declined, the West found itself cut off from its Greek philosophical and scientific roots. & \{{ \it Greek }\} \\ \midrule 
 1 & The "up bow" mark looks like a V and the "down bow mark" like a staple or a square missing its bottom side. & \{{ \it up bow mark, down bow mark }\} \\ \midrule 
 1 & Up means you should start at the tip and push the bow, and down means you should start at the frog (which is where your hand is holding the bow) and pull the bow. & \{{ \it tip, frog }\} \\ \midrule 
 1 & An up-bow usually generates a softer sound, while a down-bow is stronger and more assertive. & \{{ \it up bow, down }\} \\ \midrule 
 1 & Feel free to pencil in your own marks, but remember the printed bowing marks are there for a musical reason, so they should usually be respected. & \{{ \it  }\} \\ \midrule 
 1 & By September 17, 1939, the Polish defense was already broken, and the only hope was to retreat and reorganise along the Romanian bridgehead. & \{{ \it September }\} \\
        \bottomrule
    \end{tabular}
    \label{tab:annotations23}
\end{table*}

\begin{table*}[]
    \centering
    \fontsize{8.4}{10}\selectfont
    \begin{tabular}{cp{10.5cm}p{3.5cm}}
        \toprule
        \textbf{signer} & \textbf{reference} & \textbf{fingerspelled terms} \\
        \midrule
 1 & However, these plans were rendered obsolete nearly overnight, when over 800,000 soldiers from the Soviet's Union Red Army entered and created the Belarussian and Ukrainian fronts after invading the eastern regions of Poland in violation of the Riga Peace Treaty, the Soviet-Polish Non-Aggression Pact, and other international treaties, both bilateral and multilateral. & \{{ \it Riga Peace Treaty, bilateral }\} \\ \midrule 
 1 & Japans holds nearly 7,000 islands (the biggest being Honshu), making Japan the 7th largest island in the world! & \{{ \it Honshu }\} \\ \midrule 
 1 & Due to the cluster/group of islands Japan has, Japan is often referred to, on a geographical stance, as an "archipelago" & \{{ \it archipelago }\} \\ \midrule 
 1 & If you have watched the movie National Treasure, you may think a treasure map was written on the back of the Declaration of Independence. & \{{ \it map }\} \\ \midrule 
 1 & However, that is not true. Although there is something written on the back of the document, it is not a treasure map. & \{{ \it map }\} \\ \midrule 
 1 & Written on the back of the Declaration of Independence were the words "Original Declaration of Independence dated 4th July 1776". The text appears on the bottom of the document, upside down. & \{{ \it July }\} \\ \midrule 
 1 & While no one knows for certain who wrote it, it is known that early in its life, the large parchment document (it measures 29¾ inches by 24½ inches) was rolled up for storage. & \{{ \it  }\} \\ \midrule 
 1 & So, it is likely that the notation was added simply as a label. & \{{ \it  }\} \\ \midrule 
 1 & German pastries are quite good, and in Bavaria, are quite rich and varied, similar to those of their southern neighbor, Austria. & \{{ \it  }\} \\ \midrule 
 1 & Fruit pastries are common, with apples cooked into pastries year round, and cherries and plums making their appearances during the summer. & \{{ \it plum }\} \\ \midrule 
 1 & Many German baked goods also feature almonds, hazelnuts, and other tree nuts. Popular cakes often pair particularly well with a cup of strong coffee. & \{{ \it almond, hazelnut }\} \\ \midrule 
 1 & If you want some small though rich pastries, try what depending on region are called Berliner, Pfannkuchen or Krapfen. & \{{ \it Berliner, Pfannkuchen, Krapfen }\} \\ \midrule 
 1 & The Paralympics will take place from 24 August to 5 September 2021. Some events will be held in other locations throughout Japan. & \{{ \it Paralympics, August, September }\} \\ \midrule 
 1 & Tokyo will be the only Asian city to have hosted two summer Olympics, having hosted the games in 1964. & \{{ \it  }\} \\ \midrule 
 1 & If you booked your flights and accommodation for 2020 before the postponement was announced, you may have a tricky situation. & \{{ \it  }\} \\ \midrule 
 1 & Cancellation policies vary, but as of late March most coronavirus-based cancellation policies don't extend to July 2020, when the Olympics had been scheduled. & \{{ \it March, July }\} \\ \midrule 
 1 & It's expected that most event tickets will cost between ¥2,500 and ¥130,000, with typical tickets costing around ¥7,000. & \{{ \it  }\} \\ \midrule 
 1 & Thousands of years ago, a man called Aristarchus said that the Solar System moved around the Sun. & \{{ \it Aristarchus }\} \\ \midrule 
 1 & Some people thought he was right but many people believed the opposite; that the Solar System moved around the Earth, including the Sun (and even the other stars). & \{{ \it sun }\} \\ \midrule 
 1 & This seems sensible, because the Earth doesn't feel as if it's moving, does it? & \{{ \it  }\} \\ \midrule 
 1 & U.S. President George W. Bush welcomed the announcement. & \{{ \it US, George W. Bush }\} \\ \midrule 
 1 & Bush spokesman Gordon Johndroe called North Korea's pledge "a major step towards the goal of achieving the verifiable denuclearization of the Korean peninsula." & \{{ \it Gordon Johndroe, denuclearization }\} \\ \midrule 
 1 & The use of video recording has led to important discoveries in the interpretation of micro-expressions, facial movements which last a few milliseconds. & \{{ \it  }\} \\ \midrule 
 1 & In particular, it is claimed that one can detect whether a person is lying by interpreting micro-expressions correctly. & \{{ \it  }\} \\ \midrule 
 1 & Oliver Sacks, in his paper The President's Speech, indicated how people who are unable to understand speech because of brain damage are nevertheless able to assess sincerity accurately. & \{{ \it Oliver Sacks }\} \\ \midrule 
 1 & He even suggests that such abilities in interpreting human behavior may be shared by animals such as domestic dogs. & \{{ \it  }\} \\ \midrule 
 1 & The fission bomb works on the principle that it takes energy to put together a nucleus with many protons and neutrons. & \{{ \it fission }\} \\
        \bottomrule
    \end{tabular}
    \label{tab:annotations24}
\end{table*}

\begin{table*}[]
    \centering
    \fontsize{8.4}{10}\selectfont
    \begin{tabular}{cp{10.5cm}p{3.5cm}}
        \toprule
        \textbf{signer} & \textbf{reference} & \textbf{fingerspelled terms} \\
        \midrule
 1 & Sort of like rolling a heavy cart up a hill. Splitting the nucleus up again then releases some of that energy. & \{{ \it cart }\} \\ \midrule 
 1 & Some atoms have unstable nuclei which means that they tend to break apart with little or no nudging. & \{{ \it  }\} \\ \midrule 
 1 & The Internet combines elements of both mass and interpersonal communication. & \{{ \it  }\} \\ \midrule 
 1 & The distinct characteristics of the Internet lead to additional dimensions in terms of the uses and gratifications approach. & \{{ \it  }\} \\ \midrule 
 1 & For example, “learning” and “socialization” are suggested as important motivations for Internet use (James et al., 1995). & \{{ \it James et al }\} \\ \midrule 
 1 & “Personal involvement” and “continuing relationships” were also identified as new motivation aspects by Eighmey and McCord (1998) when they investigated audience reactions to websites. & \{{ \it Eighmey McCord }\} \\ \midrule 
 1 & Robin Uthappa made the innings highest score, 70 runs in just 41 balls by hitting 11 fours and 2 sixes. & \{{ \it Robin Uthappa, inning }\} \\ \midrule 
 1 & Middle order batsmen, Sachin Tendulkar and Rahul Dravid, performed well and made a hundred-run partnership. & \{{ \it Sachin Tendulkar, Rahul Dravid }\} \\ \midrule 
 1 & But, after losing the captain's wicket India only made 36 runs loosing 7 wickets to end the innings. & \{{ \it India, inning }\} \\ \midrule 
 1 & Generally speaking, two behaviors can emerge as managers begin to lead their former peers. One end of the spectrum is trying to remain “one of the guys” (or gals). & \{{ \it  }\} \\ \midrule 
 1 & This type of manager has difficulty making unpopular decisions, performing disciplinary action, performance evaluations, assigning responsibility, and holding people accountable. & \{{ \it  }\} \\ \midrule 
 1 & At the other end of the spectrum, one morphs into an unrecognizable individual that feels he or she must change everything the team has been doing and make it their own. & \{{ \it  }\} \\ \midrule 
 1 & After all, the leader is ultimately responsible for the success and failure of the team. & \{{ \it  }\} \\ \midrule 
 1 & This behavior oftentimes results in rifts between the leaders and the rest of the team. & \{{ \it rift }\} \\ \midrule 
 1 & Chambers had sued God for "widespread death, destruction and terrorization of millions upon millions of the Earth's inhabitants." & \{{ \it Chambers }\} \\ \midrule 
 1 & Chambers, an agnostic, argues that his lawsuit is "frivolous" and "anybody can sue anybody." & \{{ \it Chambers, agnostic }\} \\ \midrule 
 1 & A curry is a dish based on herbs and spices, together with either meat or vegetables. & \{{ \it curry, herb, spice }\} \\ \midrule 
 1 & A curry can be either "dry" or "wet" depending on the amount of liquid. & \{{ \it liquid }\} \\ \midrule 
 1 & In inland regions of Northern India and Pakistan, yogurt is commonly used in curries; in Southern India and some other coastal regions of the subcontinent, coconut milk is commonly used. & \{{ \it yogurt }\} \\ \midrule 
 1 & While project based learning should make learning easier and more interesting, scaffolding goes a step beyond. & \{{ \it  }\} \\ \midrule 
 1 & Scaffolding is not a method of learning but rather an aid that provides support to individuals whom are undergoing a new learning experience such as using a new computer program or beginning a new project. & \{{ \it individual }\} \\ \midrule 
 1 & Scaffolds can be both virtual and real, in other words, a teacher is a form of scaffold but so is the little paperclip man in Microsoft Office. & \{{ \it  }\} \\ \midrule 
 1 & Virtual Scaffolds are internalized in the software and are meant to question, prompt, and explain procedures that may have been to challenging for the student to handle alone. & \{{ \it virtual, software }\} \\ \midrule 
 1 & During his trip, Iwasaki ran into trouble on many occasions. & \{{ \it Iwasaki }\} \\ \midrule 
 1 & He was robbed by pirates, attacked in Tibet by a rabid dog, escaped marriage in Nepal and was arrested in India. & \{{ \it dog, rabid }\} \\ \midrule 
 1 & UN peacekeepers, whom arrived in Haiti after the 2010 earthquake, are being blamed for the spread of the disease which started near the troop's encampment. & \{{ \it UN }\} \\ \midrule 
 1 & According to the lawsuit, waste from the UN camp was not properly sanitized, causing bacteria to enter the tributary of the Artibonite River, one of Haiti's largest. & \{{ \it Artibonite }\} \\ \midrule 
 1 & Prior to the arrival of troops, Haiti had not encountered problems related to the disease since the 1800s. & \{{ \it  }\} \\
        \bottomrule
    \end{tabular}
    \label{tab:annotations25}
\end{table*}

\begin{table*}[]
    \centering
    \fontsize{8.4}{10}\selectfont
    \begin{tabular}{cp{10.5cm}p{3.5cm}}
        \toprule
        \textbf{signer} & \textbf{reference} & \textbf{fingerspelled terms} \\
        \midrule
 1 & The Haitian Institute for Justice and Democracy has referenced independent studies that suggest the Nepalese UN peacekeeping battalion unknowingly brought the disease to Haiti. & \{{ \it UN }\} \\ \midrule 
 1 & Danielle Lantagne, a UN expert on the disease, stated the outbreak was likely caused by the peacekeepers. & \{{ \it Danielle Lantagne }\} \\ \midrule 
 1 & Romanticism had a large element of cultural determinism, drawn from writers such as Goethe, Fichte, and Schlegel. & \{{ \it romanticism, Goethe, Fichte, Schlegel }\} \\ \midrule 
 1 & In the context of Romanticism, the geography molded individuals, and over time customs and culture related to that geography arose, and these, being in harmony with the place of the society, were better than arbitrarily imposed laws. & \{{ \it romanticism }\} \\ \midrule 
 1 & Gothic style peaked in the period between the 10th - 11th centuries and the 14th century. & \{{ \it Gothic style }\} \\ \midrule 
 1 & At the beginning dress was heavily influenced by the Byzantine culture in the east. & \{{ \it Byzantine }\} \\ \midrule 
 1 & However, due to the slow communication channels, styles in the west could lag behind by 25 to 30 year. & \{{ \it style }\} \\ \midrule 
 1 & towards the end of the Middle Ages western Europe began to develop their own style. one of the biggest developments of the time as a result of the crusades people began to use buttons to fasten clothing. & \{{ \it style }\} \\ \midrule 
 1 & Former House Speaker Newt Gingrich, Texas governor Rick Perry, and Congresswoman Michele Bachmann finished in fourth, fifth, and sixth place, respectively. & \{{ \it Newt Gingrich, Rick Perry, Michele Bachmann }\} \\ \midrule 
 1 & After the results came in, Gingrich lauded Santorum, but had tough words for Romney, on whose behalf negative campaign advertisements were aired in Iowa against Gingrich. & \{{ \it Santorum, Romney, Iowa }\} \\ \midrule 
 1 & Perry stated that he would "return to Texas to assess the results of tonight's caucus, determine whether there is a path forward for myself in this race", but later said that he would remain in the race and compete in the January 21 South Carolina primary. & \{{ \it Perry, January, South Carolina, primary }\} \\ \midrule 
 1 & Bachmann, who won the Ames Straw Poll in August, decided to end her campaign. & \{{ \it Bachmann, Ames Straw Poll, August }\} \\ \midrule 
 1 & Cochamó Valley - Chile's premier climbing destination, known as the Yosemite of South America, with a variety of granite big walls and crags. & \{{ \it Cochamo Valley, granite, crags }\} \\ \midrule 
 1 & Summits include breath-taking views from peaks. Climbers from all parts of the world are continually establishing new routes amongst its endless potential of walls. & \{{ \it  }\} \\ \midrule 
 1 & The Babylonians built each of their gods a primary temple that was considered the home of the god. & \{{ \it  }\} \\ \midrule 
 1 & People would bring sacrifices to the gods and the priests would try to attend to the needs of the gods through ceremonies and festivals. & \{{ \it  }\} \\ \midrule 
 1 & Each temple had an open temple courtyard and then an inner sanctuary that only the priests could enter. & \{{ \it  }\} \\ \midrule 
 1 & Sometimes special pyramid shaped towers, called ziggurats, were built to be a part of the temples. & \{{ \it ziggurat }\} \\ \midrule 
 1 & The top of the tower was special sanctuary for the god. & \{{ \it  }\} \\ \midrule 
 1 & "We will endeavour to cut carbon dioxide emissions per unit of GDP by a notable margin by 2020 from the 2005 level," Hu said. & \{{ \it Hu, CO2, GDP }\} \\ \midrule 
 1 & He did not set a figure for the cuts, saying they will be made based on China's economic output. & \{{ \it  }\} \\ \midrule 
 1 & Hu encouraged developing countries "to avoid the old path of polluting first and cleaning up later." & \{{ \it Hu }\} \\ \midrule 
 1 & He added that "they should not, however, be asked to take on obligations that go beyond their development stage, responsibility and capabilities." & \{{ \it  }\} \\ \midrule 
 1 & The main local beer is 'Number One', it is not a complex beer, but pleasant and refreshing. The other local beer is called "Manta". & \{{ \it Manta }\} \\ \midrule 
 1 & There are many French wines to be had, but the New Zealand and Australian wines might travel better. & \{{ \it  }\} \\ \midrule 
 1 & The local tap water is perfectly safe to drink, but bottled water is easy to find if you are fearful. & \{{ \it  }\} \\ \midrule 
 1 & For Australians, the idea of 'flat white' coffee is foreign. A short black is 'espresso', cappuccino comes heaped high with cream (not froth), and tea is served without milk. & \{{ \it flat white, espresso, froth }\} \\ \midrule 
 1 & The hot chocolate is up to Belgian standards. Fruit juices are pricey but excellent. & \{{ \it  }\} \\
        \bottomrule
    \end{tabular}
    \label{tab:annotations26}
\end{table*}

\begin{table*}[]
    \centering
    \fontsize{8.4}{10}\selectfont
    \begin{tabular}{cp{10.5cm}p{3.5cm}}
        \toprule
        \textbf{signer} & \textbf{reference} & \textbf{fingerspelled terms} \\
        \midrule
 1 & The Tibetan Buddhism is based on the teachings of Buddha, but were extended by the mahayana path of love and by a lot of techniques from Indian Yoga. & \{{ \it mahayana path, yoga }\} \\ \midrule 
 1 & In principle the Tibetan Buddhism is very simple. It consists of Kundalini Yoga, meditation and the path of all-embracing love. & \{{ \it kundalini }\} \\ \midrule 
 1 & With Kundalini Yoga the Kundalini energy (enlightenment energy) is awakened through yoga postures, breathing exercises, mantras and visualizations. & \{{ \it kundalini }\} \\ \midrule 
 1 & The center of Tibetan meditation is the Deity Yoga. Through the visualization of various deities the energy channels are cleaned, the chakras are activated and the enlightenment consciousness is created. & \{{ \it  }\} \\ \midrule 
 1 & Virtual teams are held to the same standards of excellence as conventional teams, but there are subtle differences. & \{{ \it  }\} \\ \midrule 
 1 & Virtual team members often function as the point of contact for their immediate physical group. & \{{ \it  }\} \\ \midrule 
 1 & They often have more autonomy than conventional team members as their teams may meet according to varying time zones which may not be understood by their local management. & \{{ \it  }\} \\ \midrule 
 1 & The presence of a true “invisible team” (Larson and LaFasto, 1989, p109) is also a unique component of a virtual team. & \{{ \it Larson LaFasto }\} \\ \midrule 
 1 & The “invisible team” is the management team to which each of the members report. The invisible team sets the standards for each member. & \{{ \it  }\} \\ \midrule 
 1 & The Guaraní were the most significant indigenous group inhabiting what is now Eastern Paraguay, living as semi-nomadic hunters who also practised subsistence agriculture. & \{{ \it Guarani }\} \\ \midrule 
 1 & The Chaco region was home to other groups of indigenous tribes such as the Guaycurú and Payaguá, who survived by hunting, gathering and fishing. & \{{ \it Chaco region, Guaycuru, Payagua }\} \\ \midrule 
 1 & In the 16th century Paraguay, formerly called "The Giant Province of the Indies", was born as a result of the encounter of Spanish conquerors with the native indigenous groups. & \{{ \it Indies }\} \\ \midrule 
 1 & The Spaniards started the colonization period which lasted for three centuries. & \{{ \it  }\} \\ \midrule 
 1 & Since the foundation of Asunción in 1537, Paraguay has managed to keep a lot of its indigenous character and identity. & \{{ \it Asuncion }\} \\ \midrule 
 1 & No tsunami warning has been issued, and according to the Jakarta geophysics agency, no tsunami warning will be issued because the quake did not meet the magnitude 6.5 requirement. & \{{ \it tsunami, Jakarta }\} \\ \midrule 
 1 & Despite there being no tsunami threat, residents started to panic and began to leave their businesses and homes. & \{{ \it  }\} \\ \midrule 
 1 & Twentieth century research has shown that there are two pools of genetic variation: hidden and expressed. & \{{ \it  }\} \\ \midrule 
 1 & Mutation adds new genetic variation, and selection removes it from the pool of expressed variation. & \{{ \it  }\} \\ \midrule 
 1 & Segregation and recombination shuffle variation back and forth between the two pools with each generation. & \{{ \it  }\} \\ \midrule 
 1 & Fluke wrote that the efforts by some to drown out women from speaking out about women’s health were unsuccessful. & \{{ \it Fluke }\} \\ \midrule 
 1 & She came to this conclusion due to the multitude of positive comments and encouragement sent to her by both female and male individuals urging that contraception medication be considered a medical necessity. & \{{ \it contraception }\} \\ \midrule 
 1 & The Hershey and Chase experiment was one of the leading suggestions that DNA was a genetic material. & \{{ \it Hershey, Chase, DNA }\} \\ \midrule 
 1 & Hershey and Chase used phages, or viruses, to implant their own DNA into a bacterium. & \{{ \it phages, DNA }\} \\ \midrule 
 1 & They did two experiments marking either the DNA in the phage with a radioactive phosphorus or the protein of the phage with radioactive sulfur. & \{{ \it DNA, radioactive, sulfur, radioactive }\} \\ \midrule 
 1 & Neurobiological data provide physical evidence for a theoretical approach to the investigation of cognition. Therefore it narrows the research area and makes it much more exact. & \{{ \it data, cognition }\} \\ \midrule 
 1 & The correlation between brain pathology and behaviour supports scientists in their research. & \{{ \it  }\} \\ \midrule 
 1 & It has been known for a long time that different types of brain damage, traumas, lesions, and tumours affect behaviour and cause changes in some mental functions. & \{{ \it  }\} \\
        \bottomrule
    \end{tabular}
    \label{tab:annotations27}
\end{table*}

\begin{table*}[]
    \centering
    \fontsize{8.4}{10}\selectfont
    \begin{tabular}{cp{10.5cm}p{3.5cm}}
        \toprule
        \textbf{signer} & \textbf{reference} & \textbf{fingerspelled terms} \\
        \midrule
 1 & The rise of new technologies allows us to see and investigate brain structures and processes never seen before. & \{{ \it  }\} \\ \midrule 
 1 & This provides us with a lot of information and material to build simulation models which help us to understand processes in our mind. & \{{ \it model }\} \\ \midrule 
 1 & Throughout 1960s, Brzezinski worked for John F. Kennedy as his advisor and then the Lyndon B. Johnson administration. & \{{ \it Brzezinski, JFK, Lyndon B. Johnson }\} \\ \midrule 
 1 & During the 1976 selections he advised Carter on foreign policy, then served as National Security Advisor (NSA) from 1977 to 1981, succeeding Henry Kissinger. & \{{ \it NSA, Henry Kissinger }\} \\ \midrule 
 1 & As NSA, he assisted Carter in diplomatically handling world affairs, such as the Camp David Accords, 1978; normalizing US–China relations thought the late 1970s; the Iranian Revolution, which led to the Iran hostage crisis, 1979; and the Soviet invasion in Afghanistan, 1979. & \{{ \it NSA, Camp David Accords, US }\} \\ \midrule 
 1 & In the north the region is bounded by the Sahel, and in the south and west by the Atlantic Ocean. & \{{ \it Sahel, Atlantic }\} \\ \midrule 
 1 & Women: It is recommended that any women travellers say that they are married, regardless of actual marital status. & \{{ \it  }\} \\ \midrule 
 1 & It is helpful to also wear a ring (just not one that looks too expensive. & \{{ \it  }\} \\ \midrule 
 1 & Women should realize that cultural differences may result in what they would consider harassment and it is not uncommon to be followed, grabbed by the arm, etc. & \{{ \it  }\} \\ \midrule 
 1 & Be firm in turning down men, and don't be afraid to stand your ground (cultural differences or not, it doesn't make it ok!). & \{{ \it  }\} \\ \midrule 
 1 & Plants look their best when in a natural environment, so resist the temptation to remove even "just one" specimen. & \{{ \it  }\} \\ \midrule 
 1 & If visiting a formally arranged garden, collecting "specimens" is also going to get you ejected, without discussion. & \{{ \it  }\} \\ \midrule 
 1 & Cherry blossom viewing, known as hanami, has been a part of Japanese culture since the 8th century. & \{{ \it hanami }\} \\ \midrule 
 1 & The concept came from China where plum blossoms were the flower of choice. & \{{ \it plum }\} \\ \midrule 
 1 & In Japan, the first cherry blossom parties were hosted by the emperor only for himself and other members of the aristocracy around the Imperial Court. & \{{ \it  }\} \\ \midrule 
 1 & The station's web site describes the show as "old school radio theater with a new and outrageous geeky spin!" & \{{ \it  }\} \\ \midrule 
 1 & In its early days, the show was featured solely at the long-running internet radio site TogiNet Radio, a site focused on talk radio. & \{{ \it TogiNet }\} \\ \midrule 
 1 & In late 2015, TogiNet established AstroNet Radio as a subsidiary station. & \{{ \it TogiNet, AstroNet }\} \\ \midrule 
 1 & The show originally featured amateur voice actors, local to East Texas. & \{{ \it  }\} \\ \midrule 
 1 & An increasingly more popular option for those planning a gap-year is to travel and learn. & \{{ \it  }\} \\ \midrule 
 1 & This is especially popular with school leavers, allowing them to take a year out before university, without compromising their education. & \{{ \it  }\} \\ \midrule 
 1 & In many cases, enrolling on a gap-year course abroad can actually improve your chances of moving into higher education back in your home country. & \{{ \it abroad }\} \\ \midrule 
 1 & Typically there will be a tuition fee to enroll in these educational programs. & \{{ \it tuition fee }\} \\ \midrule 
 1 & Persian has a relatively easy and mostly regular grammar. & \{{ \it Persian }\} \\ \midrule 
 1 & Therefore, reading this grammar primer would help you learn much about Persian grammar and understand phrases better. & \{{ \it  }\} \\ \midrule 
 1 & Xinhua reported that government investigators recovered two 'black box' flight recorders on Wednesday. & \{{ \it Xinhua }\} \\ \midrule 
 1 & Fellow wrestlers also paid tribute to Luna. & \{{ \it Luna }\} \\ \midrule 
 1 & Tommy Dreamer said "Luna was the first Queen of Extreme. My first manager. Luna passed away on the night of two moons. Pretty unique just like her. Strong woman." & \{{ \it Tommy Dreamer, Luna }\} \\ \midrule 
 1 & Dustin "Goldust" Runnels commented that "Luna was as freaky as me...maybe even more...love her and will miss her...hopefully she's in a better place." & \{{ \it Dustin Goldust Runnels, Luna }\} \\ \midrule 
 1 & Air accidents are common in Iran, which has an aging fleet that is poorly maintained both for civil and military operations. & \{{ \it  }\} \\
        \bottomrule
    \end{tabular}
    \label{tab:annotations28}
\end{table*}

\begin{table*}[]
    \centering
    \fontsize{8.4}{10}\selectfont
    \begin{tabular}{cp{10.5cm}p{3.5cm}}
        \toprule
        \textbf{signer} & \textbf{reference} & \textbf{fingerspelled terms} \\
        \midrule
 1 & International sanctions have meant that new aircraft cannot be purchased. & \{{ \it  }\} \\ \midrule 
 1 & Earlier this week, a police helicopter crash killed three people and wounded three more. & \{{ \it  }\} \\ \midrule 
 1 & Last month Iran saw its worst air disaster in years when an airliner heading to Armenia crashed, killing the 168 on board. & \{{ \it Armenia }\} \\ \midrule 
 1 & The same month saw another airliner overrun a runway at Mashhad and strike a wall, killing seventeen. & \{{ \it Mashhad, kill }\} \\ \midrule 
 1 & Late on Sunday, the United States President Donald Trump, in a statement delivered via the press secretary, announced US troops would be leaving Syria. & \{{ \it US, Donald Trump, US }\} \\ \midrule 
 1 & The announcement was made after Trump had a phone conversation with Turkish President Recep Tayyip Erdoğan. & \{{ \it Trump, Recep Tayyip Erdogan }\} \\ \midrule 
 1 & Turkey would also take over guarding captured ISIS fighters which, the statement said, European nations have refused to repatriate. & \{{ \it ISIS }\} \\ \midrule 
 1 & Finland is a great boating destination. The "Land of a thousand lakes" has thousands of islands too, in the lakes and in the coastal archipelagos. & \{{ \it archipelago }\} \\ \midrule 
 1 & In the archipelagos and lakes you do not necessarily need a yacht. & \{{ \it yacht }\} \\ \midrule 
 1 & Although the coastal archipelagos and the biggest lakes are indeed big enough for any yacht, smaller boats or even a kayak offer a different experience. & \{{ \it archipelago }\} \\ \midrule 
 1 & Boating is a national pastime in Finland, with a boat to every seven or eight people. & \{{ \it  }\} \\ \midrule 
 1 & This is matched by Norway, Sweden and New Zealand, but otherwise quite unique (e.g. in the Netherlands the figure is one to forty). & \{{ \it Netherlands }\} \\ \midrule 
 1 & The Great Pyramid at Giza is the only one of the seven wonders that is still standing today. & \{{ \it Giza }\} \\ \midrule 
 1 & Built by the Egyptians in the third century BCE, the Great Pyramid is one of many large pyramid structures built to honor dead Pharaoh. & \{{ \it BCE }\} \\ \midrule 
 1 & The Giza Plateau, or "Giza Necropolis" in the Egyptian Valley of the Dead contains several pyramids (of which the great pyramid is the largest), several small tombs, several temples, and the great Sphinx. & \{{ \it Giza Plateau, Giza Necropolis, tomb }\} \\ \midrule 
 1 & The great pyramid was created to honor the Pharaoh Khufu, and many of the smaller pyramids, tombs, and temples were built to honor Khufu's wives and family members. & \{{ \it Khufu, tombs, Khufu }\} \\ \midrule 
 1 & Mosasaurus was the apex predator of its time, so it feared nothing, except other mosasaurs. & \{{ \it Mosasaurus, mosasaurs }\} \\ \midrule 
 1 & Its long jaws were studded with more than 70 razor-sharp teeth, along with an extra set in the roof of its mouth, meaning that there was no escape for anything that crossed its path. & \{{ \it  }\} \\ \midrule 
 1 & We don't know for sure, but it may have had a forked tongue. Its diet included turtles, large fish, other mosasaurs, and it may even have been a cannibal. & \{{ \it mosasaurs, cannibal }\} \\ \midrule 
 1 & It also attacked anything that entered the water; even a giant dinosaur such as T. rex would be no match for it. & \{{ \it T Rex }\} \\ \midrule 
 1 & Stirrups are supports for the rider's feet that hang down on either side of the saddle. & \{{ \it stirrups }\} \\ \midrule 
 1 & They provide greater stability for the rider but can have safety concerns due to the potential for a rider's feet to get stuck in them. & \{{ \it  }\} \\ \midrule 
 1 & If a rider is thrown from a horse but has a foot caught in the stirrup, they could be dragged if the horse runs away. To minimize this risk, a number of safety precautions can be taken. & \{{ \it risk }\} \\ \midrule 
 1 & First, most riders wear riding boots with a heel and a smooth, quite narrow, sole. & \{{ \it  }\} \\ \midrule 
 1 & Next, some saddles, particularly English saddles, have safety bars that allow a stirrup leather to fall off the saddle if pulled backwards by a falling rider. & \{{ \it  }\} \\ \midrule 
 1 & Across the United States of America, there are approximately 400,000 known cases of Multiple Sclerosis (MS), leaving it as the leading neurological disease in younger and middle aged adults. & \{{ \it US, MS, Multiple Sclerosis }\} \\ \midrule 
 1 & MS is a disease that affects the central nervous system, which is made up of the brain, the spinal cord and the optic nerve. & \{{ \it MS, CNS }\} \\ \midrule 
 1 & Research has found that females are two times more likely to have MS then males. & \{{ \it MS }\} \\ \midrule 
 1 & To understand the Templars one must understand the context that prompted the creation of the order. & \{{ \it templars }\} \\ \midrule 
 1 & The age where the events took place is commonly referred as the High Middle Ages the period of European history in the 11th, 12th, and 13th centuries (AD 1000–1300). & \{{ \it AD }\} \\
        \bottomrule
    \end{tabular}
    \label{tab:annotations29}
\end{table*}

\begin{table*}[]
    \centering
    \fontsize{8.4}{10}\selectfont
    \begin{tabular}{cp{10.5cm}p{3.5cm}}
        \toprule
        \textbf{signer} & \textbf{reference} & \textbf{fingerspelled terms} \\
        \midrule
 1 & The High Middle Ages were preceded by the Early Middle Ages and followed by the Late Middle Ages, which by convention ends around 1500. & \{{ \it ages, ages, late, ages }\} \\ \midrule 
 1 & Courier companies are well paid for delivering things quickly. Frequently, time is very important with business documents, merchandise or spare parts for an urgent repair. & \{{ \it courier }\} \\ \midrule 
 1 & On some routes, the larger companies have their own planes, but for other routes and smaller firms there was a problem. & \{{ \it  }\} \\ \midrule 
 1 & If they sent things by air freight, on some routes it may have taken days to get through unloading and customs. & \{{ \it customs }\} \\ \midrule 
 1 & The only way to get it through faster was to send it as checked luggage. Airline regulations will not allow them to send luggage without a passenger, which is where you come in. & \{{ \it  }\} \\ \midrule 
 1 & When the fighting ceased after the wounded were transported to the hospital, about 40 of the other remaining inmates stayed in the yard and refused to return to their cells. & \{{ \it  }\} \\ \midrule 
 1 & Negotiators tried to rectify the situation, but the prisoners' demands are not clear. & \{{ \it  }\} \\ \midrule 
 1 & Between 10:00-11:00 pm MDT, a fire was started by the inmates in the yard. & \{{ \it MDT }\} \\ \midrule 
 1 & Soon, officers equipped with riot gear entered the yard and cornered the inmates with tear gas. & \{{ \it riot gear, gas }\} \\ \midrule 
 1 & Fire rescue crews eventually doused the fire by 11:35 pm. & \{{ \it by }\} \\ \midrule 
 1 & The document, according to the leak, will refer to the borders dispute, which Palestine wants based on the borders before the 1967 Mideast War. & \{{ \it Mideast }\} \\ \midrule 
 1 & Other topics covered reportedly include the future state of Jerusalem which is sacred to both nations and the Jordan Valley issue. & \{{ \it Jordan Valley }\} \\ \midrule 
 1 & Israel demands an ongoing military presence in the valley for ten years once an agreement is signed while the PA agrees to leave such presence only for five years. & \{{ \it PA }\} \\ \midrule 
 1 & "She’s very cute and sings quite well, too," he said according to a transcript of the news conference. & \{{ \it news }\} \\ \midrule 
 1 & "I was moved every time we did a rehearsal on this, from the bottom of my heart." & \{{ \it  }\} \\ \midrule 
 1 & Majorcan cuisine, like that of similar zones in the Mediterranean, is based on bread, vegetables and meat (specially pork), and uses olive oil throughout. & \{{ \it Majorcan, pork, olive oil }\} \\ \midrule 
 1 & A simple popular dinner, especially during the summer, is the Pa amb Oli: Bread with olive oil, tomato, and any available condiments such as cheese, tunafish, etc. & \{{ \it pa amb oli, olive oil, tunafish }\} \\ \midrule 
 1 & If you visit the Arctic or Antarctic areas in the winter you will experience the polar night, which means that the sun doesn't rise above the horizon. & \{{ \it Arctic, sun }\} \\ \midrule 
 1 & This offers a good opportunity to see the Aurora borealis, as the sky will be dark more or less around the clock. & \{{ \it Aurora borealis }\} \\ \midrule 
 1 & As the areas are sparsely populated, and light pollution therefore often not a problem, you will also be able to enjoy the stars. & \{{ \it  }\} \\ \midrule 
 1 & Anyone planning a visit to a country that could be considered a war zone should get professional training. & \{{ \it zone }\} \\ \midrule 
 1 & A search of the Internet for 'Hostile environment course' will probably provide the address of a local company. & \{{ \it  }\} \\ \midrule 
 1 & A course will normally cover all the issues discussed here in far greater detail, usually with practical experience. & \{{ \it  }\} \\ \midrule 
 1 & A course will normally be from 2-5 days and will involve role play, a lot of first aid and sometimes weapons training. & \{{ \it aid }\} \\ \midrule 
 1 & Books and magazines dealing with wilderness survival are common, but publications dealing with war zones are few. & \{{ \it  }\} \\ \midrule
 2 & During the Revolutionary War, the thirteen states first formed a weak central government—with the Congress being its only component—under the Articles of Confederation. & \{{ \it Confederation }\} \\ \midrule 
 2 & Congress lacked any power to impose taxes, and, because there was no national executive or judiciary, it relied on state authorities, who were often uncooperative, to enforce all its acts. & \{{ \it tax, executive }\} \\ \midrule 
 2 & It also had no authority to override tax laws and tariffs between states. & \{{ \it tax, tariff }\} \\
        \bottomrule
    \end{tabular}
    \label{tab:annotations30}
\end{table*}

\begin{table*}[]
    \centering
    \fontsize{8.4}{10}\selectfont
    \begin{tabular}{cp{10.5cm}p{3.5cm}}
        \toprule
        \textbf{signer} & \textbf{reference} & \textbf{fingerspelled terms} \\
        \midrule
 2 & The Articles required unanimous consent from all the states before they could be amended and states took the central government so lightly that their representatives were often absent. & \{{ \it  }\} \\ \midrule 
 2 & Argentina is well known for having one of the best polo teams and players in the world. & \{{ \it well, polo }\} \\ \midrule 
 2 & The largest tournament of the year takes place in December at the polo fields in Las Cañitas. & \{{ \it of, Las Cañitas, Dec }\} \\ \midrule 
 2 & Smaller tournaments and matches can also be seen here at other times of the year. & \{{ \it  }\} \\ \midrule 
 2 & For news on tournaments and where to buy tickets for polo matches, check Asociacion Argentina de Polo. & \{{ \it de }\} \\ \midrule 
 2 & Although pyramid-building in stone continued until the end of the Old Kingdom, the pyramids of Giza were never surpassed in their size and the technical excellence of their construction. & \{{ \it Giza }\} \\ \midrule 
 2 & New Kingdom ancient Egyptians marvelled at their predecessors monuments, which were then well over a thousand year old. & \{{ \it  }\} \\ \midrule 
 2 & Although three people were inside the house when the car impacted it, none of them were hurt. & \{{ \it  }\} \\ \midrule 
 2 & However, the driver sustained serious injuries to the head. & \{{ \it  }\} \\ \midrule 
 2 & The road where the crash happened was temporarily closed while emergency services freed the driver from the red Audi TT. & \{{ \it Audi TT }\} \\ \midrule 
 2 & He was initially hospitalised in the James Paget Hospital in Great Yarmouth. & \{{ \it James Paget, Yarmouth }\} \\ \midrule 
 2 & He was subsequently relocated to Addenbrooke's Hospital in Cambridge. & \{{ \it Addenbrooke's }\} \\ \midrule 
 2 & Ellsworth Land is the region south of the Peninsula, bounded by the Bellingshausen Sea. & \{{ \it peninsula, Ellsworth, Bellingshausen Sea }\} \\ \midrule 
 2 & The mountains of the Peninsula here merge into the plateau, then re-emerge to form the 360 km chain of the Ellsworth Mountains, bisected by the Minnesota Glacier. & \{{ \it glacier, Minn Glacier, Ellsworth, km }\} \\ \midrule 
 2 & The northern part or Sentinel Range has Antarctica's highest mountains, the Vinson Massif, peaking at 4892 m Mount Vinson. & \{{ \it Sentinel Range, Antarctica, Vinson Massif, Mt Vinson, meter }\} \\ \midrule 
 2 & Everyone participates in society and uses transportation systems. Almost everyone complains about transportation systems. & \{{ \it  }\} \\ \midrule 
 2 & In developed countries you seldom hear similar levels of complaints about water quality or bridges falling down. & \{{ \it  }\} \\ \midrule 
 2 & Why do transportation systems engender such complaints, why do they fail on a daily basis? Are transportation engineers just incompetent? Or is something more fundamental going on? & \{{ \it  }\} \\ \midrule 
 2 & The Iraq Study Group presented its report at 12.00 GMT today. & \{{ \it Iraq }\} \\ \midrule 
 2 & It warns No one can guarantee that any course of action in Iraq at this point will stop sectarian warfare, growing violence, or a slide toward chaos. & \{{ \it chaos }\} \\ \midrule 
 2 & The Report opens with plea for open debate and the formation of a consensus in the United States about the policy towards the Middle East. & \{{ \it US }\} \\ \midrule 
 2 & The Report is highly critical of almost every aspect of the present policy of the Executive towards Iraq and it urges an immediate change of direction. & \{{ \it  }\} \\ \midrule 
 2 & First among its 78 recommendations is that a new diplomatic initiative should be taken before the end of this year to secure Iraq’s borders against hostile interventions and to re-establish diplomatic relations with its neighbors. & \{{ \it  }\} \\ \midrule 
 2 & Internal poisoning may not be immediately apparent. Symptoms, such as vomiting are sufficiently general that an immediate diagnosis cannot be made. & \{{ \it  }\} \\ \midrule 
 2 & The best indication of internal poisoning may be the presence of an open container of medication or toxic household chemicals. & \{{ \it  }\} \\ \midrule 
 2 & Check the label for specific first aid instructions for that specific poison. & \{{ \it aid }\} \\ \midrule 
 2 & The fatalities were Nicholas Alden, 25, and Zachary Cuddeback, 21. Cuddeback had been the driver. & \{{ \it Nicholas Alden,  Zachary Cuddeback }\} \\ \midrule 
 2 & Edgar Veguilla received arm and jaw wounds while Kristoffer Schneider was left requiring reconstructive surgery for his face. & \{{ \it Edgar Veguilla, Kristoffer Schneider }\} \\
        \bottomrule
    \end{tabular}
    \label{tab:annotations31}
\end{table*}

\begin{table*}[]
    \centering
    \fontsize{8.4}{10}\selectfont
    \begin{tabular}{cp{10.5cm}p{3.5cm}}
        \toprule
        \textbf{signer} & \textbf{reference} & \textbf{fingerspelled terms} \\
        \midrule
 2 & Uka's weapon failed whilst pointed at a fifth man's head. Schneider has ongoing pain, blindness in one eye, a missing section of skull and a face rebuilt from titanium. & \{{ \it titanium, Uka, weapon }\} \\ \midrule 
 2 & Schneider testified via videolink from a USAF base in his homeland. & \{{ \it Schneider, US }\} \\ \midrule 
 2 & The U.S. Corps of Engineers estimated that 6 inches of rainfall could breach the previously damaged levees. & \{{ \it US Corps of, inch }\} \\ \midrule 
 2 & The Ninth Ward, which saw flooding as high as 20 feet during Hurricane Katrina, is currently in waist-high water as the nearby levee was overtopped. & \{{ \it Ward, feet, Katrina }\} \\ \midrule 
 2 & Water is spilling over the levee in a section 100 feet wide. & \{{ \it feet }\} \\ \midrule 
 2 & A former Filipino policeman has kept Hong Kong tourists hostage by hijacking their bus in Manila, the capital of the Philippines. & \{{ \it Manila, hostage }\} \\ \midrule 
 2 & Rolando Mendoza fired his M16 rifle at the tourists. & \{{ \it Rolando Mendoza, M16 }\} \\ \midrule 
 2 & Several hostages have been rescued and least six have been confirmed dead so far. & \{{ \it  }\} \\ \midrule 
 2 & Six hostages, including the children and elderly, were released early, as were the Filipino photographers. & \{{ \it  }\} \\ \midrule 
 2 & The photographers later took the place of an aged lady as she needed the lavatory. Mendoza was gunned down. & \{{ \it Mendoza }\} \\ \midrule 
 2 & While most of their food would be familiar to us, Romans did have their share of strange or unusual feast items, including wild boar, peacock, snails, and a type of rodent called a dormouse & \{{ \it dormouse }\} \\ \midrule 
 2 & Another difference was that while the poor people and the woman ate their food while sitting in chairs, the rich men liked to have banquets together where they would lounge on their sides while they ate their meals. & \{{ \it  }\} \\ \midrule 
 2 & Ancient Roman meals couldn't have included foods that came to Europe from America or from Asia in later centuries. & \{{ \it  }\} \\ \midrule 
 2 & For instance, they didn't have corn, nor tomatoes, nor potatoes, nor cocoa, and no ancient Roman ever tasted a turkey. & \{{ \it  }\} \\ \midrule 
 2 & Goma is a tourist city of the Democratic Republic of Congo in the extreme east near Rwanda. & \{{ \it Congo, Goma, Rwanda }\} \\ \midrule 
 2 & In 2002 Goma was destroyed by lava from the Nyiragongo volcano which buried most of the town’s streets, particularly the town centre. & \{{ \it Nyiragongo }\} \\ \midrule 
 2 & While Goma is reasonably safe, any visits outside of Goma should be researched to understand the state of the fighting that persists in the North Kivu province. & \{{ \it Goma, Kivu province }\} \\ \midrule 
 2 & The city is also the base to climb the Nyiragongo volcano along with some of the cheapest Mountain Gorilla tracking in Africa. & \{{ \it Goma, Nyiragongo, of }\} \\ \midrule 
 2 & You can use boda-boda (motorcycle taxi) to get around Goma. The normal (local) price is \textasciitilde{}500 Congolese Francs for the short ride. & \{{ \it taxi, boda-boda, Congolese Francs }\} \\ \midrule 
 2 & Voyagers planning sex reassignment surgery abroad must ensure they're carrying valid documents for the return trip. & \{{ \it sex }\} \\ \midrule 
 2 & The willingness of governments to issue passports with gender not stated (X) or documents updated to match a desired name and gender varies. & \{{ \it gender, gender }\} \\ \midrule 
 2 & Willingness of foreign governments to honour these documents is just as widely variable. & \{{ \it  }\} \\ \midrule 
 2 & Searches at security checkpoints have also become far more intrusive in the post-September 11, 2001 era. & \{{ \it Sept, at }\} \\ \midrule 
 2 & Pre-operative transgender people should not expect to pass through the scanners with their privacy and dignity intact. & \{{ \it  }\} \\ \midrule 
 2 & Ocelots like to eat small animals. They will catch monkeys, snakes, rodents and birds if they can. Almost all of the animals that the ocelot hunts are far smaller than it is. & \{{ \it ocelot, ocelot }\} \\ \midrule 
 2 & Scientists think that ocelots follow and find animals to eat (prey) by smell, sniffing for where they've been on the ground. & \{{ \it ocelots, ocelot }\} \\ \midrule 
 2 & They can see very well in the dark with night vision, and move very stealthily, too. Ocelots hunt their prey by blending in with their surroundings then pouncing on their prey. & \{{ \it ocelot }\} \\ \midrule 
 2 & In remote locations, without cell phone coverage, a satellite phone may be your only option. & \{{ \it  }\} \\ \midrule 
 2 & A satellite phone is not generally a replacement for a mobile phone, as you have to be outdoors with clear line of sight to the satellite to make a phone call. & \{{ \it  }\} \\
        \bottomrule
    \end{tabular}
    \label{tab:annotations32}
\end{table*}

\begin{table*}[]
    \centering
    \fontsize{8.4}{10}\selectfont
    \begin{tabular}{cp{10.5cm}p{3.5cm}}
        \toprule
        \textbf{signer} & \textbf{reference} & \textbf{fingerspelled terms} \\
        \midrule
 2 & The service is frequently used by shipping, including pleasure craft, as well as expeditions who have remote data and voice needs. & \{{ \it  }\} \\ \midrule 
 2 & Your local telephone service provider should be able to give more information about connecting to this service. & \{{ \it  }\} \\ \midrule 
 2 & The strain of bird flu lethal to humans, H5N1, has been confirmed to have infected a dead wild duck, found on Monday, in marshland near Lyon in the east of France. & \{{ \it flu, strain, H5N1, Lyon, marsh, H5N1 }\} \\ \midrule 
 2 & France is the seventh country in the European Union to suffer this virus; following Austria, Germany, Slovenia, Bulgaria, Greece and Italy. & \{{ \it EU }\} \\ \midrule 
 2 & Suspected cases of H5N1 in Croatia and Denmark remain unconfirmed. & \{{ \it H5N1 }\} \\ \midrule 
 2 & During the 1920s, the prevailing attitudes of most citizens and nations was that of pacifism and isolation. & \{{ \it  }\} \\ \midrule 
 2 & After seeing the horrors and atrocities of war during World War I, nations desired to avoid such a situation again in the future. & \{{ \it WW1 }\} \\ \midrule 
 2 & Children are placed in Foster Care for a wide variety of reasons that range from neglect, to abuse, and even to extortion. & \{{ \it foster, even }\} \\ \midrule 
 2 & No child should ever have to grow up in an environment that is not nurturing, caring, and educational, but they do. & \{{ \it  }\} \\ \midrule 
 2 & We perceive the Foster Care System to be a safety zone for these children. & \{{ \it at }\} \\ \midrule 
 2 & Our foster care system is supposed to provide safe homes, loving caregivers, stable education, and reliable health care. & \{{ \it  }\} \\ \midrule 
 2 & Foster care is supposed to provide all the necessities that were lacking in the home they were previously taken from. & \{{ \it  }\} \\ \midrule 
 2 & With 17,000 islands to choose from, Indonesian food is an umbrella term covering a vast variety of regional cuisines found across the nation. & \{{ \it  }\} \\ \midrule 
 2 & But, if used without further qualifiers, the term tends to mean the food originally from the central and eastern parts of the main island Java. & \{{ \it Java }\} \\ \midrule 
 2 & Now widely available throughout the archipelago, Javanese cuisine features an array of simply seasoned dishes, the predominant flavorings the Javanese favor being peanuts, chillies, sugar (especially Javanese coconut sugar) and various aromatic spices. & \{{ \it Java, Java, Java, spices }\} \\ \midrule 
 2 & The cities of Harappa and Mohenjo-daro had a flush toilet in almost every house, attached to a sophisticated sewage system. & \{{ \it Harappa, Mohenjo-daro }\} \\ \midrule 
 2 & Remains of sewage systems have been found in the houses of the Minoan cities of Crete and Santorini in Greece. & \{{ \it Minoan, Crete, Santorini }\} \\ \midrule 
 2 & There were also toilets in ancient Egypt, Persia and China. In Roman civilization, toilets were sometimes part of public bath houses where men and women were together in mixed company. & \{{ \it Persia }\} \\ \midrule 
 2 & Pre-modern observatories are usually obsolete today, and remain as museums, or sites of education. & \{{ \it or }\} \\ \midrule 
 2 & As light pollution in their heyday was not the kind of problem it is today, they are usually located in cities or at campuses, easier to reach than those built in modern times. & \{{ \it  }\} \\ \midrule 
 2 & Most modern research telescopes are enormous facilities in remote areas with favorable atmospheric conditions. & \{{ \it  }\} \\ \midrule 
 2 & The program started at 8:30 p.m. local time (15.00 UTC). & \{{ \it UTC }\} \\ \midrule 
 2 & Famous singers across the country presented bhajans, or devotional songs, to Shri Shyam's feet. & \{{ \it bhajans, Shri Shyam's, feet }\} \\ \midrule 
 2 & Singer Sanju Sharma started the evening, followed by Jai Shankar Choudhary. esented the chhappan bhog bhajan as well. Singer, Raju Khandelwal was accompanying him. & \{{ \it Sanju Sharma, Jai Shankar Choudhary, chhappan bhog bhajan, Raju Khandelwal }\} \\ \midrule 
 2 & Then, Lakkha Singh took the lead in singing the bhajans. & \{{ \it Lakkha Singh, bhajans }\} \\ \midrule 
 2 & 108 plates of Chhappan Bhog (in Hinduism, 56 different edible items, like, sweets, fruits, nuts, dishes etc. which are offered to deity) were served to Baba Shyam. & \{{ \it Baba Shyam, chhappan bhog, Hindu, deity }\} \\ \midrule 
 2 & Lakkha Singh presented the chhappan bhog bhajan as well. Singer, Raju Khandelwal was accompanying him. & \{{ \it Lakkha Singh, Raju Khandelwal, chhappan bhog bhajan }\} \\ \midrule 
 2 & At Thursday's keynote presentation of the Tokyo Game Show, Nintendo president Satoru Iwata unveiled the controller design for the company's new Nintendo Revolution console. & \{{ \it Nintendo, Satoru Iwata, Nintendo, console }\} \\
        \bottomrule
    \end{tabular}
    \label{tab:annotations33}
\end{table*}

\begin{table*}[]
    \centering
    \fontsize{8.4}{10}\selectfont
    \begin{tabular}{cp{10.5cm}p{3.5cm}}
        \toprule
        \textbf{signer} & \textbf{reference} & \textbf{fingerspelled terms} \\
        \midrule
 2 & Resembling a television remote, the controller uses two sensors placed near the user's television to triangulate its position in three-dimensional space. & \{{ \it TV, TV, sensor, 3D }\} \\ \midrule 
 2 & This will allow players to control actions and movements in video games by moving the device through the air. & \{{ \it  }\} \\ \midrule 
 2 & The war expenditures of U.S. imperialism in the conquest of the Philippines were paid for by the Filipino people themselves. & \{{ \it US, US }\} \\ \midrule 
 2 & They were compelled to pay taxes to the U.S. colonial regime to defray a major part of the expenditures and the interest on bonds floated in the name of the Philippine government through the Wall Street banking houses. & \{{ \it tax, US, bond, Wall Street banks }\} \\ \midrule 
 2 & Of course, the superprofits derived from the protracted exploitation of the Filipino people would constitute the basic gains of U.S. imperialism. & \{{ \it US, US }\} \\ \midrule 
 2 & The medical charity Mangola, Medecines Sans Frontieres and the World Health Organisation say it is the worst outbreak recorded in the country. & \{{ \it Mangola, Medecines Sans Frontiers, WHO }\} \\ \midrule 
 2 & Spokesman for Medecines Sans Frontiere Richard Veerman said: "Angola is heading for its worst ever outbreak and the situation remains very bad in Angola," he said. & \{{ \it Medecines Sans Frontieres, Richard Veerman, Angola }\} \\ \midrule 
 2 & The official Falklands currency is the Falkland pound (FKP) whose value is set equivalent to that of one British pound (GBP). & \{{ \it Falkland, Falkland pound, FKP, pound, GBP }\} \\ \midrule 
 2 & Money can be exchanged at the only bank in the islands which is located in Stanley across from the FIC West store. & \{{ \it bank, Stanley, FIC West, bank }\} \\ \midrule 
 2 & British pounds will generally be accepted anywhere in the islands and within Stanley credit cards and United States dollars are also often accepted. & \{{ \it Stanley, US }\} \\ \midrule 
 2 & On the outlying islands credit cards will probably not be accepted, although British and United States currency may be taken; check with the owners in advance to determine what is an acceptable payment method. & \{{ \it US }\} \\ \midrule 
 2 & It is nearly impossible to exchange Falklands currency outside of the islands, so exchange money prior to leaving the islands. & \{{ \it Falkland, Falkland }\} \\ \midrule 
 2 & Hamilton confirmed Howard University Hospital admitted the patient in stable condition. & \{{ \it Hamilton, Howard }\} \\ \midrule 
 2 & The patient had been to Nigeria, where some cases of the Ebola virus have occurred. & \{{ \it Ebola }\} \\ \midrule 
 2 & The hospital has followed protocol for infection control, including separating the patient from others to prevent possible infection of others. & \{{ \it  }\} \\ \midrule 
 2 & Cuomo, 53, began his governorship earlier this year and signed a bill last month legalizing same-sex marriage. & \{{ \it Cuomo, bill, sex }\} \\ \midrule 
 2 & He referred to the rumors as "political chatter and silliness". & \{{ \it  }\} \\ \midrule 
 2 & He is speculated to make a run for president in 2016. & \{{ \it  }\} \\ \midrule 
 2 & Alloys are basically a mixture of two or more metals. Don't forget that there are many elements on the periodic table. & \{{ \it alloy, of, alloys, elements, periodic }\} \\ \midrule 
 2 & Elements like calcium and potassium are considered metals. Of course, there are also metals like silver and gold. & \{{ \it elements, calcium, potassium }\} \\ \midrule 
 2 & You can also have alloys that include small amounts of non-metallic elements like carbon. & \{{ \it alloy, carbon }\} \\ \midrule 
 2 & Mutations can have a variety of different effects depending on the type of mutation, the significance of the piece of genetic material affected and whether the cells affected are germ-line cells. & \{{ \it mutations, gene, of, gene, cells, germ-line, sperm, cells }\} \\ \midrule 
 2 & Only mutations in germ-line cells can be passed on to children, while mutations elsewhere can cause cell-death or cancer. & \{{ \it cell, or }\} \\ \midrule 
 2 & Casinos typically make many efforts to maximize time and money spent by guests. Windows and clocks are usually absent, and exits can be hard to find. & \{{ \it  }\} \\ \midrule 
 2 & They usually have special food, drink and entertainment offers, to keep guests in a good mood, and keep them at the premise. & \{{ \it  }\} \\ \midrule 
 2 & Some venues offer alcoholic beverages on the house. However, drunkenness impairs judgement, and all good gamblers know the importance of staying sober. & \{{ \it  }\} \\ \midrule 
 2 & There's a tradition to pass the Easter night awake at some exposed point to see the sunrise. & \{{ \it  }\} \\ \midrule 
 2 & There are of course Christian theological explanations for this tradition, but it may well be a pre-Christian Spring and Fertility ritual. & \{{ \it ritual }\} \\
        \bottomrule
    \end{tabular}
    \label{tab:annotations34}
\end{table*}

\begin{table*}[]
    \centering
    \fontsize{8.4}{10}\selectfont
    \begin{tabular}{cp{10.5cm}p{3.5cm}}
        \toprule
        \textbf{signer} & \textbf{reference} & \textbf{fingerspelled terms} \\
        \midrule
 2 & More traditional churches often hold an Easter Vigil on Saturday night during the Easter weekend, with the congregations often breaking into celebration at the stroke of midnight to celebrate Christ's resurrection. & \{{ \it vigil }\} \\ \midrule 
 2 & U.S. President George W. Bush arrived in Singapore the morning of November 16, beginning a week-long tour of Asia. & \{{ \it Nov, US, George W. Bush }\} \\ \midrule 
 2 & He was greeted by Singapore's Deputy Prime Minister Wong Kan Seng and discussed trade and terrorism issues with the Singapore Prime Minister Lee Hsien Loong. & \{{ \it PM, Wong Kan Seng, PM, Lee Hsien Loong, terrorism }\} \\ \midrule 
 2 & After a week of losses in the midterm election, Bush told an audience about the expansion of trade in Asia. & \{{ \it term, Bush }\} \\ \midrule 
 2 & The village of Haldarsvík offer views of the nearby island Eysturoy and has an unusual octagonal church. & \{{ \it Haldarsvík, Eysturoy }\} \\ \midrule 
 2 & In the churchyard, there are interesting marble sculptures of doves over some tombs. & \{{ \it dove, marble }\} \\ \midrule 
 2 & It's worth half an hour to stroll about the intriguing village. & \{{ \it  }\} \\ \midrule 
 2 & USA Gymnastics supports the United States Olympic Committee's letter and accepts the absolute need of the Olympic family to promote a safe environment for all of our athletes. & \{{ \it US, US }\} \\ \midrule 
 2 & We agree with the USOC's statement that the interests of our athletes and clubs, and their sport, may be better served by moving forward with meaningful change within our organization, rather than decertification. & \{{ \it USOC, clubs }\} \\ \midrule 
 2 & USA Gymnastics supports an independent investigation that may shine light on how abuse of the proportion described so courageously by the survivors of Larry Nassar could have gone undetected for so long and embraces any necessary and appropriate changes. & \{{ \it US, Larry Nassar, USOC, US }\} \\ \midrule 
 2 & USA Gymnastics and the USOC have the same goal — making the sport of gymnastics, and others, as safe as possible for athletes to follow their dreams in a safe, positive and empowered environment. & \{{ \it USOC, US, so }\} \\ \midrule 
 2 & Most of the distinct Baltic Cruises feature an extended stay in St. Petersburg, Russia. & \{{ \it Baltic, Baltic }\} \\ \midrule 
 2 & This means you can visit the historic city for a couple of full days while returning and sleeping on the ship at night. & \{{ \it  }\} \\ \midrule 
 2 & If you only go ashore using shipboard excursions you will not need a separate visa (as of 2009). & \{{ \it visa, visa }\} \\ \midrule 
 2 & Some cruises feature Berlin, Germany in the brochures. As you can see from the map above Berlin is no where near the sea and a visit to the city is not included in the price of the cruise. & \{{ \it map }\} \\ \midrule 
 2 & They found the Sun operated on the same basic principles as other stars: The activity of all stars in the system was found to be driven by their luminosity, their rotation, and nothing else. & \{{ \it  }\} \\ \midrule 
 2 & The luminosity and rotation are used together to determine a star's Rossby number, which is related to plasma flow. & \{{ \it Rossby, plasma flow }\} \\ \midrule 
 2 & The smaller the Rossby number, the less active the star with respect to magnetic reversals. & \{{ \it Rossby, magnetic }\} \\ \midrule 
 2 & Though many of the animals in the park are used to seeing humans, the wildlife is nonetheless wild and should not be fed or disturbed. & \{{ \it park, wildlife, wild }\} \\ \midrule 
 2 & According to park authorities, stay at least 100 yards/meters away from bears and wolves and 25 yards/meters from all other wild animals! & \{{ \it park, yards, meters, yards, meters }\} \\ \midrule 
 2 & No matter how docile they may look, bison, elk, moose, bears, and nearly all large animals can attack. & \{{ \it  }\} \\ \midrule 
 2 & Each year, dozens of visitors are injured because they didn't keep a proper distance. These animals are large, wild, and potentially dangerous, so give them their space. & \{{ \it  }\} \\ \midrule 
 2 & In addition, be aware that odors attract bears and other wildlife, so avoid carrying or cooking odorous foods and keep a clean camp. & \{{ \it  }\} \\ \midrule 
 2 & Why would an organization want to go through the time consuming process of establishing a learning organization? One goal for putting organizational learning concepts into practice is innovation. & \{{ \it  }\} \\ \midrule 
 2 & When all available resources are effectively used across the functional departments of an organization, creativity and ingenuity can transpire. & \{{ \it  }\} \\ \midrule 
 2 & As a result, the process of an organization working together to overcome an obstacle can lead to a new innovative process to serve the customer's need. & \{{ \it  }\} \\
        \bottomrule
    \end{tabular}
    \label{tab:annotations35}
\end{table*}

\begin{table*}[]
    \centering
    \fontsize{8.4}{10}\selectfont
    \begin{tabular}{cp{10.5cm}p{3.5cm}}
        \toprule
        \textbf{signer} & \textbf{reference} & \textbf{fingerspelled terms} \\
        \midrule
 2 & Before an organization can be innovative, leadership must create a culture of innovation as well as shared knowledge and organizational learning. & \{{ \it  }\} \\ \midrule 
 2 & Angel (2006), explains the Continuum approach as a method being used to help organizations reach a higher level of performance. & \{{ \it Angel }\} \\ \midrule 
 2 & Severe weather is the generic term for any dangerous weather phenomenon with the potential to cause damage, serious social disruption, or loss of human life. & \{{ \it  }\} \\ \midrule 
 2 & Severe weather can occur anywhere in the world, and there are different types of it, which can depend on geography, topography, and atmospheric conditions. & \{{ \it  }\} \\ \midrule 
 2 & High winds, hail, excessive precipitation, and wildfires are forms and effects of severe weather, as are thunderstorms, tornadoes, waterspouts, and cyclones. & \{{ \it  }\} \\ \midrule 
 2 & Regional and seasonal severe weather phenomena include blizzards, snowstorms, ice storms, and dust storms. & \{{ \it  }\} \\ \midrule 
 2 & Travellers are strongly advised to be aware of any risk of severe weather affecting their area as they may affect any travel plans. & \{{ \it  }\} \\ \midrule 
 2 & The terrified King Louis XVI, Queen Marie Antoinette their two young children (11 year old Marie Therese and four year old Louis-Charles) and the King's sister, Madam Elizabeth, on the 6th October 1789 were forced back to Paris from Versailles by a mob of market women. & \{{ \it Oct, Versailles, Louis XVI, Marie Antoinette, Marie Therese, Louis-Charles, Madam Elizabeth }\} \\ \midrule 
 2 & In a carriage, they traveled back to Paris surrounded by a mob of people screaming and shouting threats against the King and Queen. & \{{ \it  }\} \\ \midrule 
 2 & The mob of people forced the King And Queen to have their carriage windows wide open. & \{{ \it  }\} \\ \midrule 
 2 & At one point a member of the mob waved the head of a royal guard killed at Versailles in front of the terrified Queen. & \{{ \it royal, Versailles }\} \\ \midrule 
 2 & In France, voting has traditionally been a low-tech experience: voters isolate themselves in a booth, put a pre-printed sheet of paper indicating their candidate of choice into an envelope. & \{{ \it  }\} \\ \midrule 
 2 & After officials verify the voter's identity, the voter drops the envelope into the ballot box and signs the voting roll. & \{{ \it  }\} \\ \midrule 
 2 & French electoral law rather strictly codifies the proceedings. & \{{ \it  }\} \\ \midrule 
 2 & Since 1988, ballot boxes must be transparent so that voters and observers can witness that no envelopes are present at the start of the vote and that no envelopes are added except those of the duly counted and authorized voters. & \{{ \it  }\} \\ \midrule 
 2 & Candidates can send representatives to witness every part of the process. In the evening, votes are counted by volunteers under heavy supervision, following specific procedures. & \{{ \it  }\} \\ \midrule 
 2 & Final results from Namibian presidential and parliamentary elections have indicated that the incumbent president, Hifikepunye Pohamba, has been reelected by a large margin. & \{{ \it Hifikepunye Pohamba }\} \\ \midrule 
 2 & The ruling party, South West Africa People's Organisation (SWAPO), also retained a majority in the parliamentary elections. & \{{ \it SWAPO }\} \\ \midrule 
 2 & After the accident occurred, Gibson was transported to a hospital but died shortly afterwards. & \{{ \it Gibson }\} \\ \midrule 
 2 & The truck driver, who is aged 64, was not injured in the crash. & \{{ \it truck }\} \\ \midrule 
 2 & The vehicle itself was taken away from the scene of the accident at approximately 1200 GMT on the same day. & \{{ \it GMT }\} \\ \midrule 
 2 & A person working in a garage near where the accident occurred said: "There were children waiting to cross the road and they were all screaming and crying." & \{{ \it  }\} \\ \midrule 
 2 & They all ran back from where the accident had happened. & \{{ \it  }\} \\ \midrule 
 2 & Out of 1,400 people polled prior to the 2010 federal election, those who oppose Australia becoming a republic grew by 8 per cent since 2008. & \{{ \it  }\} \\ \midrule 
 2 & Caretaker Prime Minister Julia Gillard claimed during the campaign of the 2010 federal election that she believed Australia should become a republic at the end of Queen Elizabeth II's reign. & \{{ \it care, PM, Julia Gillard, Elizabeth II }\} \\ \midrule 
 2 & 34 per cent of those in the poll share this view, wanting Queen Elizabeth II to be Australia's last monarch. & \{{ \it Elizabeth II }\} \\ \midrule 
 2 & At the extremes of the poll, 29 per cent of those surveyed believe Australia should become a republic as soon as possible, while 31 per cent believe Australia should never become a republic. & \{{ \it  }\} \\
        \bottomrule
    \end{tabular}
    \label{tab:annotations36}
\end{table*}

\begin{table*}[]
    \centering
    \fontsize{8.4}{10}\selectfont
    \begin{tabular}{cp{10.5cm}p{3.5cm}}
        \toprule
        \textbf{signer} & \textbf{reference} & \textbf{fingerspelled terms} \\
        \midrule
 2 & Ancient China had a unique way of showing different time periods; each stage of China or each family that was in power was a distinctive dynasty. & \{{ \it  }\} \\ \midrule 
 2 & Also between each dynasty was an unstable age of divided provinces. The best-known of these periods was the Three Kingdoms epoch taking place for 60 years between the Han and the Jin Dynasty. & \{{ \it Han, Jin }\} \\ \midrule 
 2 & During these periods fierce warfare took place between many nobles fighting for the throne. & \{{ \it  }\} \\ \midrule 
 2 & The Three Kingdoms was one of the bloodiest eras in Ancient China’s history thousands of people died fighting to sit in the highest seat in the grand palace at Xi’an. & \{{ \it Xian, grand palace }\} \\ \midrule 
 2 & Pronunciation is relatively easy in Italian since most words are pronounced exactly how they are written & \{{ \it  }\} \\ \midrule 
 2 & The main letters to watch out for are c and g, since their pronunciation varies based on the following vowel. & \{{ \it  }\} \\ \midrule 
 2 & Also, make sure to pronounce r and rr differently: caro means dear, whereas carro means chariot. & \{{ \it caro, carro, caro, carro }\} \\ \midrule 
 2 & Television reports show white smoke coming from the plant. & \{{ \it TV, plant }\} \\ \midrule 
 2 & Local authorities are warning residents in the vicinity of the plant to stay indoors, turn off air-conditioners and not to drink tap water. & \{{ \it AC, off }\} \\ \midrule 
 2 & According to Japan's nuclear agency, radioactive caesium and iodine has been identified at the plant. & \{{ \it nuclear, radioactive, caesium, iodine, radioactive }\} \\ \midrule 
 2 & Authorities speculate that this indicates that containers holding uranium fuel at the site may have ruptured and are leaking. & \{{ \it uranium fuel }\} \\ \midrule 
 2 & Using ships to transport goods is by far the most efficient way to move large amounts of people and goods across oceans. & \{{ \it  }\} \\ \midrule 
 2 & The job of navies has traditionally been to ensure that your country maintains the ability to move your people and goods, while at the same time, interfering with your enemy's ability to move his people and goods. & \{{ \it navy }\} \\ \midrule 
 2 & One of the most noteworthy recent examples of this was the North Atlantic campaign of WWII. The Americans were trying to move men and materials across the Atlantic Ocean to help Britain. & \{{ \it WW2, Atlantic, Atlantic }\} \\ \midrule 
 2 & At the same time, the German navy, using mainly U-boats, was trying to stop this traffic. & \{{ \it U-boat }\} \\ \midrule 
 2 & Had the Allies failed, Germany probably would have been able to conquer Britain as it had the rest of Europe. & \{{ \it  }\} \\ \midrule 
 2 & The term bug is used by entomologists in a formal sense for this group of insects. & \{{ \it bug }\} \\ \midrule 
 2 & This term derives from ancient familiarity with Bed-bugs, which are insects highly adapted to parasitize humans. & \{{ \it bug, insect, parasite }\} \\ \midrule 
 2 & Both Assassin-bugs and Bed-bugs are nidicolous, adapted to living in nest or housing of their host. & \{{ \it Assassin bug, nest }\} \\ \midrule 
 2 & A doctor who worked at Children's Hospital of Pittsburgh, Pennsylvania will be charged with aggravated murder after her mother was found dead in the trunk of her car Wednesday, authorities in Ohio say. & \{{ \it PA }\} \\ \midrule 
 2 & Dr. Malar Balasubramanian, 29, was found in Blue Ash, Ohio, a suburb approximately 15 miles north of Cincinnati lying on the ground beside the road in a T-shirt and underwear in an apparently heavily medicated state. & \{{ \it Cincinnati, miles, Ash, Malar Balasubramanian }\} \\ \midrule 
 2 & She directed officers to her black Oldsmobile Intrigue which was 500 feet away. & \{{ \it feet, Oldsmobile Intrigue }\} \\ \midrule 
 2 & There, they found the body of Saroja Balasubramanian, 53, covered with blood-stained blankets. & \{{ \it Saroja Balasubramanian }\} \\ \midrule 
 2 & Police said that the body appeared to have been there for about a day. & \{{ \it  }\} \\ \midrule 
 2 & One bomb exploded outside the governor general's office. & \{{ \it  }\} \\ \midrule 
 2 & Three more bombs exploded near government buildings in a period of two hours. & \{{ \it  }\} \\ \midrule 
 2 & Some reports put the official death toll at eight, and official reports confirm that up to 30 were injured; but final numbers are not yet known. & \{{ \it  }\} \\ \midrule 
 2 & Everything in the Universe is made of matter. All matter is made of tiny particles called atoms. & \{{ \it of, matter, of, atoms }\} \\
        \bottomrule
    \end{tabular}
    \label{tab:annotations37}
\end{table*}

\begin{table*}[]
    \centering
    \fontsize{8.4}{10}\selectfont
    \begin{tabular}{cp{10.5cm}p{3.5cm}}
        \toprule
        \textbf{signer} & \textbf{reference} & \textbf{fingerspelled terms} \\
        \midrule
 2 & Atoms are so incredibly tiny that trillions of them could fit into the period at the end of this sentence. & \{{ \it trillions, of }\} \\ \midrule 
 2 & Before The Simpsons Simon had worked on several shows in various positions. & \{{ \it Simon, TV, The Simpsons }\} \\ \midrule 
 2 & During the 1980s he worked on shows such as Taxi, Cheers, and The Tracy Ullman Show. & \{{ \it Taxi, Cheers, Tracy Ullman Show }\} \\ \midrule 
 2 & In 1989 he helped create The Simpsons with Brooks and Groening, and was responsible for hiring the show's first writing team. & \{{ \it Simon, Brooks, Groening, The Simpsons, Simon }\} \\ \midrule 
 2 & Despite leaving the show in 1993 he kept the title of executive producer, and continued to receive tens of millions of dollars every season in royalties. & \{{ \it Simon }\} \\ \midrule 
 2 & During this period of European history, the Catholic Church, which had become rich and powerful, came under scrutiny. & \{{ \it  }\} \\ \midrule 
 2 & For over a thousand years the Christian religion had bound European states together despite differences in language and customs. I & \{{ \it  }\} \\ \midrule 
 2 & Its all-pervading power affected everyone from king to commoner. & \{{ \it  }\} \\ \midrule 
 2 & One of the main Christian tenets is that wealth should be used to alleviate suffering and poverty and that the monetary funds of the church are there specifically for that reason. & \{{ \it  }\} \\ \midrule 
 2 & The central authority of the church had been in Rome for over a thousand years and this concentration of power and money led many to question whether this tenet was being met. & \{{ \it Rome }\} \\ \midrule 
 2 & All nouns, alongside the word Sie for you, always begin with a capital letter, even in the middle of a sentence. & \{{ \it noun, Sie, noun, noun }\} \\ \midrule 
 2 & This is an important way to distinguish between some verbs and objects. & \{{ \it verb, object }\} \\ \midrule 
 2 & It also arguably makes reading easier, though writing is somewhat complicated by the need to find out whether a verb or adjective is used in a substantivized form. & \{{ \it verb, adjective, noun }\} \\ \midrule 
 2 & To the north and within easy reach is the romantic and fascinating town of Sintra and which was made famous to foreigners after a glowing account of its splendours recorded by Lord Byron. & \{{ \it Sintra, Byron }\} \\ \midrule 
 2 & Scotturb Bus 403 travels regularly to Sintra, stopping at Cabo da Roca. & \{{ \it bus, Scotturb, Cabo de Roca }\} \\ \midrule 
 2 & Also to the north visit the great Sanctuary of Our Lady of Fatima (Shrine), a place of worldwide famous Marian apparitions. & \{{ \it Sanctuary, of, of, Fatima, shrine }\} \\ \midrule 
 2 & Five minutes into the display a wind starts rolling in, about a minute later, the wind is reaching 70km/h... then the rain comes, but so hard and so large that it slaps your skin like a needle, then hail fell from the sky, people panicking and screaming and running over each other. & \{{ \it km, ice }\} \\ \midrule 
 2 & I lost my sister and her friend, and on my way there were two disabled people in wheelchairs, people just jumping over and pushing them," Armand Versace said. & \{{ \it Armand Versace }\} \\ \midrule 
 2 & Gold may be worked into all sorts of shapes. It can be rolled into tiny shapes. & \{{ \it  }\} \\ \midrule 
 2 & It can be pulled into thin wire, which can be twisted and plaited. It can be hammered or rolled into sheets. & \{{ \it  }\} \\ \midrule 
 2 & It can be made very thin, and stuck onto other metal. It can be made so thin that it was sometimes used to decorate the hand-painted pictures in books called "illuminated manuscripts". & \{{ \it  }\} \\ \midrule 
 2 & We can start living more friendly to the environment, we can join to the environmental movement, and we can even be activists in order to reduce the future suffering in some degree. & \{{ \it  }\} \\ \midrule 
 2 & This is just like symptomatic treatment in many cases. However, if we do not only want a temporary solution, then we should find the root of the problems, and we should deactivate them. & \{{ \it root cause }\} \\ \midrule 
 2 & It is obvious enough that the world has changed much because of humankind's scientific and technological advancements, and problems have become greater because of overpopulation and mankind's extravagant lifestyle. & \{{ \it style }\} \\ \midrule 
 2 & A study published on Thursday in the journal Science reported on formation of a new bird species on the Ecuadorean Galápagos Islands. & \{{ \it Galapagos }\} \\ \midrule 
 2 & Researchers from Princeton University in the United States and Uppsala University in Sweden reported the new species evolved in just two generations, though this process had been believed to take much longer, due to breeding between an endemic Darwin finch, Geospiza fortes, and the immigrant cactus finch, Geospiza conirostris. & \{{ \it Princeton, US, Uppsala, Darwin finch, Geospiza fortes, cactus finch, Geospiza conirostris }\} \\
        \bottomrule
    \end{tabular}
    \label{tab:annotations38}
\end{table*}

\begin{table*}[]
    \centering
    \fontsize{8.4}{10}\selectfont
    \begin{tabular}{cp{10.5cm}p{3.5cm}}
        \toprule
        \textbf{signer} & \textbf{reference} & \textbf{fingerspelled terms} \\
        \midrule
 2 & Technological determinism is a term that encompasses a wide range of ideas in practice, from technology-push or the technological imperative to a strict sense that human destiny is driven by an underlying logic associated with scientific laws and their manifestation in technology. & \{{ \it determinism, ism, destiny, by, manifestation }\} \\ \midrule 
 2 & Most interpretations of technological determinism share two general ideas: that the development of technology itself follows a path largely beyond cultural or political influence, and that technology in turn has "effects" on societies that are inherent, rather than socially conditioned. & \{{ \it or, or, to, or, of }\} \\ \midrule 
 2 & For example, one might say that the motor car necessarily leads to the development of roads. & \{{ \it  }\} \\ \midrule 
 2 & However, a nationwide road network is not economically viable for just a handful of cars, so new methods of production are developed to reduce the cost of car ownership. & \{{ \it so }\} \\ \midrule 
 2 & Mass car ownership also leads to a higher incidence of accidents on the roads, which leads to the invention of new techniques in healthcare for repairing damaged bodies. & \{{ \it  }\} \\ \midrule 
 2 & Plants make oxygen which humans breathe, and they take in carbon-dioxide which humans exhale (that is, breathe out). & \{{ \it O2, CO2, CO2 }\} \\ \midrule 
 2 & Plants make their food from the sun by photosynthesis. They also provide shade. & \{{ \it photosynthesis, shade }\} \\ \midrule 
 2 & We make our houses from plants and make clothes from plants. Most foods that we eat are plants. Without plants, animals could not survive. & \{{ \it  }\} \\ \midrule 
 2 & The surface of the Moon is made of rocks and dust. The outer layer of the Moon is called the crust. & \{{ \it surface, crust }\} \\ \midrule 
 2 & The crust is about 70 km thick on the near side and 100 km thick on the far side. & \{{ \it km, km }\} \\ \midrule 
 2 & It is thinner under the maria and thicker under the highlands. & \{{ \it maria, land }\} \\ \midrule 
 2 & There may be more maria on the near side because the crust is thinner. It was easier for lava to rise up to the surface. & \{{ \it maria, crust, lava }\} \\ \midrule 
 2 & Travellers may encounter animal pests that they are not familiar with in their home regions. & \{{ \it pest }\} \\ \midrule 
 2 & Pests can spoil food, cause irritation, or in a worse case cause allergic reactions, spread venom, or transmit infections. & \{{ \it pest }\} \\ \midrule 
 2 & Infectious diseases themselves, or dangerous animals that can injure or kill people by force, do not usually qualify as pests. & \{{ \it or, pest }\} \\ \midrule 
 2 & The wheel has changed the world in incredible ways. The biggest thing that the wheel has done for us is given us much easier and faster transportation. & \{{ \it wheel }\} \\ \midrule 
 2 & It has brought us the train, the car, and many other transportation devices. & \{{ \it  }\} \\ \midrule 
 2 & A civilization is a singular culture shared by a significant large group of people who live and work co-operatively, a society. & \{{ \it  }\} \\ \midrule 
 2 & The word civilization comes from the Latin civilis, meaning civil, related to the Latin civis, meaning citizen, and civitas, meaning city or city-state, and that also somehow defines the size of the society. & \{{ \it civilis, civil, civis, citizen, civitas }\} \\ \midrule 
 2 & City-states are the precursors of nations. A civilizational culture implies the passing on of knowledge across several generations, a lingering cultural footprint and fair dissemination. & \{{ \it  }\} \\ \midrule 
 2 & Minor cultures often vanish without leaving relevant historic evidence and fail to be recognized as proper civilizations. & \{{ \it  }\} \\ \midrule 
 2 & Field trips are a large part of any classroom. Quite often a teacher would love to take her students places to which a bus trip is not an option. & \{{ \it  }\} \\ \midrule 
 2 & Technology offers the solution with virtual field trips. Students can look at museum artifacts, visit an aquarium, or admire beautiful art while sitting with their class. & \{{ \it aquarium }\} \\ \midrule 
 2 & Sharing a field trip virtually is also a great way to reflect a on a trip and share experiences with future classes. & \{{ \it  }\} \\ \midrule 
 2 & For example, each year students from Bennet School in North Carolina design a website about their trip to the State Capital, each year the website gets remodeled, but old versions are kept online to serve as a scrapbook. & \{{ \it NC, Bennet, website, trip, website }\} \\ \midrule 
 2 & While visiting the monasteries, women are required to wear skirts covering the knees and have their shoulders covered, too. & \{{ \it monasteries, skirts, knee }\} \\
        \bottomrule
    \end{tabular}
    \label{tab:annotations39}
\end{table*}

\begin{table*}[]
    \centering
    \fontsize{8.4}{10}\selectfont
    \begin{tabular}{cp{10.5cm}p{3.5cm}}
        \toprule
        \textbf{signer} & \textbf{reference} & \textbf{fingerspelled terms} \\
        \midrule
 2 & Most of the monasteries do provide wraps for women who come unprepared, but if you bring your own, especially one with bright colors, you'll get a smile from the monk or nun at the entrance. & \{{ \it wrap, own, monk, nun }\} \\ \midrule 
 2 & Along the same line, men are required to wear trousers covering the knees. & \{{ \it knee }\} \\ \midrule 
 2 & This too can be borrowed from the stock at the entrance but that clothing isn't washed after every user so you may not feel comfortable wearing these skirts. One size fits all for men! & \{{ \it  }\} \\ \midrule
3 & "It's going to behave like water. It's transparent just the way water is. & \{{ \it  }\} \\ \midrule 
 3 & So if you were standing by the shoreline, you would be able to see down to whatever pebbles or gunk that was on the bottom. & \{{ \it pebbles, gunk }\} \\ \midrule 
 3 & As far as we know, there is only one planetary body that displays more dynamism than Titan, and its name is Earth," added Stofan. & \{{ \it Stofan, Titan, Earth }\} \\ \midrule 
 3 & England had experienced a long period of peace after the reconquest of the Danelaw. & \{{ \it Danelaw }\} \\ \midrule 
 3 & However in 991 Ethelred was faced with a Viking fleet larger than any since Guthrum's a century earlier. & \{{ \it Ethelred, Guthrum }\} \\ \midrule 
 3 & This fleet was led by Olaf Trygvasson, a Norwegian with ambitions to reclaim his country from Danish domination. & \{{ \it Olaf Trygvasson, Danish }\} \\ \midrule 
 3 & After initial military setbacks, Ethelred was able to agree to terms with Olaf, who returned to Norway to try to gain his kingdom with mixed success. & \{{ \it Ethelred, Olaf }\} \\ \midrule 
 3 & Peter Costello, Australian treasurer and the man most likely to succeed Prime Minister John Howard as Liberal party leader has thrown his support behind a nuclear power industry in Australia. & \{{ \it Peter Costello, PM John Howard, nuclear }\} \\ \midrule 
 3 & Mr Costello said that when nuclear power generation becomes economically viable, Australia should pursue its use. & \{{ \it Mr. Costello, nuclear }\} \\ \midrule 
 3 & "If it becomes commercial, we should have it. That is, there's no in-principle objection to nuclear energy" Mr Costello said. & \{{ \it Mr. Costello, nuclear }\} \\ \midrule 
 3 & When the capsule gets to Earth and enters the atmosphere, at about 5am (eastern time), it is expected to put on quite a light show for folks in Northern California, Oregon, Nevada, and Utah. & \{{ \it capsule, atmosphere, Nev, Utah }\} \\ \midrule 
 3 & The capsule will look much like a shooting star going across the sky. & \{{ \it  }\} \\ \midrule 
 3 & The capsule will be traveling at about 12.8 km or 8 miles per second, fast enough to go from San Francisco to Los Angeles in one minute. & \{{ \it km, mps, SF, LA }\} \\ \midrule 
 3 & Stardust will set a new all-time record for being the fastest spacecraft to return to Earth, breaking the previous record set in May of 1969 during the return of the Apollo X command module. & \{{ \it Stardust, May, Apollo X }\} \\ \midrule 
 3 & “It will move over the west coast of northern California and will light the sky from California through central Oregon and on through Nevada and Idaho and into Utah,” Tom Duxbury, Stardust's project manager said. & \{{ \it Tom Duxbury, Nev, Idaho, Utah }\} \\ \midrule 
 3 & Mr. Rudd's decision to sign the Kyoto climate accord isolates the United States, which will now be the only developed nation not to ratify the agreement. & \{{ \it Mr. Rudd, Kyoto climate accord, US }\} \\ \midrule 
 3 & Australia's former conservative government refused to ratify Kyoto, saying it would damage the economy with its heavy reliance on coal exports, while countries like India and China were not bound by emissions targets. & \{{ \it Kyoto, coal exports, emissions }\} \\ \midrule 
 3 & Travel agencies have been around since the 19th century. A travel agent is usually a good option for a journey which extends beyond a traveller's previous experience of nature, culture, language or low-income countries. & \{{ \it agent, agent }\} \\ \midrule 
 3 & Although most agencies are willing to take on most regular bookings, many agents specialise in particular types of travel, budget ranges or destinations. & \{{ \it agents }\} \\ \midrule 
 3 & It can be better to use an agent who frequently books similar trips to yours. & \{{ \it agent }\} \\ \midrule 
 3 & Take a look at what trips the agent is promoting, whether on a website or in a shop window. & \{{ \it agent }\} \\ \midrule 
 3 & The Twin Otter had been trying to land at Kokoda yesterday as Airlines PNG Flight CG4684, but had aborted once already. & \{{ \it Twin Otter, Kokoda, Airlines PNG, CG4684 }\} \\ \midrule 
 3 & About ten minutes before it was due to land from its second approach it disapeared. & \{{ \it  }\} \\ \midrule 
 3 & The crash site was located today and is so inaccessable that two policemen were dropped into the jungle in order to hike to the scene and seek survivors. & \{{ \it  }\} \\
        \bottomrule
    \end{tabular}
    \label{tab:annotations40}
\end{table*}

\begin{table*}[]
    \centering
    \fontsize{8.4}{10}\selectfont
    \begin{tabular}{cp{10.5cm}p{3.5cm}}
        \toprule
        \textbf{signer} & \textbf{reference} & \textbf{fingerspelled terms} \\
        \midrule
 3 & The search had been hampered by the same bad weather that had caused the aborted landing. & \{{ \it did }\} \\ \midrule 
 3 & Born in Hong Kong, Ma studied at New York University and Harvard Law School and once held an American permanent resident "green card". & \{{ \it Ma, NYU }\} \\ \midrule 
 3 & Hsieh implied during the election that Ma might flee the country during a time of crisis. & \{{ \it Hsieh, Ma }\} \\ \midrule 
 3 & Hsieh also argued that the photogenic Ma was more style than substance. & \{{ \it Hsieh, Ma }\} \\ \midrule 
 3 & Despite these accusations, Ma won handily on a platform advocating closer ties with the Chinese mainland. & \{{ \it Ma, mainland }\} \\ \midrule 
 3 & A tornado is a spinning column of very low-pressure air, which sucks the surrounding air inward and upward. & \{{ \it air }\} \\ \midrule 
 3 & They generate high winds (often 100-200 miles/hour) and can lift heavy objects into the air, carrying them as the tornado moves. & \{{ \it mph }\} \\ \midrule 
 3 & They begin as funnels descending from storm clouds, and become "tornadoes" when they touch the ground. & \{{ \it  }\} \\ \midrule 
 3 & Proper nutritional practices alone cannot generate elite performances, but they can significantly affect young athletes' overall wellness. & \{{ \it  }\} \\ \midrule 
 3 & Maintaining a healthy energy balance, practicing effective hydration habits, and understanding the various aspects of supplementation practices can help athletes improve their performance and increase their enjoyment of the sport. & \{{ \it supplementation }\} \\ \midrule 
 3 & A hospitality exchange network is the organization that connects travelers with locals in the cities they are going to visit. & \{{ \it hospitality }\} \\ \midrule 
 3 & Joining such a network usually just requires filling out an online form; although some networks offer or require additional verification. & \{{ \it  }\} \\ \midrule 
 3 & A listing of available hosts is then provided either in print and/or online, sometimes with references and reviews by other travelers. & \{{ \it hosts, ref }\} \\ \midrule 
 3 & Couchsurfing was founded in January 2004 after computer programmer Casey Fenton found a cheap flight to Iceland but did not have a place to stay. & \{{ \it Couchsurfing, Jan, Casey Fenton }\} \\ \midrule 
 3 & He emailed students at the local university and received an overwhelming number of offers for free accommodation. & \{{ \it  }\} \\ \midrule 
 3 & Unlike other primates, hominids no longer use their hands in locomotion or bearing weight or swinging through the trees. & \{{ \it primates, hominids, locomotion }\} \\ \midrule 
 3 & The chimpanzee's hand and foot are similar in size and length, reflecting the hand's use for bearing weight in knuckle walking. & \{{ \it chimpanzee, foot }\} \\ \midrule 
 3 & The human hand is shorter than the foot, with straighter phalanges. & \{{ \it human, foot }\} \\ \midrule 
 3 & Fossil hand bones two million to three million years old reveal this shift in specialization of the hand from locomotion to manipulation. & \{{ \it fossil, locomotion, manipulation }\} \\ \midrule 
 3 & Jardín de la Unión. This space was built as the atrium for a 17th-century convent, of which the Templo de San Diego is the sole surviving building. & \{{ \it Jardin de la Union, atrium, convent, Templo de San Diego }\} \\ \midrule 
 3 & It now functions as the central plaza, and always has a lot of things going on, day and night. & \{{ \it plaza }\} \\ \midrule 
 3 & There are a number of restaurants surrounding the garden, and in the afternoons and evening there free concerts are often given from the central gazebo. & \{{ \it gazebo }\} \\ \midrule 
 3 & Callejon del Beso (Alley of the Kiss). Two balconies separated by only 69 centimeters is home of an old love legend. & \{{ \it Callejon del Beso, Alley of the, balconies, cm }\} \\ \midrule 
 3 & For a few pennies some children will tell you the story. & \{{ \it  }\} \\ \midrule 
 3 & In developed countries today, providing deluxe bed and breakfasts has been raised to a sort of art-form. & \{{ \it form }\} \\ \midrule 
 3 & At the top end, B\&Bs obviously compete mainly on two main things: bedding and breakfast. & \{{ \it  }\} \\ \midrule 
 3 & Accordingly, at the finest such establishments one is apt to find the most luxurious bedding, maybe a handmade quilt or an antique bed. & \{{ \it quilt }\} \\ \midrule 
 3 & Breakfast may include seasonal delights of the region or the host's speciality dish. & \{{ \it host, dish }\} \\ \midrule 
 3 & The setting might be an historic old building with antique furnishings, manicured grounds and a swimming pool. & \{{ \it pool }\} \\
        \bottomrule
    \end{tabular}
    \label{tab:annotations41}
\end{table*}

\begin{table*}[]
    \centering
    \fontsize{8.4}{10}\selectfont
    \begin{tabular}{cp{10.5cm}p{3.5cm}}
        \toprule
        \textbf{signer} & \textbf{reference} & \textbf{fingerspelled terms} \\
        \midrule
 3 & The ACMA also found that despite the video being streamed on the Internet, Big Brother had not breached online content censorship laws as the media had not been stored on Big Brother's website. & \{{ \it ACMA, Big Brother, censorship, media }\} \\ \midrule 
 3 & The Broadcasting Services Act provides for the regulation of Internet content, however to be considered Internet content, it must physically reside on a server. & \{{ \it broadcasting, act, server }\} \\ \midrule 
 3 & Brazil is the largest Roman Catholic country on Earth, and the Roman Catholic Church has consistently opposed the legalization of same-sex marriage in the country. & \{{ \it sex }\} \\ \midrule 
 3 & The National Congress of Brazil has debated legalization for 10 years, and such civil marriages are currently only legal in Rio Grande do Sul. & \{{ \it civil, Rio Grande do Sul }\} \\ \midrule 
 3 & The original bill was drafted by former mayor of São Paulo, Marta Suplicy. The proposed legislation, after being amended, is now in the hands of Roberto Jefferson. & \{{ \it bill, mayor, Sao Paulo, Marta Suplicy, Roberto Jefferson }\} \\ \midrule 
 3 & Protesters hope to collect a petition of 1.2 million signatures to present to the National Congress in November. & \{{ \it Nov }\} \\ \midrule 
 3 & While you may not need a visa for short visits to certain countries as a tourist or for business, going there as an international student generally requires a longer stay than going there just as a casual tourist. & \{{ \it visa }\} \\ \midrule 
 3 & In general, staying in any foreign country for an extended period of time will require you to obtain a visa in advance. & \{{ \it visa }\} \\ \midrule 
 3 & Student visas generally have different requirements and application procedures from normal tourist or business visas. & \{{ \it visa, tourist, visa }\} \\ \midrule 
 3 & For most countries, you will need an offer letter from the institution you wish to study at, and also evidence of funds to support yourself for at least the first year of your course. & \{{ \it  }\} \\ \midrule 
 3 & Check with the institution, as well as the immigration department for the country you wish to study in for detailed requirements. & \{{ \it  }\} \\ \midrule 
 3 & Following the race, Keselowski remains the Drivers' Championship leader with 2,250 points. & \{{ \it Keselowki }\} \\ \midrule 
 3 & Seven points behind, Johnson is second with 2,243. & \{{ \it Johnson }\} \\ \midrule 
 3 & In third, Hamlin is twenty points behind, but five ahead of Bowyer. Kahne and Truex, Jr. are fifth and sixth respectively with 2,220 and 2,207 points. & \{{ \it Hamlin, Bowyer, Kahne, Truex Jr. }\} \\ \midrule 
 3 & Stewart, Gordon, Kenseth, and Harvick round out the top-ten positions for the Drivers' Championship with four races remaining in the season. & \{{ \it Stewart, Gordon, Kenseth, Harvick }\} \\ \midrule 
 3 & In the last 3 months, over 80 arrestees were released from the Central Booking facility without being formally charged. & \{{ \it booking }\} \\ \midrule 
 3 & In April this year, a temporary restaining order was issued by Judge Glynn against the facility to enforce the release of those held more than 24 hours after their intake who did not receive a hearing by a court commissioner. & \{{ \it April, Glynn, intake, date }\} \\ \midrule 
 3 & The commissioner sets bail, if granted, and formalizes the charges filed by the arresting officer. The charges are then entered into the state's computer system where the case is tracked. & \{{ \it bail, case }\} \\ \midrule 
 3 & The hearing also marks the date for the suspect’s right to a speedy trial. & \{{ \it date }\} \\ \midrule 
 3 & Karno is a well-known but controversial English tutor who taught under Modern Education and King's Glory who claimed to have 9,000 students at the peak of his career. & \{{ \it Karno, ed }\} \\ \midrule 
 3 & In his notes he used words which some parents considered coarse, and he reportedly used profanity in class. & \{{ \it notes }\} \\ \midrule 
 3 & Modern Education accused him of printing large advertisements on buses without authorisation and lying by saying that he was the chief English tutor. & \{{ \it ed, bus }\} \\ \midrule 
 3 & He has also been accused previously of copyright infringement, but was not charged. & \{{ \it  }\} \\ \midrule 
 3 & A former student said that he 'used slang in class, taught dating skills in notes, and was just like the students' friend.' & \{{ \it slang }\} \\ \midrule 
 3 & The station maintained its attitude, despite the loss of a gyroscope earlier in the space station mission, until the end of the spacewalk. & \{{ \it space station, attitude, space, gyroscope, space station mission }\} \\ \midrule 
 3 & Chiao and Sharipov reported being a safe distance from the attitude adjustment thrusters. & \{{ \it Chiao, Sharipov, attitude, thrusters }\} \\ \midrule 
 3 & Russian ground control activated the jets and normal attitude of the station was regained. & \{{ \it jets, station, attitude }\} \\
        \bottomrule
    \end{tabular}
    \label{tab:annotations42}
\end{table*}

\begin{table*}[]
    \centering
    \fontsize{8.4}{10}\selectfont
    \begin{tabular}{cp{10.5cm}p{3.5cm}}
        \toprule
        \textbf{signer} & \textbf{reference} & \textbf{fingerspelled terms} \\
        \midrule
 3 & Using GPS navigation apps on your smartphone can be the easiest and most convenient way to navigate when out of your home country. & \{{ \it GPS, app }\} \\ \midrule 
 3 & It can save money over buying new maps for a GPS, or a standalone GPS device or renting one from a car rental company. & \{{ \it map, GPS }\} \\ \midrule 
 3 & If you don't have a data connection for your phone, or when it is out of range, their performance can be limited or unavailable. & \{{ \it data }\} \\ \midrule 
 3 & Bowen Island is a popular day trip or weekend excursion offering kayaking, hiking, shops, restaurants, and more. & \{{ \it Bowen }\} \\ \midrule 
 3 & This authentic community is located in Howe Sound just off Vancouver, and is easily accessed via scheduled water taxis departing Granville Island in downtown Vancouver. & \{{ \it Howe Sound, taxi, Granville }\} \\ \midrule 
 3 & For those who enjoy outdoor activities, a trek up the Sea to Sky corridor is essential. & \{{ \it sky, corridor }\} \\ \midrule 
 3 & Whistler (1.5 hour drive from Vancouver) is expensive but well-known because of the 2010 Winter Olympics. & \{{ \it  }\} \\ \midrule 
 3 & In the winter, enjoy some of the best skiing in North America, and in the summer try some authentic mountain biking. & \{{ \it  }\} \\ \midrule 
 3 & The band canceled the show at Maui's War Memorial Stadium, which was set to be attended by 9,000 people, and apologized to fans. & \{{ \it fans }\} \\ \midrule 
 3 & The band's management company, HK Management Inc., gave no initial reason when they canceled on September 20, but blamed logistical reasons by the next day. & \{{ \it HK, Inc, Sept }\} \\ \midrule 
 3 & The militaristic nature of the Roman empire aided in the development of medical advances. & \{{ \it  }\} \\ \midrule 
 3 & Doctors began to be recruited by Emperor Augustus and even formed the first Roman Medical Corps for use in the aftermath of battles. & \{{ \it Augustus, corps }\} \\ \midrule 
 3 & Surgeons had knowledge of various sedatives including morphine from extracts of poppy seeds and scopolamine from herbane seeds. & \{{ \it sedatives, morphine, poppy seeds, scopolamine, herbane seeds }\} \\ \midrule 
 3 & They became proficient at amputation to save patients from gangrene as well as tourniquets and arterial clamps to stem blood flow. & \{{ \it gangrene, arterial }\} \\ \midrule 
 3 & Over multiple centuries, the Roman empire led to great gains in the field of medicine and formed much of the knowledge we know today. & \{{ \it  }\} \\ \midrule 
 3 & Luxembourg has a long history but its independence dates from 1839. & \{{ \it Luxembourg }\} \\ \midrule 
 3 & Present-day parts of Belgium were part of Luxembourg in the past but became Belgian after the 1830s Belgian Revolution. & \{{ \it Luxembourg }\} \\ \midrule 
 3 & Luxembourg has always tried to remain a neutral country but it was occupied in both World War I and World War II by Germany. & \{{ \it Luxembourg, WW1, WW2 }\} \\ \midrule 
 3 & In 1957 Luxembourg became a founding member of the organization which is today known as the European Union. & \{{ \it Luxembourg }\} \\ \midrule 
 3 & The storm, situated about 645 miles (1040 km) west of the Cape Verde islands, is likely to dissipate before threatening any land areas, forecasters say. & \{{ \it Cape Verde, miles, km }\} \\ \midrule 
 3 & Fred currently has winds of 105 miles per hour (165 km/h) and is moving towards the northwest. & \{{ \it Fred, mph, km/h }\} \\ \midrule 
 3 & Fred is the strongest tropical cyclone ever recorded so far south and east in the Atlantic since the advent of satellite imagery, and only the third major hurricane on record east of 35°W. & \{{ \it Fred, tropical cyclone, Atlantic, of }\} \\ \midrule 
 3 & News spread in the Red Lake community today as funerals for Jeff Weise and three of the nine victims were held that another student was arrested in connection with the school shootings of March 21. & \{{ \it Lake, March, Jeff Weise, news }\} \\ \midrule 
 3 & Authorities said little officially beyond confirming today's arrest. & \{{ \it  }\} \\ \midrule 
 3 & However, a source with knowledge of the investigation told the Minneapolis Star-Tribune that it was Louis Jourdain, 16-year old son of Red Lake Tribal Chairman Floyd Jourdain. & \{{ \it Star Tribune, Louis Jourdain, Lake, tribal, Floyd Jourdain }\} \\ \midrule 
 3 & It is not known at this time what charges will be laid or what led authorities to the boy but juvenile proceedings have begun in federal court. & \{{ \it juvenile }\} \\ \midrule 
 3 & A Hangi Meal is cooked in a hot pit in the ground. & \{{ \it Hangi Meal, pit }\} \\ \midrule 
 3 & The pit is either heated with hot stones from a fire, or in some places geothermal heat makes areas of ground naturally hot. & \{{ \it pit, geothermal }\} \\
        \bottomrule
    \end{tabular}
    \label{tab:annotations43}
\end{table*}

\begin{table*}[]
    \centering
    \fontsize{8.4}{10}\selectfont
    \begin{tabular}{cp{10.5cm}p{3.5cm}}
        \toprule
        \textbf{signer} & \textbf{reference} & \textbf{fingerspelled terms} \\
        \midrule
 3 & The hangi is often used to cook a traditional roast style dinner. & \{{ \it hangi, roast }\} \\ \midrule 
 3 & Several places in Rotorua offer geothermal hangi, while other hangi can be sampled in Christchurch, Wellington and elsewhere. & \{{ \it Rotorua, geothermal hangi, hangi, Christchurch, Wellington }\} \\ \midrule 
 3 & The comments, live on television, were the first time that senior Iranian sources have admitted that the sanctions are having any effect. & \{{ \it live, TV, sanctions }\} \\ \midrule 
 3 & They include financial restrictions and a ban by the European Union on the export of crude oil, from which the Iranian economy receives 80\% of its foreign income. & \{{ \it sanctions, crude oil }\} \\ \midrule 
 3 & In its most recent monthly report, OPEC said exports of crude had fallen to their lowest level for two decades at 2.8 million barrels per day. & \{{ \it OPEC, crude oil }\} \\ \midrule 
 3 & The country's supreme leader, Ayatollah Ali Khamenei, has described the dependency on oil as "a trap" dating from before Iran's Islamic revolution in 1979 and from which the country should free itself. & \{{ \it Ayatollah Ali Khamenei, oil, Islamic }\} \\ \midrule 
 3 & The American plan relied on launching coordinated attacks from three different directions. & \{{ \it  }\} \\ \midrule 
 3 & General John Cadwalder would launch a diversionary attack against the British garrison at Bordentown, in order to block off any reinforcements. & \{{ \it John Cadwalder, Bordentown }\} \\ \midrule 
 3 & General James Ewing would take 700 militia across the river at Trenton Ferry, seize the bridge over the Assunpink Creek and prevent any enemy troops from escaping. & \{{ \it James Ewing, Trenton Ferry, Assunpink Creek }\} \\ \midrule 
 3 & The main assault force of 2,400 men would cross the river nine miles north of Trenton, and then split into two groups, one under Greene and one under Sullivan, in order to launch a pre-dawn attack. & \{{ \it Trenton, miles, Greene, Sullivan }\} \\ \midrule 
 3 & The invention of spoke wheels made Assyrian chariots lighter, faster, and better prepared to outrun soldiers and other chariots. & \{{ \it spoke, Assyrian chariots, chariots }\} \\ \midrule 
 3 & Arrows from their deadly crossbows could penetrate the armor of rival soldiers. About 1000 B.C., the Assyrians introduced the first cavalry. & \{{ \it crossbows, arrow, BC, Assyrians, cavalry }\} \\ \midrule 
 3 & A cavalry is an army that fights on horseback. The saddle had not yet been invented, so the Assyrian cavalry fought on the bare backs of their horses. & \{{ \it cavalry, saddle, Assyrian }\} \\ \midrule 
 3 & Architecture is concerned with the design and construction of buildings. The architecture of a place is often a tourist attraction in its own right. & \{{ \it tourist }\} \\ \midrule 
 3 & Many buildings are quite beautiful to look at and the view from a tall building or from a cleverly-positioned window can be a beauty to behold. & \{{ \it  }\} \\ \midrule 
 3 & Architecture overlaps considerably with other fields including urban planning, civil engineering, decorative arts, interior design and landscape design. & \{{ \it urban, civil }\} \\ \midrule 
 3 & Because calls are routed over the Internet, you do not need to use a phone company located where you live or where you travel. & \{{ \it  }\} \\ \midrule 
 3 & There is also no requirement that you obtain a local number from the community in which you live; you can obtain a satellite Internet connection in the wilds of Chicken, Alaska and select a number which claims you're in sunny Arizona. & \{{ \it Chicken }\} \\ \midrule 
 3 & Often, you have to buy a global number separately that allows PSTN phones to call you. Where the number is from makes a difference for people calling you. & \{{ \it PSTN }\} \\ \midrule 
 3 & Make and carry copies of your policy and your insurer's contact details with you. & \{{ \it  }\} \\ \midrule 
 3 & They need to show the insurer's e-mail address and international phone numbers for advice/authorizations and making claims. & \{{ \it claim }\} \\ \midrule 
 3 & Have another copy in your luggage and online (e-mail to yourself with attachment, or stored in the "cloud"). & \{{ \it  }\} \\ \midrule 
 3 & If traveling with a laptop or tablet, store a copy in its memory or disc (accessible without the internet). & \{{ \it disk }\} \\ \midrule 
 3 & Also give policy/contact copies to traveling companions and relatives or friends back home willing to help. & \{{ \it  }\} \\ \midrule 
 3 & Felicia, once a Category 4 storm on the Saffir-Simpson Hurricane Scale, weakened to a tropical depression before dissipating Tuesday. & \{{ \it Felicia, Saffir Simpson, tropical depression }\} \\ \midrule 
 3 & Its remnants produced showers across most of the islands, though as of yet, no damage or flooding has been reported. & \{{ \it  }\} \\ \midrule 
 3 & The precipitation, reaching 6.34 inches at a gauge on Oahu, was described as "beneficial". & \{{ \it inches, Oahu }\} \\ \midrule 
 3 & Some of the rainfall was accompanied by thunderstorms and frequent lightning. & \{{ \it  }\} \\
        \bottomrule
    \end{tabular}
    \label{tab:annotations44}
\end{table*}

\begin{table*}[]
    \centering
    \fontsize{8.4}{10}\selectfont
    \begin{tabular}{cp{10.5cm}p{3.5cm}}
        \toprule
        \textbf{signer} & \textbf{reference} & \textbf{fingerspelled terms} \\
        \midrule
 3 & A cell is the smallest structural and functional unit of a living (things) organism. & \{{ \it cell, unit, organism }\} \\ \midrule 
 3 & Cell comes from the Latin word cella which means small room. & \{{ \it cell, cella }\} \\ \midrule 
 3 & If you look at living things under a microscope, you will see that they are made of small squares or balls. & \{{ \it  }\} \\ \midrule 
 3 & Robert Hooke, a biologist from England, saw small squares in cork with a microscope. & \{{ \it Robert Hooke, cork }\} \\ \midrule 
 3 & They looked like rooms. He was the first person to observe dead cells & \{{ \it Robert, cells }\} \\ \midrule 
 3 & The Mumbai attackers arrived via boat on Novemeber 26, 2008, bringing with them grenades, automatic weapons and hit multiple targets including the crowded Chhatrapati Shivaji Terminus railway station and the famous Taj Mahal Hotel. & \{{ \it Nov, Mumbai, Chhatrapati Shivaji Terminus, Taj Mahal }\} \\ \midrule 
 3 & David Headley's scouting and information gathering had helped to enable the operation by the 10 gunmen from the Pakistani militant group Laskhar-e-Taiba. & \{{ \it David Headley, Laskhar-e-Taiba }\} \\ \midrule 
 3 & The attack put a huge strain on relations between India and Pakistan. & \{{ \it  }\} \\ \midrule 
 3 & During the 18th century Cambodia found itself squeezed between two powerful neighbors, Thailand and Vietnam. & \{{ \it 18th }\} \\ \midrule 
 3 & The Thais invaded Cambodia several times in the 18th century and in 1772 they destroyed Phnom Phen. & \{{ \it 18th, Phnom Phen }\} \\ \midrule 
 3 & In the last years of the 18th century the Vietnamese also invaded Cambodia. & \{{ \it 18th }\} \\ \midrule 
 3 & Since Pakistani independence from British rule in 1947, the Pakistani President has appointed "Political Agents" to govern FATA, who exercise near-complete autonomous control over the areas. & \{{ \it agents, FATA }\} \\ \midrule 
 3 & These agents are responsible for providing government and judicial services under Article 247 of the Pakistani Constitution. & \{{ \it agents, article }\} \\ \midrule 
 3 & Every corner store is filled with a confusing array of pre-paid phone cards that can be used from pay phones or ordinary telephones. & \{{ \it  }\} \\ \midrule 
 3 & While most cards are good for calling anywhere, some specialise in providing favourable call rates to specific groups of countries. & \{{ \it rates }\} \\ \midrule 
 3 & Access to these services is often through a toll-free telephone number that can be called from most phones without charge. & \{{ \it toll free }\} \\ \midrule 
 3 & The famous Greek lawyers, Sakis Kechagioglou and George Nikolakopoulos have been imprisoned in the Athens' jail of Korydallus, as they were found guilty of graft and corruption. & \{{ \it Sakis Kechagioglou, George Nikolakopoulos, Korydallus, graft, corruption }\} \\ \midrule 
 3 & As a result of this, a big scandal within the Greek legal community has been raised through the exposure of illegal actions that judges, lawyers, solicitors and attorneys have done during the previous years. & \{{ \it solicitors, attorneys }\} \\ \midrule 
 3 & A few weeks ago, after the information published by the journalist Makis Triantafylopoulos in his popular Television show "Zoungla" in Alpha TV, the member of Parliament and lawyer, Petros Mantouvalos was abdicated as members of his office had been involved in illegal graft and corruption. & \{{ \it Makis Triantafylopoulos, TV, Zoungla, Alpha TV, parliament, Petros Mantouvalos, graft, corruption }\} \\ \midrule 
 3 & Moreover, top judge Evangelos Kalousis is imprisoned as he found guilty of corruption and degenerate behaviour. & \{{ \it Evangelos Kalousis, corruption, degenerate }\} \\ \midrule 
 3 & Bird flu, or more formally avian influenza, can infect both birds and mammals. & \{{ \it flu, avian influenza, mammals }\} \\ \midrule 
 3 & Fewer than a thousand cases have ever been reported in humans, but some of them have been fatal. & \{{ \it cases }\} \\ \midrule 
 3 & Most have involved people who work with poultry, but there is also some risk to birdwatchers. & \{{ \it poultry, risk }\} \\ \midrule 
 3 & On Monday, scientists from the Stanford University School of Medicine announced the invention of a new diagnostic tool that can sort cells by type: a tiny printable chip that can be manufactured using standard inkjet printers for possibly about one U.S. cent each. & \{{ \it Stanford, dx, tool, cells, chip, inkjet }\} \\ \midrule 
 3 & Lead researchers say this may bring early detection of cancer, tuberculosis, HIV and malaria to patients in low-income countries, where the survival rates for illnesses such as breast cancer can be half those of richer countries. & \{{ \it chip, cancer, tuberculosis, HIV, malaria, rate, cancer }\} \\ \midrule 
 3 & Elements and compounds can move from one state to another and not change. & \{{ \it elements, compounds }\} \\ \midrule 
 3 & Nitrogen as a gas still has the same properties as liquid nitrogen. The liquid state is denser but the molecules are still the same. & \{{ \it nitrogen, liquid nitrogen, properties, liquid, dense, molecules }\} \\
        \bottomrule
    \end{tabular}
    \label{tab:annotations45}
\end{table*}

\begin{table*}[]
    \centering
    \fontsize{8.4}{10}\selectfont
    \begin{tabular}{cp{10.5cm}p{3.5cm}}
        \toprule
        \textbf{signer} & \textbf{reference} & \textbf{fingerspelled terms} \\
        \midrule
 3 & Water is another example. The compound water is made up of two hydrogen atoms and one oxygen atom. & \{{ \it compound, hydrogen atoms, atom }\} \\ \midrule 
 3 & It has the same molecular structure whether it is a gas, liquid, or solid. & \{{ \it molecule, gas, liquid, solid }\} \\ \midrule 
 3 & Although its physical state may change, its chemical state remains the same. & \{{ \it  }\} \\ \midrule 
 3 & If confirmed, the find completes Allen's eight-year search for the Musashi. & \{{ \it Musashi, Allen }\} \\ \midrule 
 3 & Following seabed mapping the wreck was found using an ROV. & \{{ \it seabed mapping, ROV }\} \\ \midrule 
 3 & One of the world's richest people, Allen has reportedly invested much of his wealth in marine exploration and began his quest to find the Musashi out of a lifelong interest in the war. & \{{ \it Allen, of, marine, Musashi }\} \\ \midrule 
 3 & Airlines such as Emirates, Etihad Airways, Qatar Airways \& Turkish Airlines have greatly expanded their services to Africa, and offer connections to many major African cities at competitive rates than other European airlines. & \{{ \it Emirates, Etihad, Qatar }\} \\ \midrule 
 3 & Turkish Airlines flies to 39 destinations in 30 African countries as of 2014. & \{{ \it  }\} \\ \midrule 
 3 & If you have additional travel time, check to see how your total fare quote to Africa compares with a round-the-world fare. & \{{ \it  }\} \\ \midrule 
 3 & Don't forget to add in the extra costs of additional visas, departure taxes, ground transportation, etc. for all those places outside of Africa. & \{{ \it visas, taxes }\} \\ \midrule 
 3 & Unwanted murals or scribble is known as graffiti. & \{{ \it murals, scribble, graffiti }\} \\ \midrule 
 3 & While it's far from a modern phenomenon, most people probably associate it with youth vandalizing public and private property using spray paint. & \{{ \it  }\} \\ \midrule 
 3 & However, nowadays there are established graffiti artists, graffiti events and "legal" walls. Graffiti paintings in this context often resemble artworks rather than illegible tags. & \{{ \it graffiti, graffiti, graffiti }\} \\ \midrule 
 3 & Humans have been making and using lenses for magnification for thousands and thousands of years. & \{{ \it lenses }\} \\ \midrule 
 3 & However, the first true telescopes were made in Europe in the late 16th century. & \{{ \it 16th }\} \\ \midrule 
 3 & These telescopes used a combination of two lenses to make distant objects appear both nearer and larger. & \{{ \it  }\} \\ \midrule 
 3 & There are varying estimates for how many people speak Hindi. It is estimated to be between the second and fourth most commonly spoken language in the world. & \{{ \it Hindi }\} \\ \midrule 
 3 & The number of native speakers varies depending on whether or not very closely related dialects are counted. & \{{ \it dialects }\} \\ \midrule 
 3 & Estimates range from 340 million to 500 million speakers, and as many as 800 million people can understand the language. & \{{ \it  }\} \\ \midrule 
 3 & Hindi and Urdu are similar in vocabulary but different in script; in everyday conversations, speakers of both languages can usually understand each other. & \{{ \it Hindi, Urdu }\} \\ \midrule 
 3 & According to Ansa, "police were concerned by a couple of top-level hits they feared might spark a full-blown war of succession. & \{{ \it Ansa }\} \\ \midrule 
 3 & Police said Lo Piccolo had the upper hand because he had been Provenzano's right-hand man in Palermo and his greater experience won him the respect of the older generation of bosses as they pursued Provenzano's policy of keeping as low as possible while strengthening their power network. & \{{ \it Lo Piccolo, Provenzano, Palermo, Provenzano }\} \\ \midrule 
 3 & These bosses had been reined in by Provenzano when he put an end to the Riina-driven war against the state that claimed the lives of Mafia crusaders Giovanni Falcone and Paolo Borsellino in 1992." & \{{ \it Provenzano, Riina, Mafia, Giovanni Falcone, Paolo Borsellino }\} \\ \midrule 
 3 & Drink alcoholic drinks with moderation. Alcohol affects everyone differently, and knowing your limit is very important. & \{{ \it  }\} \\ \midrule 
 3 & Possible long term health events from excessive drinking can include liver damage and even blindness and death. The potential danger is increased when consuming illegally produced alcohol. & \{{ \it liver }\} \\ \midrule 
 3 & Illegal spirits can contain various dangerous impurities including methanol, which can cause blindness or death even in small doses. & \{{ \it spirits, impurities, methanol }\} \\ \midrule 
 3 & As of the night of August 9, the eye of Morakot was about seventy kilometres away from the Chinese province of Fujian. & \{{ \it Aug, province, Fujian, km, Morakot }\} \\ \midrule 
 3 & The typhoon is estimated to be moving toward China at eleven kph. & \{{ \it kph }\} \\
        \bottomrule
    \end{tabular}
    \label{tab:annotations46}
\end{table*}

\begin{table*}[]
    \centering
    \fontsize{8.4}{10}\selectfont
    \begin{tabular}{cp{10.5cm}p{3.5cm}}
        \toprule
        \textbf{signer} & \textbf{reference} & \textbf{fingerspelled terms} \\
        \midrule
3 & How would Io's gravity pull on me? If you stood on the surface of Io, you would weigh less than you do on Earth. & \{{ \it Io, Io }\} \\ \midrule 
 3 & A person who weighs 200 pounds (90kg) on Earth would weigh about 36 pounds (16kg) on Io. So the gravity, of course, pulls less on you. & \{{ \it kg, Io, kg }\} \\ \midrule 
 3 & Eighteen percent of Venezuelans are unemployed, and most of those who are employed work in the informal economy. & \{{ \it  }\} \\ \midrule 
 3 & Two thirds of Venezuelans who work do so in the service sector, nearly a quarter work in industry and a fifth work in agriculture. & \{{ \it sector }\} \\ \midrule 
 3 & An important industry for Venezuelans is oil, where the country is a net exporter, even though only one percent work in the oil industry. & \{{ \it oil, oil }\} \\ \midrule 
 3 & Born in the Croatian capital, Zagreb, Bobek gained fame while playing for Partizan Belgrade. & \{{ \it Bobek, Zagreb, Partizan Belgrade }\} \\ \midrule 
 3 & He joined them in 1945 and stayed until 1958. & \{{ \it  }\} \\ \midrule 
 3 & During his time with the team, he scored 403 goals in 468 appearances. & \{{ \it  }\} \\ \midrule 
 3 & No one else has ever made more appearances or scored more goals for the club than Bobek. & \{{ \it club, Bobek }\} \\ \midrule 
 3 & In 1995 he was voted the best player in Partizan's history. & \{{ \it Partizan }\} \\ \midrule 
 3 & At least 100 people had attended the party, in order to celebrate the first anniversary of a couple whose wedding was held last year. & \{{ \it  }\} \\ \midrule 
 3 & A formal anniversary event was scheduled for a later date, officials said. & \{{ \it date }\} \\ \midrule 
 3 & The couple had married in Texas one year ago and came to Buffalo to celebrate with friends and relatives. & \{{ \it  }\} \\ \midrule 
 3 & The 30-year-old husband, who was born in Buffalo, was one of the four killed in the shooting, but his wife was not hurt. & \{{ \it of }\} \\ \midrule 
 3 & Today's announcement also extended the government's commitment made in March of this year to fund extra carriages. & \{{ \it March, carriages }\} \\ \midrule 
 3 & An additional 300 brings the total to 1,300 carriages to be acquired to relieve overcrowding. & \{{ \it  }\} \\ \midrule 
 3 & Be assured that once you hit the marinas everything will be pretty obvious. You will meet other boat hitchhikers and they will share their information with you. & \{{ \it marinas }\} \\ \midrule 
 3 & Basically you'll be putting up notices offering your help, pacing the docks, approaching people cleaning their yachts, trying to make contact with sailors in the bar, etc. & \{{ \it docks }\} \\ \midrule 
 3 & Try to talk to as many people as possible. After a while everyone will know you and will give you hints as to which boat is looking for someone. & \{{ \it  }\} \\ \midrule 
 3 & In summer, especially, you'll need to watch out for mosquitoes if you decide to hike through the rainforest. & \{{ \it  }\} \\ \midrule 
 3 & Even if you're driving through the subtropical rainforest, a few seconds with the doors open while you get inside the vehicle is enough time for mosquitoes to get in the vehicle with you. & \{{ \it subtropical }\} \\ \midrule 
 3 & In the heat of the day, travelers can experience mirages that give the illusion of water (or other things). & \{{ \it mirages }\} \\ \midrule 
 3 & These can be dangerous if the traveler pursue the mirage, wasting precious energy and remaining water. & \{{ \it  }\} \\ \midrule 
 3 & Even the hottest of deserts can become extremely cold at night. Hypothermia is a real risk without warm clothing. & \{{ \it hypothermia }\} \\ \midrule 
 3 & Japanese judoka Hitoshi Saito, winner of two Olympic gold medals, has died at the age of 54. & \{{ \it judoka, Hitoshi Saito }\} \\ \midrule 
 3 & The cause of death was announced as intrahepatic bile duct cancer. & \{{ \it intrahepatic bile duct }\} \\ \midrule
 3 & He died in Osaka on Tuesday. & \{{ \it Osaka }\} \\ \midrule 
 3 & As well as a former Olympic and World champion, Saito was the All Japan Judo Federation training committee chairman at the time of his death. & \{{ \it Saito, Judo }\} \\ \midrule 
 3 & The Cook Islands are an island country in free association with New Zealand, located in Polynesia, in the middle of the South Pacific Ocean. & \{{ \it Cook, Polynesia, Pacific }\} \\ \midrule 
 3 & It is an archipelago with 15 islands spread out over 2.2 million km2 of ocean. & \{{ \it archipelago, km }\} \\
        \bottomrule
    \end{tabular}
    \label{tab:annotations47}
\end{table*}

\begin{table*}[]
    \centering
    \fontsize{8.4}{10}\selectfont
    \begin{tabular}{cp{10.5cm}p{3.5cm}}
        \toprule
        \textbf{signer} & \textbf{reference} & \textbf{fingerspelled terms} \\
        \midrule
 3 & With the same time zone as Hawaii, the islands are sometimes thought of as "Hawaii down under". & \{{ \it zone }\} \\ \midrule 
 3 & Though smaller, it reminds some elderly visitors of Hawaii before statehood without all the large tourist hotels and other development. & \{{ \it tourist }\} \\ \midrule 
 3 & The Cook Islands do not have any cities but are composed of 15 different islands. The main ones are Rarotonga and Aitutaki. & \{{ \it Cook, Rarotonga, Aitutaki }\} \\ \midrule 
 3 & Make sure your hand is as relaxed as possible while still hitting all the notes correctly - also try not to make much extraneous motion with your fingers. & \{{ \it  }\} \\ \midrule 
 3 & This way, you will tire yourself out as little as possible. Remember there's no need to hit the keys with a lot of force for extra volume like on the piano. & \{{ \it  }\} \\ \midrule 
 3 & On the accordion, to get extra volume, you use the bellows with more pressure or speed. & \{{ \it  }\} \\ \midrule 
 3 & On September 24, 1759, Arthur Guinness signed a 9,000 year lease for the St James' Gate Brewery in Dublin, Ireland. & \{{ \it Sept, Arthur Guinness, lease, St. James Gate }\} \\ \midrule 
 3 & 250 years later, Guinness has grown to a global business that turns over 10 billion euros (US\$14.7 billion) every year. & \{{ \it Guinness }\} \\ \midrule 
 3 & The Dominican Republic (Spanish: República Dominicana) is a Caribbean country that occupies the eastern half of the island of Hispaniola, which it shares with Haiti & \{{ \it Caribbean, Hispaniola }\} \\ \midrule 
 3 & Besides white sand beaches and mountain landscapes, the country is home to the oldest European city in the Americas, now part of Santo Domingo. & \{{ \it Santo Domingo }\} \\ \midrule 
 3 & The island was first inhabited by the Taínos and Caribes. The Caribes were an Arawakan-speaking people who had arrived around 10,000 BCE. & \{{ \it Tainos, Caribes, Caribes, Arawakan, BCE }\} \\ \midrule 
 3 & Within a few short years following the arrival of European explorers, the population of Tainos had significantly been reduced by the Spanish conquerors & \{{ \it Taino }\} \\ \midrule 
 3 & Based on Fray Bartolomé de las Casas (Tratado de las Indias) between 1492 and 1498 the Spanish conquerors killed around 100,000 Taínos. & \{{ \it Fray Bartolome de las Casa, Indias, Tainos }\} \\ \midrule 
 3 & Hangeul is the only purposely invented alphabet in popular daily use. The alphabet was invented in 1444 during the reign of King Sejong (1418 – 1450). & \{{ \it Hangeul, ABC, ABC, Sejong }\} \\ \midrule 
 3 & King Sejong was the fourth king of the Joseon Dynasty and is one of the most highly regarded. & \{{ \it Sejong, Joseon, dynasty }\} \\ \midrule 
 3 & He originally named the Hangeul alphabet Hunmin Jeongeum, which means "the correct sounds for the instruction of the people". & \{{ \it Hangeul, ABC, Hunmin Jeongeum }\} \\ \midrule 
 3 & MySpace is the third most popular website used in the United States and has 54 million profiles currently. & \{{ \it MySpace, profiles }\} \\ \midrule 
 3 & These websites have gotten a lot of attention, especially in the education setting. & \{{ \it ed }\} \\ \midrule 
 3 & There are positive aspects to these websites, which include, being able to easily setup a class page which can include blogs, videos, photos, and other features. & \{{ \it blogs, features }\} \\ \midrule 
 3 & This page can easily be accessed by providing just one web address, which makes it easy to remember and easy to type in for students who may have trouble using the keyboard or with spelling. & \{{ \it URL }\} \\ \midrule 
 3 & It can be customized to make it easy to read and also with as much or little color as desired. & \{{ \it  }\} \\ \midrule 
 3 & The tourist season for the hill stations generally peaks during the Indian summer. & \{{ \it tourist, hill stations }\} \\ \midrule 
 3 & However, they have a different kind of beauty and charm during winter, with many hill stations receiving healthy amounts of snow and offering activities such as skiing and snowboarding. & \{{ \it hill stations }\} \\
 3 & While immigration check is usually absent or a formality when you arrive in your homeland, customs control can be a hassle. & \{{ \it customs }\} \\ \midrule 
 3 & Make sure you know what you can and cannot bring in and declare anything over the legal limits. & \{{ \it  }\} \\ \midrule 
 3 & For a long time during the nineteenth and twentieth centuries, it was believed the first inhabitants of New Zealand were the Maori people, who hunted giant birds called moas. & \{{ \it 19th, 20th, Maori, moas }\} \\ \midrule 
 3 & The theory then established the idea that the Maori people migrated from Polynesia in a Great Fleet and took New Zealand from the Moriori, establishing an agricultural society. & \{{ \it Maori, Polynesia, Moriori }\} \\ \midrule 
 3 & However, new evidence suggests that the Moriori were a group of mainland Maori who migrated from New Zealand to the Chatham Islands, developing their own distinctive, peaceful culture. & \{{ \it Moriori, Maori, Chatham, own }\} \\
        \bottomrule
    \end{tabular}
    \label{tab:annotations48}
\end{table*}

\clearpage
\begin{table*}[]
    \centering
    \fontsize{8.4}{10}\selectfont
    \begin{tabular}{cp{10.5cm}p{3.5cm}}
        \toprule
        \textbf{signer} & \textbf{reference} & \textbf{fingerspelled terms} \\
        \midrule
 3 & There was also another tribe on the Chatham islands these were Maori who migrated away from New Zealand. & \{{ \it tribe, Chatham, Maori }\} \\ \midrule 
 3 & They called themselves the Moriori there were a few skirmishes and in the end, the Moriori were wiped out & \{{ \it tribe, Moriori, Moriori }\} \\ \midrule 
 3 & The Sun doesn't have a crust like the Earth that you can stand on. The whole Sun is made out of gases, fire, and plasma. & \{{ \it crust, gas, plasma }\} \\ \midrule 
 3 & The gas becomes thinner as you go farther from the center of the Sun. & \{{ \it gas }\} \\ \midrule 
 3 & The outer-part we see when we look at the Sun is called the photosphere, which means "ball of light". & \{{ \it photosphere }\} \\ \midrule 
 3 & To return to their previous energy level, they must get rid of the extra energy they got from the light. & \{{ \it  }\} \\ \midrule 
 3 & They do this by emitting a tiny particle of light called a "photon". & \{{ \it photon }\} \\ \midrule 
 3 & Scientists call this process "stimulated emission of radiation" because the atoms are stimulated by the bright light, causing the emission of a photon of light, and light is a type of radiation. & \{{ \it radiation, atoms, photon, radiation }\} \\ \midrule 
 3 & The next picture shows the atoms emitting photons. Of course, in reality photons are a lot smaller than those in the picture. & \{{ \it atoms, photons, photons }\} \\ \midrule 
 3 & Photons are even smaller than the stuff that makes up atoms! & \{{ \it photons, atoms }\} \\ \midrule 
 3 & Learning to create interactive media requires conventional and traditional skills, as well as tools mastered in interactive classes (storyboarding, audio and video editing, story telling, etc.) & \{{ \it media, tools }\} \\ \midrule 
 3 & Interactive design requires that you re-assess your assumptions about media production and learn to think in a non-linear ways. & \{{ \it media }\} \\ \midrule 
 3 & Interactive design requires that components of a project connect to each other, but also make sense as a separate entity. & \{{ \it  }\} \\ \midrule 
 3 & Given how remote many of the pueblos are, you won't be able to find a significant amount of nightlife without traveling to Albuquerque or Santa Fe. & \{{ \it pueblos, Albuq, SF }\} \\ \midrule 
 3 & However, nearly all of the casinos listed above serve drinks, and several of them bring in name-brand entertainment (primarily the large ones immediately surrounding Albuquerque and Santa Fe). & \{{ \it Albuq, SF }\} \\ \midrule 
 3 & Beware: small-town bars here are not always good places for the out-of-state visitor to hang out. & \{{ \it  }\} \\ \midrule 
 3 & For one thing, northern New Mexico has significant problems with drunk driving, and the concentration of intoxicated drivers is high close to small-town bars. & \{{ \it NM }\} \\ \midrule 
 3 & Middle distance running is a relatively inexpensive sport; however, there are many misconceptions regarding the few pieces of equipment required to participate. & \{{ \it  }\} \\ \midrule 
 3 & Products can be purchased as needed, but most will have little or no real impact on performance. & \{{ \it  }\} \\ \midrule 
 3 & Athletes may feel that they prefer a product even when it provides no real benefits. & \{{ \it  }\} \\ \midrule 
 3 & Ecuador requires that Cuban citizens receive an invitation letter prior to entering Ecuador through international airports or frontier admission points. & \{{ \it  }\} \\ \midrule 
 3 & This letter must be legalized by the Ecuadorian Ministry of Foreign Affairs, and comply with certain requirements. & \{{ \it ministry, affairs }\} \\ \midrule 
 3 & These requirements are designed to provide an organized migratory flow between both countries. & \{{ \it  }\} \\ \midrule 
 3 & Cuban citizens who are US green card holders should visit an Ecuadorian Consulate to obtain an exemption to this requirement. & \{{ \it US, consulate }\} \\ \midrule 
 3 & Your passport must be valid for at least 6 months beyond your travel dates. A round/onward trip ticket is needed to prove the length of your stay. & \{{ \it  }\} \\ \midrule 
 3 & The major organ of the circulatory system is the heart, which pumps the blood. & \{{ \it circulatory, organ }\} \\ \midrule
 3 & Blood goes away from the heart in tubes called arteries and comes back to the heart in tubes called veins. The smallest tubes are called capillaries. & \{{ \it tubes, arteries, veins, tubes, capillaries }\} \\ \midrule 
 3 & The disadvantage of zoom lenses is that the focal complexity and number of lens elements required to achieve a range of focal lengths is much greater than for prime lenses. & \{{ \it focal, lens, prime }\} \\
        \bottomrule
    \end{tabular}
    \label{tab:annotations49}
\end{table*}

\clearpage 
\begin{table*}[]
    \centering
    \fontsize{8.4}{10}\selectfont
    \begin{tabular}{cp{10.5cm}p{3.5cm}}
        \toprule
        \textbf{signer} & \textbf{reference} & \textbf{fingerspelled terms} \\
        \midrule
 3 & This is becoming less of an issue as lens manufacturers achieve higher standards in lens production. & \{{ \it  }\} \\ \midrule 
 3 & This has allowed zoom lenses to produce pictures of a quality comparable to that achieved by lenses with fixed focal length. & \{{ \it  }\} \\ \midrule 
 3 & Another disadvantage of zoom lenses is that the maximum aperture (the speed) of the lens is usually lower. & \{{ \it aperture }\} \\ \midrule 
 3 & This makes inexpensive zoom lenses hard to use in low-light conditions without a flash. & \{{ \it  }\} \\ \midrule 
 3 & Tornadoes strike a small area compared to other violent storms, but they can destroy everything in their path. & \{{ \it  }\} \\ \midrule 
 3 & Tornadoes uproot trees, rip boards from buildings, and fling cars up into the sky. The most violent two percent of tornadoes last more than three hours. & \{{ \it  }\} \\ \midrule 
 3 & These monster storms have winds up to 480 km/h (133 m/s; 300 mph). & \{{ \it kmph, m/s, mph }\} \\ \midrule 
 3 & It is the biggest acquisition in eBay's history. & \{{ \it eBay }\} \\ \midrule 
 3 & The company hopes to diversify its profit sources and gain popularity in areas where Skype holds a strong position, such as China, Eastern Europe, and Brazil. & \{{ \it co, Skype }\} \\ \midrule 
 3 & According to reports, an apartment on Macbeth Street exploded due to a gas leak. & \{{ \it apt, Macbeth St, gas leak }\} \\ \midrule 
 3 & An official with the gas company was reporting to the scene after a neighbor called about a gas leak. & \{{ \it gas co, gas leak }\} \\ \midrule 
 3 & When the official arrived, the apartment exploded. & \{{ \it  }\} \\ \midrule 
 3 & No major injuries were reported, but at least five people on scene at the time of the explosion were treated for symptoms of shock. & \{{ \it  }\} \\ \midrule 
 3 & No one was inside the apartment. & \{{ \it apt }\} \\ \midrule 
 3 & At the time, nearly 100 residents were evacuated from the area. & \{{ \it  }\} \\ \midrule 
 3 & The temple of Artemis at Ephesus was destroyed on July 21, 356 BCE in an act of arson committed by Herostratus. & \{{ \it July, BCE, of Artemis, Ephesus, Herostratus }\} \\ \midrule 
 3 & According to the story, his motivation was fame at any cost. The Ephesians, outraged, announced that Herostratus' name never be recorded. & \{{ \it Ephesians, Herostratus }\} \\ \midrule 
 3 & The Greek historian Strabo later noted the name, which is how we know today. The temple was destroyed on the same night that Alexander the Great was born. & \{{ \it Strabo, Alexander the Great }\} \\ \midrule 
 3 & Alexander, as king, offered to pay to rebuild the temple, but his offer was denied. Later, after Alexander died, the temple was rebuilt in 323 BCE. & \{{ \it Alexander, BCE }\} \\ \midrule 
 3 & Arias tested positive for a mild case of the virus, Presidential Minister Rodrigo Arias said. & \{{ \it Minister Rodrigo Arias, mild case, virus }\} \\ \midrule 
 3 & The president's condition is stable, though he will be isolated at home for several days. & \{{ \it  }\} \\ \midrule 
 3 & "Apart from the fever and a sore throat, I feel well and in good shape to carry out my work by telecommuting. & \{{ \it  }\} \\ \midrule 
 3 & I expect to return to all my duties on Monday," Arias said in a statement. & \{{ \it  }\} \\ \midrule 
 3 & At its peak, Tropical Cyclone Gonu, named for a bag of palm leaves in the language of the Maldives, reached sustained winds of 240 kilometers an hour (149 miles per hour) . & \{{ \it Tropical, Gonu, bag, palm, Maldives, kmph, mph }\} \\ \midrule 
 3 & By early today, winds were around 83 km/h, and it was expect to keep weakening. & \{{ \it kmph }\} \\ \midrule 
 3 & With the change from the quarter to the half mile run, speed becomes of much less importance and endurance becomes an absolute necessity. & \{{ \it mile }\} \\ \midrule 
 3 & Of course a first-class half-miler, a man who can beat two minutes, must be possessed of a fair amount of speed, but endurance must be cultivated at all hazards. & \{{ \it mile, hazards }\} \\ \midrule 
 3 & Some cross country running during the winter, combined with gymnasium work for the upper part of the body, is the best preparation for the running season. & \{{ \it season }\} \\ \midrule 
 3 & Personal VPN (virtual private network) providers are an excellent way of circumventing both political censorship and commercial IP-geofiltering. & \{{ \it VPN, virtual, IP geofiltering }\} \\ \midrule
 3 & They are superior to web proxies for several reasons: They re-route all Internet traffic, not only http. & \{{ \it web proxies, http }\} \\ \midrule 
 3 & They normally offer higher bandwidth and better quality of service. They are encrypted and thus harder to spy on. & \{{ \it BW }\} \\
        \bottomrule
    \end{tabular}
    \label{tab:annotations50}
\end{table*}

\clearpage 
\begin{table*}[]
    \centering
    \fontsize{8.4}{10}\selectfont
    \begin{tabular}{cp{10.5cm}p{3.5cm}}
        \toprule
        \textbf{signer} & \textbf{reference} & \textbf{fingerspelled terms} \\
        \midrule
 3 & The United States embassy located in Nairobi, Kenya has issued a warning that "extremists from Somalia" are planning to launch suicide bomb attacks in Kenya and Ethiopia. & \{{ \it US embassy, Nairobi, extremists, suicide }\} \\ \midrule 
 3 & The U.S. says it has received information from an undisclosed source that specifically mentions the use of suicide bombers to will blow up "prominent landmarks" in Ethiopia and Kenya. & \{{ \it US, source, suicide }\} \\ \midrule 
 3 & Long before The Daily Show and The Colbert Report, Heck and Johnson envisioned a publication that would parody the news—and news reporting—when they were students at UW in 1988. & \{{ \it The Daily Show, The Colbert Report, UW, Heck, Johnson, news, news }\} \\ \midrule 
 3 & Since its inception, The Onion has become a veritable news parody empire, with a print edition, a website that drew 5,000,000 unique visitors in the month of October, personal ads, a 24 hour news network, podcasts, and a recently launched world atlas called Our Dumb World. & \{{ \it the, news, Oct, ads, news, podcasts, atlas }\} \\ \midrule 
 3 & Al Gore and General Tommy Franks casually rattle off their favorite headlines (Gore's was when The Onion reported he and Tipper were having the best sex of their lives after his 2000 Electoral College defeat). & \{{ \it Al Gore, Tommy Franks, Gore, Tipper, sex, college }\} \\ \midrule 
 3 & Many of their writers have gone on to wield great influence on Jon Stewart and Stephen Colbert's news parody shows. & \{{ \it Jon Stewart, Stephen Colbert, news }\} \\ \midrule 
 3 & The world has over 5,000 different languages, including more than twenty with 50 million or more speakers. & \{{ \it  }\} \\ \midrule 
 3 & Written words are often easier to understand than spoken words, also. This is especially true of addresses, which are often difficult to pronounce intelligibly. & \{{ \it  }\} \\ \midrule 
 3 & Many entire nations are completely fluent in English, and in even more you can expect a limited knowledge - especially among younger people. & \{{ \it  }\} \\ \midrule 
 3 & Imagine, if you will, a Mancunian, Bostonian, Jamaican and Sydneysider sitting around a table having dinner at a restaurant in Toronto. & \{{ \it Mancunian }\} \\ \midrule 
 3 & They're regaling each other with stories from their hometowns, told in their distinct accents and local argot. & \{{ \it argot }\} \\ \midrule 
 3 & Asynchronous communication encourages time for reflection and reaction to others. & \{{ \it asynchronous }\} \\ \midrule 
 3 & It allows students the ability to work at their own pace and control the pace of instructional information. & \{{ \it own }\} \\ \midrule 
 3 & In addition, there are fewer time restrictions with the possibility of flexible working hours. (Bremer, 1998) & \{{ \it Bremer }\} \\ \midrule 
 3 & The use of the Internet and the World Wide Web allows learners to have access to information at all times. & \{{ \it WWW }\} \\ \midrule 
 3 & Students can also submit questions to instructors at any time of day and expect reasonably quick responses, rather than waiting until the next face-to-face meeting. & \{{ \it  }\} \\ \midrule 
 3 & Army ant colonies march and nest in different phases as well. & \{{ \it nest }\} \\ \midrule 
 3 & In the nomadic phase, army ants march at night and stop to camp during the day. & \{{ \it nomadic }\} \\ \midrule 
 3 & The colony begins a nomadic phase when available food has decreased. During this phase, the colony makes temporary nests that are changed everyday. & \{{ \it nomadic, colony, nests }\} \\ \midrule 
 3 & Each of these nomadic rampages or marches lasts for approximately 17 days. & \{{ \it nomadic }\} \\ \midrule 
 3 & Christopher Garcia, a spokesperson of the Los Angeles Police Department, said the suspected male offender is being investigated for trespassing rather than vandalism. & \{{ \it Christopher Garcia, LA, offender, vandalism }\} \\ \midrule 
 3 & The sign was not physically damaged; the modification was done using black tarpaulins decorated with signs of peace and heart to alter the "O" to read lowercase "e". & \{{ \it tarp, O, e }\} \\ \midrule 
 3 & One of the most common problems when trying to convert a movie to DVD format is the overscan. & \{{ \it DVD, overscan }\} \\ \midrule 
 3 & Most televisions are made in a way to please the general public. & \{{ \it TV }\} \\ \midrule 
 3 & For that reason, everything you see on the TV had the borders cut, top, bottom and sides. & \{{ \it TV }\} \\ \midrule 
 3 & This is made to ensure that the image covers the whole screen. That is called overscan. & \{{ \it overscan }\} \\ \midrule 
 3 & Unfortunately, when you make a DVD, it's borders will most likely be cut too, and if the video had subtitles too close to the bottom, they won't be fully shown. & \{{ \it DVD, subtitles }\} \\ \midrule
  3 & Greed and selfishness will always be with us and it is the nature of cooperation that when the majority benefit there will always be more to gain in the short term by acting selfishly & \{{ \it greed }\} \\
        \bottomrule
    \end{tabular}
    \label{tab:annotations51}
\end{table*}

\clearpage
\begin{table*}[]
    \centering
    \fontsize{8.4}{10}\selectfont
    \begin{tabular}{cp{10.5cm}p{3.5cm}}
        \toprule
        \textbf{signer} & \textbf{reference} & \textbf{fingerspelled terms} \\
        \midrule
 3 & Hopefully, most people will realise that their long term best option is to work together with others. & \{{ \it  }\} \\ \midrule
 3 & Vision, or the ability to see depends on visual system sensory organs or eyes. & \{{ \it sensory organs }\} \\ \midrule 
 3 & There are many different constructions of eyes, ranging in complexity depending on the requirements of the organism. & \{{ \it organism }\} \\ \midrule 
 3 & The different constructions have different capabilities, are sensitive to different wave-lengths and have differing degrees of acuity, also they require different processing to make sense of the input and different numbers to work optimally. & \{{ \it wavelengths, acuity, input }\} \\ \midrule 
 3 & Tomb of Tutankhamun (KV62). KV62 may be the most famous of the tombs in the Valley, the scene of Howard Carter's 1922 discovery of the almost intact royal burial of the young king. & \{{ \it tomb, Tutankhamun, KV62, Howard Carter }\} \\ \midrule 
 3 & Compared to most of the other royal tombs, however, the tomb of Tutankhamun is barely worth visiting, being much smaller and with limited decoration. & \{{ \it Tutankhamun }\} \\ \midrule 
 3 & Anyone interested in seeing evidence of the damage to the mummy done during attempts to remove it from the coffin will be disappointed as only the head and shoulders are visible. & \{{ \it mummy }\} \\ \midrule 
 3 & The fabulous riches of the tomb are no longer in it, but have been removed to the Egyptian Museum in Cairo. & \{{ \it Cairo }\} \\ \midrule 
 3 & Visitors with limited time would be best to spend their time elsewhere. & \{{ \it  }\} \\ \midrule 
 3 & A triceratops' teeth would have been able to crush not only leaves but even very tough branches and roots. & \{{ \it triceratops }\} \\ \midrule 
 3 & Some scientists think Triceratops ate cycads, which are a type of plant that was common in the Cretaceous. & \{{ \it cycads, Cretaceous }\} \\ \midrule 
 3 & These plants look like a small palm tree with a crown of sharp, spiky leaves. & \{{ \it palm }\} \\ \midrule 
 3 & A Triceratops could have used its strong beak to strip off the leaves before eating the trunk. & \{{ \it  }\} \\ \midrule 
 3 & Other scientists argue that these plants are very poisonous so it is unlikely that any dinosaur ate them, even though today the sloth and other animals like the parrot (a descendant of the dinosaurs) can eat poisonous leaves or fruit. & \{{ \it sloth, parrot }\} \\ \midrule 
 3 & Rules regarding regular photography also apply to video recording, possibly even more so. & \{{ \it  }\} \\ \midrule 
 3 & If just taking a photo of something isn't allowed, then you shouldn't even think of recording a video of it. & \{{ \it  }\} \\ \midrule 
 3 & If using a drone, check well in advance as to what you are allowed to film and what permits or additional licensing are required. & \{{ \it drone }\} \\ \midrule 
 3 & Flying a drone near an airport or over a crowd is almost always a bad idea, even if it's not illegal in your area. & \{{ \it  }\} \\ \midrule 
 3 & Dr. Lee also expressed his concern about reports that children in Turkey have now become infected with the A(H5N1) avian influenza virus without becoming ill. & \{{ \it Lee, A(H5N1) avian influenza virus }\} \\ \midrule 
 3 & Some studies suggest that the disease must become less lethal before it can cause a global epidemic, he noted. & \{{ \it lethal }\} \\ \midrule 
 3 & There is concern that patients may continue to infect more people by going through their daily routines if the flu symptoms remain mild. & \{{ \it flu }\} \\ \midrule 
 3 & The glaciers are not stable, but flow down the mountain. This will cause cracks, crevasses, which may be obscured by snow bridges. & \{{ \it glaciers }\} \\ \midrule 
 3 & The walls and roofs of ice caves can collapse and cracks can get closed. & \{{ \it ice cave }\} \\ \midrule 
 3 & At the edge of glaciers huge blocks break loose, fall down and perhaps jump or roll farther from the edge. & \{{ \it glaciers }\} \\ \midrule 
 3 & Crime, and ill-will toward foreigners in general, is virtually unknown in Greenland. Even in the towns, there are no "rough areas." & \{{ \it crime, Greenland }\} \\ \midrule 
 3 & Cold weather is perhaps the only real danger the unprepared will face. & \{{ \it  }\} \\ \midrule 
 3 & If you visit Greenland during cold seasons (considering that the further north you go, the colder it will be), it is essential to bring warm enough clothing. & \{{ \it Greenland }\} \\ \midrule 
 3 & The very long days in the summer can lead to problems getting sufficient sleep and associated health issues. & \{{ \it  }\} \\
        \bottomrule
    \end{tabular}
    \label{tab:annotations52}
\end{table*}

\begin{table*}[]
    \centering
    \fontsize{8.4}{10}\selectfont
    \begin{tabular}{cp{10.5cm}p{3.5cm}}
        \toprule
        \textbf{signer} & \textbf{reference} & \textbf{fingerspelled terms} \\
        \midrule
 3 & During the summer, also watch out for the Nordic mosquitoes. Although they do not transmit any diseases, they can be irritating. & \{{ \it Nordic }\} \\ \midrule 
 3 & Many people dream of the day when humans can travel to another star and explore other worlds, some people wonder what's out there some belive that aliens or other life may live on another plant. & \{{ \it form }\} \\ \midrule
 3 & But, if this ever does happen probably won't happen for a very long time. The stars are so spread out that there are trillions of miles between stars that are "neighbors". & \{{ \it trillions, miles }\} \\ \midrule 
 3 & Maybe one day, your great grandchildren will be standing atop an alien world wondering about their ancient ancestors? & \{{ \it great, grand }\} \\ \midrule 
 3 & Hesperonychus elizabethae is a species of the family Dromaeosauridae and is a cousin of Velociraptor . & \{{ \it Hesperonychus elizabethae, species, Dromaesauridae, Velociraptor }\} \\ \midrule 
 3 & This fully feathered, warm blooded bird of prey was believed to have walked upright on two legs with claws like the Velociraptor. & \{{ \it did, prey, Velociraptor }\} \\ \midrule 
 3 & Its second claw was larger, giving rise to the name Hesperonychus which means "western claw." & \{{ \it Hesperonychus, western claw }\} \\ \midrule 
 3 & Traveling to Walt Disney World represents a major pilgrimage for many American families. & \{{ \it  }\} \\ \midrule 
 3 & The "typical" visit involves flying into Orlando International Airport, busing to an on-site Disney hotel, spending about a week without leaving Disney property, and returning home. & \{{ \it bus }\} \\ \midrule 
 3 & There are infinite variations possible, but this remains what most people mean when they talk of "going to Disney World". & \{{ \it  }\} \\ \midrule 
 3 & Many tickets sold online through auction websites such as eBay or Craigslist are partially used multi-day park-hopper tickets. & \{{ \it eBay, CL, park }\} \\ \midrule 
 3 & While this is a very common activity, it is prohibited by Disney: the tickets are non-transferable. & \{{ \it  }\} \\ \midrule 
 3 & The atom can be considered to be one of the fundamental building blocks of all matter. & \{{ \it matter, atom }\} \\ \midrule 
 3 & Its a very complex entity which consists, according to a simplified Bohr model, of a central nucleus orbited by electrons, somewhat similar to planets orbiting the sun - see Figure 1.1. & \{{ \it Bohr, nucleus, electrons, Figure }\} \\ \midrule 
 3 & The nucleus consists of two particles - neutrons and protons. & \{{ \it nucleus, particles, neutrons, protons }\} \\ \midrule 
 3 & Protons have a positive electric charge while neutrons have no charge. The electrons have a negative electric charge. & \{{ \it protons, neutrons, electrons }\} \\ \midrule 
 3 & The Colonists, seeing this activity, had also called for reinforcements. & \{{ \it colonist }\} \\ \midrule 
 3 & Troops reinforcing the forward positions included the 1st and 3rd New Hampshire regiments of 200 men, under Colonels John Stark and James Reed (both later became generals). & \{{ \it troops, NH, regiments, Colonel John Stark, James Reed }\} \\ \midrule 
 3 & Stark's men took positions along the fence on the north end of the Colonist's position. & \{{ \it Stark }\} \\ \midrule 
 3 & When low tide opened a gap along the Mystic River along the northeast of the peninsula, they quickly extended the fence with a short stone wall to the north ending at the water's edge on a small beach. & \{{ \it tide, Mystic, beach }\} \\ \midrule 
 3 & Gridley or Stark placed a stake about 100 feet (30 m) in front of the fence and ordered that no one fire until the regulars passed it. & \{{ \it Gridley, Stark, feet, m, regulars }\} \\ \midrule 
 3 & Every year around October nearly 1.5 million herbivores travel towards the southern plains, crossing the Mara River, from the northern hills for the rains. & \{{ \it Oct, herbivores, hills, Mara }\} \\ \midrule 
 3 & And then back to the north through the west, once again crossing the Mara river, after the rains in around April. & \{{ \it April, Mara }\} \\ \midrule 
 3 & The Serengeti region contains the Serengeti National Park, the Ngorongoro Conservation Area and Maswa Game Reserve in Tanzania and the Maasai Mara National Reserve in Kenya. & \{{ \it Serengeti, Seregeti, Ngorongoro, Maswa Game Reserve, Maasai Mara, reserve }\} \\ \midrule 
 3 & The traditional medieval castle has long inspired the imagination, conjuring up images of jousts, banquets and Arthurian chivalry. & \{{ \it medieval, Arthurian }\} \\ \midrule 
 3 & Even standing amidst thousand year-old ruins it is easy to bring to mind the sounds and smells of battles long gone, to almost hear the clatter of hooves on the cobbles and to smell the fear rising from the dungeon pits. & \{{ \it pits }\} \\
        \bottomrule
    \end{tabular}
    \label{tab:annotations53}
\end{table*}

\begin{table*}[]
    \centering
    \fontsize{8.4}{10}\selectfont
    \begin{tabular}{cp{10.5cm}p{3.5cm}}
        \toprule
        \textbf{signer} & \textbf{reference} & \textbf{fingerspelled terms} \\
        \midrule
 3 & But is our imagination based on reality? Why were castles built in the first place? How were they designed and built? & \{{ \it  }\} \\ \midrule 
 3 & Unless you are a diplomat, working overseas generally means that you will have to file income tax in the country you are based in. & \{{ \it diplomat, tax }\} \\ \midrule 
 3 & Income tax is structured differently in different countries, and the tax rates and brackets vary widely from one country to another. & \{{ \it tax, tax, rates, brackets }\} \\ \midrule 
 3 & In some federal countries, such as the United States and Canada, income tax is levied both at the federal level and at the local level, so the rates and brackets can vary from region to region. & \{{ \it US, tax, rates, brackets }\} \\ \midrule 
 3 & Only a few airlines still offer bereavement fares, which slightly discount the cost of last-minute funeral travel. & \{{ \it Co, fares }\} \\ \midrule 
 3 & Airlines that offer these include Air Canada, Delta Air Lines, Lufthansa for flights originating from the U.S. or Canada, and WestJet. & \{{ \it Co, Air, Delta, Lufthansa, US, Jet }\} \\ \midrule 
 3 & In all cases, you must book by phone directly with the airline. & \{{ \it Co }\} \\ \midrule 
 3 & After it became apparent that many families were seeking legal help to fight the evictions, a meeting was held on March 20 at the East Bay Community Law Center for the victims of the housing scam. & \{{ \it March, Bay, scam }\} \\ \midrule 
 3 & When the tenants started sharing what had occurred to them, most of the families involved suddenly realized that Carolyn Wilson of the OHA had stolen their security deposits, and skipped out of town. & \{{ \it tenants, Carolyn Wilson, OHA }\} \\ \midrule 
 3 & Tenants at Lockwood Gardens believe that there may be another 40 families or more to face eviction, since they learned that OHA police are also investigating other public housing properties in Oakland that may be caught up in the housing scam. & \{{ \it Tenants, Lockwood Gardens, OHA, public, scam }\} \\ \midrule 
 3 & The proposed amendment already passed both houses in 2011. & \{{ \it  }\} \\ \midrule 
 3 & A change was made this legislative session when the second sentence was deleted first by the House of Representatives and then was passed in a similar form by the Senate Monday. & \{{ \it  }\} \\ \midrule 
 3 & The failure of the second sentence, which proposes to ban same-sex civil unions, could possibly open the door for civil unions in the future. & \{{ \it civil union, civil union }\} \\ \midrule 
 3 & Following the process, HJR-3 will be reviewed again by the next elected legislature in either 2015 or 2016 to remain in process. & \{{ \it HJR-3 }\} \\ \midrule 
 3 & Coffee is one of the world's most traded commodities, and you can probably find many types in your home region. & \{{ \it goods }\} \\ \midrule 
 3 & Nevertheless, there are many distinctive ways of drinking coffee around the world that are worth experiencing. & \{{ \it  }\} \\ \midrule 
 3 & "Based on this fossil, that means the split is much earlier than has been anticipated by the molecular evidence. & \{{ \it fossil, molecular }\} \\ \midrule 
 3 & That means everything has to be put back," said researcher at the Rift Valley Research Service in Ethiopia and a co-author of the study, Berhane Asfaw. & \{{ \it Rift Valley, Berhane Asfaw }\} \\ \midrule 
 3 & The Luno had 120–160 cubic metres of fuel aboard when it broke down and high winds and waves pushed it into the breakwater. & \{{ \it Luno, cubic meters, fuel, breakwater }\} \\ \midrule 
 3 & Helicopters rescued the twelve crewmembers and the only injury was a broken nose. & \{{ \it crew }\} \\ \midrule 
 3 & The 100-metre ship was en route to pick up its usual fertiliser cargo and initially officials feared the vessel could spill a load. & \{{ \it meters, fertilizer cargo, did }\} \\ \midrule 
 3 & Most of the temples do have an Annual festival starting from November end to Mid of May, which varies according to each temple's annual calendar. & \{{ \it Nov, May }\} \\ \midrule 
 3 & Most of the temple festivals is celebrated as part of temple's anniversary or presiding deity's birthday or any other major event associated with the temple. & \{{ \it  }\} \\ \midrule 
 3 & Kerala's temple festivals are very interesting to see, with regular procession of decorated elephants, temple orchestra and other festivities. & \{{ \it Kerala }\} \\ \midrule 
 3 & Until now, AOL has been able to move and develop the IM market at its own pace, due to its widespread use within the United States. & \{{ \it AOL, IM, US }\} \\ \midrule 
 3 & With this arrangement in place, this freedom might end. & \{{ \it  }\} \\ \midrule 
 3 & The number of users of the Yahoo! and Microsoft services combined will rival the number of AOL's customers. & \{{ \it Yahoo, MS, AOL }\} \\
        \bottomrule
    \end{tabular}
    \label{tab:annotations54}
\end{table*}

\begin{table*}[]
    \centering
    \fontsize{8.4}{10}\selectfont
    \begin{tabular}{cp{10.5cm}p{3.5cm}}
        \toprule
        \textbf{signer} & \textbf{reference} & \textbf{fingerspelled terms} \\
        \midrule
 3 & Scientists have suspected Enceladus as geologically active and a possible source of Saturn's icy E ring. & \{{ \it Enceladus, Saturn Icy E }\} \\ \midrule 
 3 & Enceladus is the most reflective object in the solar system, reflecting about 90 percent of the sunlight that hits it. & \{{ \it Enceladus }\} \\ \midrule 
 3 & A population is the collection of organisms of a particular species within a given geographic area. & \{{ \it population, organisms, species }\} \\ \midrule 
 3 & When all individuals in a population are identical with regard to a particular phenotypic trait they are known as monomorphic. & \{{ \it population, phenotypic trait, monomorphic }\} \\ \midrule 
 3 & When the individuals show several variants of a particular trait they are polymorphic. & \{{ \it trait, polymorphic }\} \\ \midrule 
 3 & At home, due to this constant exposure to the local germs, odds are very high that you're already immune to them. & \{{ \it germs }\} \\ \midrule 
 3 & But in other parts of the world, where the bacteriological fauna are new to you, you're far more likely to run into problems. & \{{ \it bacteriological fauna }\} \\ \midrule 
 3 & Also, in warmer climates bacteria both grow more quickly and survive longer outside the body. & \{{ \it bacteria }\} \\ \midrule 
 3 & Thus the scourges of Delhi Belly, the Pharaoh's Curse, Montezuma's Revenge, and their many friends. & \{{ \it Delhi Belly, Montezuma }\} \\ \midrule 
 3 & As with respiratory problems in colder climates, intestinal problems in hot climates are fairly common and in most cases are distinctly annoying but not really dangerous. & \{{ \it intestinal }\} \\ \midrule 
 3 & With the battle for France over, Germany began to get ready to invade the island of Britain. & \{{ \it  }\} \\ \midrule 
 3 & Germany code-named the attack “Operation Sealion”. Most of the British Army's heavy weapons and supplies had been lost when it evacuated from Dunkirk, so the army was fairly weak. & \{{ \it code, Operation Sealion, weapons, supplies, Dunkirk }\} \\ \midrule 
 3 & But the Royal Navy was still much stronger than the German Navy (“Kriegsmarine”) and could have destroyed any invasion fleet sent across the English Channel. & \{{ \it Kriegsmarine, English Channel }\} \\ \midrule 
 3 & However, very few Royal Navy ships were based near the likely invasion routes as the admirals were afraid they would be sunk by German air attack. & \{{ \it  }\} \\ \midrule 
 3 & Some hotels have a heritage from the golden age of steam railways and ocean liners; before the Second World War, in the 19th or early 20th centuries. & \{{ \it steam, WW2, 19th, 20th }\} \\ \midrule 
 3 & These hotels were where the rich and the famous of the day would stay, and often had fine dining and nightlife. & \{{ \it  }\} \\ \midrule 
 3 & The old fashioned fittings, the lack of the latest amenities, and a certain graceful agedness are also part of their character. & \{{ \it  }\} \\ \midrule 
 3 & While they are usually privately owned, they sometimes accommodate visiting heads of state and other dignitaries. & \{{ \it dignitaries }\} \\ \midrule 
 3 & A traveller with piles of money might consider a round the world flight, broken up with stays in many of these hotels. & \{{ \it  }\} \\ \midrule 
 3 & The media companies routinely lie about the purpose of this, claiming that it is to "prevent piracy". & \{{ \it media, co, piravy }\} \\ \midrule 
 3 & In fact, region codes have absolutely no effect on illegal copying; a bit-for-bit copy of a disk will play just fine on any device where the original will. & \{{ \it codes, bit, bit, disk, device }\} \\ \midrule 
 3 & The actual purpose is to give those companies more control over their markets; it's all about money spinning. & \{{ \it co }\} \\ \midrule 
 3 & In 1994, the ethnically Armenian Nagorno-Karabakh region of Azerbaijan waged war against the Azeris. & \{{ \it ethnically Armenian, Nagorno-Karabakh, Azerbaijan, Azeris }\} \\ \midrule 
 3 & With Armenian backing, a new republic was created. However, no established nation - not even Armenia - officially recognizes it. & \{{ \it Armenian, republic, nation, Armenia }\} \\ \midrule 
 3 & Diplomatic arguments over the region continue to mar relations between Armenia and Azerbaijan. & \{{ \it diplomatic, Armenia, Azerbaijan }\} \\ \midrule 
 3 & The unmanned lunar orbiter Chandrayaan-1 ejected its Moon Impact Probe (MIP), which hurtled across the surface of the Moon at 1.5 kilometres per second (3000 miles per hour), and successfully crash landed near the Moon's south pole. & \{{ \it orbiter, Chandrayaan, impact probe, MIP, kilometers per sec, mph, pole }\} \\ \midrule 
 3 & Besides carrying three important scientific instruments, the lunar probe also carried the image of the Indian national flag, painted on all sides. & \{{ \it instruments }\} \\
        \bottomrule
    \end{tabular}
    \label{tab:annotations55}
\end{table*}

\begin{table*}[]
    \centering
    \fontsize{8.4}{10}\selectfont
    \begin{tabular}{cp{10.5cm}p{3.5cm}}
        \toprule
        \textbf{signer} & \textbf{reference} & \textbf{fingerspelled terms} \\
        \midrule
 3 & The Drukgyal Dzong is a ruined fortress and Buddhist monastery in the upper part of the Paro District (in Phondey Village). & \{{ \it Drukgyal Dzong, Paro, Phondey }\} \\ \midrule 
 3 & It is said that in 1649, Zhabdrung Ngawang Namgyel created the fortress to commemorate his victory against the Tibetan-Mongol forces. & \{{ \it Zhabdrung Ngawang Namgyel }\} \\ \midrule 
 3 & In 1951, a fire caused for only some of the relics of the Drukgyal Dzong to remain, such as the image of Zhabdrung Ngawang Namgyal. & \{{ \it relics, Drukgyal Dzong, Zhabdrung Ngawang Namgyal }\} \\ \midrule 
 3 & After the fire, the fortress was preserved and protected, remaining to be one of Bhutan's most sensational attractions. & \{{ \it  }\} \\ \midrule 
 3 & Mysticism is the pursuit of communion with, identity with, or conscious awareness of an ultimate reality, divinity, spiritual truth, or God. & \{{ \it mysticism, reality, divinity }\} \\ \midrule 
 3 & The believer seeks a direct experience, intuition, or insight into divine reality/the deity or dieties. & \{{ \it divine }\} \\ \midrule 
 3 & Followers pursue certain ways of living, or practices that are intended to nurture those experiences. & \{{ \it  }\} \\ \midrule 
 3 & Mysticism can be distinguished from other forms of religious belief and worship by its emphasis on the direct personal experience of a unique state of consciousness, particularly those of a peaceful, insightful, blissful, or even ecstatic character. & \{{ \it mysticism }\} \\ \midrule 
 3 & The aircraft had been headed to Irkutsk and was being operated by interior troops. & \{{ \it Irkutsk }\} \\ \midrule 
 3 & An inquiry was established to investigate. & \{{ \it  }\} \\ \midrule 
 3 & The Il-76 has been a major component of both the Russian and Soviet military since the 1970s, and had already seen a serious accident in Russia last month. & \{{ \it Il-76, Soviet, 1970s }\} \\ \midrule 
 3 & On October 7 an engine separated on takeoff, without injuries. Russia briefly grounded Il-76s after that accident. & \{{ \it Oct, Il-76 }\} \\ \midrule 
 3 & We know many Greek politicians, scientists, and artists. Possibly the most known person of this culture is Homer, the legendary blind poet, who composed two masterpieces of Greek literature: the poems Iliad and Odyssey. & \{{ \it Homer, lit, Iliad, Odyssey }\} \\ \midrule 
 3 & Sophocles and Aristophanes are still popular playwrights and their plays are considered to be among the greatest works of world literature. & \{{ \it Sophocles, Aristophanes, lit }\} \\ \midrule 
 3 & Another famous Greek is a mathematician Pythagoras, mostly known for his famous theorem about relations of the sides of right triangles. & \{{ \it Pythagoras }\} \\ \midrule 
 3 & An immigration checkpoint is usually the first stop when disembarking from a plane, a ship, or another vehicle. & \{{ \it  }\} \\ \midrule 
 3 & In some cross-border trains inspections are done on the running train and you should have valid ID with you when boarding one of those trains. & \{{ \it ID }\} \\ \midrule 
 3 & On night sleeper trains, passports may be collected by the conductor so that you do not have your sleep interrupted. & \{{ \it  }\} \\ \midrule 
 3 & Vikings used the Russian waterways to get to the Black Sea and Caspian Sea. Parts of these routes can still be used. Check possible need for special permits, which may be difficult to get. & \{{ \it Sea, Capsian Sea }\} \\ \midrule 
 3 & The White Sea–Baltic Canal connects the Arctic Ocean to the Baltic Sea, via Lake Onega, Lake Ladoga and Saint Petersburg, mostly by rivers and lakes. & \{{ \it Sea, Baltic Canal, Arctic, Baltic Canal, Lake Onega, Lake Ladoga }\} \\ \midrule 
 3 & Lake Onega is also connected to Volga, so coming from the Caspian Sea through Russia is still possible. & \{{ \it Lake Onega, Volga, Caspian }\} \\ \midrule 
 3 & MetroRail has two classes on commuter trains in and around Cape Town: MetroPlus (also called First Class) and Metro (called Third Class). & \{{ \it Cape Town, MetroRail, Metro, Metro }\} \\ \midrule 
 3 & MetroPlus is more comfortable and less crowded but slightly more expensive, though still cheaper than normal metro tickets in Europe. & \{{ \it Metro, metro }\} \\ \midrule 
 3 & Every train has both MetroPlus and Metro coaches; the MetroPlus coaches are always on the end of the train nearest Cape Town. & \{{ \it Metro, Metro, Metro }\} \\ \midrule 
 3 & There are many things you have to take into consideration before and when you travel somewhere. & \{{ \it  }\} \\ \midrule 
 3 & When you travel, expect things not to be like they are "back home". Manners, laws, food, traffic, lodging, standards, language and so on will to some extent differ from where you live. & \{{ \it  }\} \\
        \bottomrule
    \end{tabular}
    \label{tab:annotations56}
\end{table*}

\begin{table*}[]
    \centering
    \fontsize{8.4}{10}\selectfont
    \begin{tabular}{cp{10.5cm}p{3.5cm}}
        \toprule
        \textbf{signer} & \textbf{reference} & \textbf{fingerspelled terms} \\
        \midrule
 3 & This is something you always need to keep in mind, to avoid disappointment or perhaps even distaste over local ways to do things. & \{{ \it  }\} \\ \midrule 
 3 & Nadia, born on September 17, 2007, by Cesarean section at a maternity clinic in Aleisk, Russia, weighed in at a massive 17 pounds 1 ounce. & \{{ \it Sept, Nadia, Aleisk, oz }\} \\ \midrule 
 3 & "We were all simply in shock," the mother stated. & \{{ \it  }\} \\ \midrule 
 3 & When asked what the father said, she answered "He couldn't say a thing - he just stood there blinking." & \{{ \it  }\} \\ \midrule 
 3 & Wildlife is among the most challenging motifs for a photographer, and needs a combination of good luck, patience, experience and good equipment. & \{{ \it wildlife }\} \\ \midrule 
 3 & Wildlife photography is often taken for granted, but like photography in general, a picture is worth a thousand words. & \{{ \it wildlife }\} \\ \midrule 
 3 & Wildlife photography often requires a long telephoto lens, though things like a flock of birds or a tiny creature need other lenses. & \{{ \it wildlife }\} \\ \midrule 
 3 & Many exotic animals are hard to find, and parks sometimes have rules about taking photographs for commercial purposes. & \{{ \it exotic, park }\} \\ \midrule 
 3 & Wild animals might either be shy or aggressive. The environment might be cold, hot, or otherwise hostile. & \{{ \it wild }\} \\ \midrule 
 3 & This is believed to be the fifth time in history that people have observed what turned out to be chemically confirmed martian material falling to Earth. & \{{ \it martian }\} \\ \midrule 
 3 & Out of the approximately 24,000 known meteorites to have fallen to Earth, only about 34 have been verified to be martian in origin. & \{{ \it meteorites, martian }\} \\ \midrule 
 3 & Fifteen of these rocks are attributed to the meteorite shower last July. & \{{ \it meteorite shower, July }\} \\ \midrule 
 3 & Some of the rocks, which are very rare on Earth, are being sold from US\$11,000 to \$22,500 per ounce, which is about ten times more than the cost of gold. & \{{ \it rare, per oz }\} \\ \midrule 
 3 & The easiest way to get started in the business of travel writing is to hone your skills on an established travel blog website. & \{{ \it blog }\} \\ \midrule 
 3 & After you become comfortable with formatting and editing on the web, then later, you might create your own website. & \{{ \it formatting, own }\} \\ \midrule 
 3 & Children develop an awareness of race and racial stereotypes quite young and these racial stereotypes affect behavior. & \{{ \it race, racial, racial }\} \\ \midrule 
 3 & For instance, children who identify with a racial minority that is stereotyped as not doing well in school tend to not do well in school once they learn about the stereotype associated with their race. & \{{ \it racial minority }\} \\ \midrule 
 3 & The celebrations started with a special show by the world-renowned group Cirque du Soleil. & \{{ \it  }\} \\ \midrule 
 3 & It was followed by the Istanbul State Symphony Orchestra, a Janissary band, and the singers Fatih Erkoç and Müslüm Gürses. & \{{ \it Janissary, Fatih Erkoc, Muslum Gurses }\} \\ \midrule 
 3 & Then Whirling Dervishes took to the stage. & \{{ \it Whirling Dervishes }\} \\ \midrule 
 3 & Turkish diva Sezen Aksu performed with the Italian tenor Alessandro Safina and Greek singer Haris Alexiou. & \{{ \it diva Sezen Aksu, tenor, Alessandro Safina, Haris Alexiou }\} \\ \midrule 
 3 & To finish, Turkish dance group Fire of Anatolia performed the show "Troy". & \{{ \it Fire of Anatolia, Troy }\} \\ \midrule 
 3 & The Nervous System maintains homeostasis by sending nerve impulses through out the body to keep the flow of blood going as well as undisturbed. & \{{ \it homeostasis }\} \\ \midrule 
 3 & These nerve impulses can be sent so quickly throughout the body which helps keep the body safe from any potential threat. & \{{ \it  }\} \\ \midrule
4 & Leslie Aun, a spokesperson for the Komen Foundation, said the organization adopted a new rule that does not allow grants or funding to be awarded to organizations that are under legal investigation. & \{{ \it Leslie Aun, Komen, Komen, grants }\} \\ \midrule 
 4 & Komen's policy disqualified Planned Parenthood due to a pending investigation on how Planned Parenthood spends and reports its money that is being conducted by Representative Cliff Stearns. & \{{ \it Komen, disq, hood, Cliff Stearns }\} \\ \midrule 
 4 & Stearns is investigating whether taxes are used to fund abortions through Planned Parenthood in his role as chairman of the Oversight and Investigations Subcommittee, which is under the umbrella of the House Energy and Commerce Committee. & \{{ \it Stearns, taxes, hood, of, sub }\} \\
        \bottomrule
    \end{tabular}
    \label{tab:annotations57}
\end{table*}

\begin{table*}[]
    \centering
    \fontsize{8.4}{10}\selectfont
    \begin{tabular}{cp{10.5cm}p{3.5cm}}
        \toprule
        \textbf{signer} & \textbf{reference} & \textbf{fingerspelled terms} \\
        \midrule
 4 & Internationally, there are nearly 200 running tour organizations. Most of them operate independently. & \{{ \it  }\} \\ \midrule 
 4 & The Global Running Tours successor, Go Running Tours networks dozens of sightrunning providers on four continents. & \{{ \it Tours, Go, continents }\} \\ \midrule 
 4 & With roots in Barcelona's Running Tours Barcelona and Copenhagen's Running Copenhagen, it was quickly joined by Running Tours Prague based in Prague and others. & \{{ \it Tours, Copenhagen, Prague }\} \\ \midrule 
 4 & Iraq's Abu Ghraib prison has been set alight during a riot. & \{{ \it Iraq, Abu Ghraib }\} \\ \midrule 
 4 & The jail became notorious after prisoner abuse was discovered there after US forces took over. & \{{ \it out, US }\} \\ \midrule 
 4 & Tours are cheaper for larger groups, so if you're by yourself or with just one friend, try to meet other people and form a group of four to six for a better per-person rate. & \{{ \it tours }\} \\ \midrule 
 4 & However, this shouldn't really be off your concern, because often tourists are shuffled around to fill the cars. & \{{ \it van }\} \\ \midrule 
 4 & It seems actually to be more a way of tricking people into believing they have to pay more. & \{{ \it  }\} \\ \midrule 
 4 & The PBS show has more than two-dozen Emmy awards, and its run is shorter only than Sesame Street and Mister Rogers' Neighborhood. & \{{ \it Emmy, Sesame Street, Mister Rogers' }\} \\ \midrule 
 4 & Each episode of the show would focus on a theme in a specific book and then explore that theme through multiple stories. & \{{ \it episode, theme }\} \\ \midrule 
 4 & Each show would also provide recommendations for books that children should look for when they went to their library. & \{{ \it theme }\} \\ \midrule 
 4 & John Grant, from WNED Buffalo (Reading Rainbow's home station) said "Reading Rainbow taught kids why to read,... the love of reading — [the show] encouraged kids to pick up a book and to read." & \{{ \it so, John Grant, WNED, station }\} \\ \midrule 
 4 & It is believed by some, including John Grant, that both the funding crunch and a shift in the philosophy of educational television programming contributed to ending the series. & \{{ \it John Grant, crunch, ed, TV }\} \\ \midrule 
 4 & Hiking is an outdoor activity which consists of walking in natural environments, often on hiking trails. & \{{ \it hiking, hiking trails }\} \\ \midrule 
 4 & Day hiking involves distances of less than a mile up to longer distances that can be covered in a single day. & \{{ \it mile }\} \\ \midrule 
 4 & For a day hike along an easy trail little preparations are needed, and any moderately fit person can enjoy them. & \{{ \it fit, fit }\} \\ \midrule 
 4 & Families with small children may need more preparations, but a day outdoors is easily possible even with babies and pre-school children. & \{{ \it  }\} \\ \midrule 
 4 & If you want to see the world on the cheap, for necessity, lifestyle or challenge, there are some ways to do that. & \{{ \it style, or }\} \\ \midrule 
 4 & Basically, they fall into two categories: Either work while you travel or try and limit your expenses. This article is focused on the latter. & \{{ \it  }\} \\ \midrule 
 4 & For those willing to sacrifice comfort, time and predictability to push expenses down close to zero, see minimum budget travel. & \{{ \it or }\} \\ \midrule 
 4 & The advice assumes that travellers do not steal, trespass, participate in the illegal market, beg, or otherwise exploit other people for their own gain. & \{{ \it trespass, illegal }\} \\ \midrule 
 4 & Carrying for others - Never let your bags out of your sight, especially when you are crossing international borders. & \{{ \it or }\} \\ \midrule 
 4 & You could find yourself being used as a drug carrier without your knowledge, which will land you in a great deal of trouble. & \{{ \it drug carrier }\} \\ \midrule 
 4 & This includes waiting in line, as drug-sniffing dogs could be used at any time without notice. & \{{ \it dogs, drugs }\} \\ \midrule 
 4 & Some countries have exceedingly draconian punishments even for first time offenses; these can include prison sentences of over 10 years or death. & \{{ \it or }\} \\ \midrule 
 4 & Unattended bags are a target for theft and can also attract attention from authorities wary of bomb threats. & \{{ \it so, or, bomb }\} \\ \midrule 
 4 & The city of Beijing in China will be the host city of the Olympic Winter Games in 2022, which will make it the first city to have hosted both the summer and winter Olympics. & \{{ \it Beijing }\} \\ \midrule 
 4 & Beijing will host the opening and closing ceremonies and the indoor ice events. & \{{ \it Beijing, ice }\} \\
        \bottomrule
    \end{tabular}
    \label{tab:annotations58}
\end{table*}

\begin{table*}[]
    \centering
    \fontsize{8.4}{10}\selectfont
    \begin{tabular}{cp{10.5cm}p{3.5cm}}
        \toprule
        \textbf{signer} & \textbf{reference} & \textbf{fingerspelled terms} \\
        \midrule
 4 & Other skiing events will be at the Taizicheng ski area in Zhangjiakou, about 220 km (140 miles) from Beijing. & \{{ \it km, miles, Beijing, Taizicheng, Zhangjiakou }\} \\ \midrule 
 4 & Scientists have discovered landforms littered across the moon's surface called lobate scarps that have apparently resulted from the moon's shrinking very slowly. & \{{ \it landforms, lobate scarps }\} \\ \midrule 
 4 & These scarps were found all over the moon and appear to be minimally weathered, indicating that the geologic events that created them were fairly recent. & \{{ \it scarps, geological }\} \\ \midrule 
 4 & This theory contradicts the claim that the moon is completely devoid of geologic activity. & \{{ \it geological }\} \\ \midrule 
 4 & Organisers of the protest said about 100,000 people turned up in German cities such as Berlin, Cologne, Hamburg, and Hanover. & \{{ \it of, Berlin, Cologne, Hamburg, Hanover }\} \\ \midrule 
 4 & In Berlin, police estimated 6,500 protestors. & \{{ \it  }\} \\ \midrule 
 4 & Protests also took place in Paris, Sofia in Bulgaria, Vilnius in Lithuania, Valetta in Malta, Tallinn in Estonia, and Edinburgh and Glasgow in Scotland. & \{{ \it Paris, Sofia, Vilnius, Lithuania, Valetta, Malta, Tallinn, Estonia, Edinburgh, Glasgow }\} \\ \midrule 
 4 & In London, about 200 people protested outside some major copyright holders' offices. & \{{ \it London, office }\} \\ \midrule 
 4 & Last month, there were major protests in Poland when that country signed ACTA, which has led to the Polish government deciding not to ratify the agreement, for now. & \{{ \it ACTA }\} \\ \midrule 
 4 & Latvia and Slovakia have both delayed the process of joining ACTA. & \{{ \it Latvia, Slovakia, ACTA }\} \\ \midrule 
 4 & "Thanks for those who supported a convict like me," Siriporn was quoted as saying at a press conference. & \{{ \it convict, Siriporn, press }\} \\ \midrule 
 4 & "Some may not agree but I don't care. & \{{ \it  }\} \\ \midrule 
 4 & I am happy that there are people willing to support me. & \{{ \it  }\} \\ \midrule 
 4 & Jerusalem is the capital and largest city of Israel, though most other countries and the United Nations do not recognize it as Israel's capital. & \{{ \it Jerusalem, UN, United }\} \\ \midrule 
 4 & The ancient city in the Judean Hills has a fascinating history spanning thousands of years. & \{{ \it Judean Hills }\} \\ \midrule 
 4 & The city is holy to the three monotheistic religions - Judaism, Christianity and Islam, and serves as a spiritual, religious, and cultural center. & \{{ \it monotheistic, Judaism, Christianity, Islam }\} \\ \midrule 
 4 & Due to the religious significance of the city, and in particular the many sites of the Old City area, Jerusalem is one of the main tourist destinations in Israel. & \{{ \it of, city }\} \\ \midrule 
 4 & Jerusalem has many historic, archeological and cultural sites, along with vibrant and crowded shopping centers, cafés, and restaurants. & \{{ \it archeological, sites, vibrant }\} \\ \midrule 
 4 & "Panama Papers" is an umbrella term for roughly ten million documents from Panamanian law firm Mossack Fonseca, leaked to the press in spring 2016. & \{{ \it Panama, firm, Mossack Fonseca, press }\} \\ \midrule 
 4 & The documents showed fourteen banks helped wealthy clients hide billions of US dollars of wealth to avoid taxes and other regulations. & \{{ \it banks, of, US, taxes, fees }\} \\ \midrule 
 4 & British newspaper The Guardian suggested Deutsche Bank controlled roughly a third of the 1200 shell companies used to accomplish this. & \{{ \it The Guardian, Bank, of, shell }\} \\ \midrule 
 4 & There were protests worldwide, several criminal prosecutions, and the leaders of the governments of Iceland and Pakistan both resigned. & \{{ \it so, criminal, of }\} \\ \midrule 
 4 & Stars give off light and heat because of the energy made when hydrogen atoms are merged (or fused) together to form heavier elements. & \{{ \it hydrogen atoms, elements }\} \\ \midrule 
 4 & Scientists are working to create a reactor that can make energy in the same way. & \{{ \it reactor, react }\} \\ \midrule 
 4 & This, however, is a very hard problem to solve and will take many years before we see useful fusion reactors built. & \{{ \it fusion reactor }\} \\ \midrule 
 4 & You should choose your Frequent Flyer airline in an alliance carefully. & \{{ \it alliance }\} \\ \midrule 
 4 & Although you may think it is intuitive to join the airline you fly most, you should be aware that privileges offered are often different and frequent flyer points may be more generous under a different airline in the same alliance. & \{{ \it  }\} \\ \midrule 
 4 & The most common cause for accidents in winter is slippery roads, pavements (sidewalks) and especially steps. & \{{ \it roads, sidewalk }\} \\ \midrule 
 4 & At a minimum, you need footwear with suitable soles. Summer shoes are usually very slippery on ice and snow, even some winter boots are deficient. & \{{ \it ice, boots, boots }\} \\
        \bottomrule
    \end{tabular}
    \label{tab:annotations59}
\end{table*}

\begin{table*}[]
    \centering
    \fontsize{8.4}{10}\selectfont
    \begin{tabular}{cp{10.5cm}p{3.5cm}}
        \toprule
        \textbf{signer} & \textbf{reference} & \textbf{fingerspelled terms} \\
        \midrule
4 & The pattern should be deep enough, 5 mm (1/5 inch) or more, and the material soft enough in cold temperatures. & \{{ \it mm, inch }\} \\ \midrule 
 4 & Some boots have studs and there is studded add-on equipment for slippery conditions, suitable for most shoes and boots, for the heels or heels and sole. & \{{ \it studs, or, heels, soles }\} \\ \midrule 
 4 & Heels should be low and wide. Sand, gravel or salt (calcium chloride) is often scattered on roads or paths to improve traction. & \{{ \it gravel, calcium chloride, side }\} \\ \midrule 
 4 & While authorities are unsure of the credibility of the threat, the Maryland Transportaion Authority made the closure with the urging of the FBI. & \{{ \it MD, FBI, MD }\} \\ \midrule 
 4 & Dump trucks were used to block tube entrances and assistance of 80 police were on hand to direct motorists to detours. & \{{ \it route, detour }\} \\ \midrule 
 4 & There were no heavy traffic delays reported on the beltway, the city's alternate route. & \{{ \it  }\} \\ \midrule 
 4 & Regarding the global financial situation, Zapatero continued by saying that "the financial system is a part of the economy, a crucial part. & \{{ \it Zapatero }\} \\ \midrule 
 4 & We have a year-long financial crisis, which has had its most acute moment in the past two months, and I think now the financial markets are beginning to recover." & \{{ \it  }\} \\ \midrule 
 4 & Jarque was practicing during pre-season training at Coverciano in Italy earlier in the day. & \{{ \it Jarque, Coverciano }\} \\ \midrule 
 4 & He was staying in the team hotel ahead of a match planned for Sunday against Bolonia. & \{{ \it Bolonia }\} \\ \midrule 
 4 & Passengers were given water as they waited in 90(F)-degree heat. & \{{ \it passengers }\} \\ \midrule 
 4 & Fire captain Scott Kouns said, "It was a hot day in the Santa Clara with temperatures in the 90s. & \{{ \it Santa Clara, Scott Kouns }\} \\ \midrule 
 4 & Any length of time trapped on a roller coaster would be uncomfortable, to say the least, and it took at least an hour to get the first person off the ride." & \{{ \it  }\} \\ \midrule 
 4 & Leonov, also known as "cosmonaut No. 11", was part of the Soviet Union's original team of cosmonauts. & \{{ \it Leonov, cosmonaut, of, Soviet Union, of }\} \\ \midrule 
 4 & On March 18, 1965, he performed the first manned extravehicular activity (EVA), or "spacewalk", remaining alone outside the spacecraft for just over twelve minutes. & \{{ \it March, extravehicular, EVA, spacewalk, of }\} \\ \midrule 
 4 & He received the "Hero of the Soviet Union", the Soviet Union's highest honor, for his work. & \{{ \it of, Soviet Union }\} \\ \midrule 
 4 & Ten years later, he led the Soviet part of the Apollo–Soyuz mission symbolizing that the Space Race was over. & \{{ \it Soviet, Apollo-Soyuz, mission, space }\} \\ \midrule 
 4 & While San Francisco's economy is linked to it being a world-class tourist attraction, its economy is diversified. & \{{ \it SF, SF }\} \\ \midrule 
 4 & The largest employment sectors are professional services, government, finance, trade, and tourism. & \{{ \it tourism }\} \\ \midrule 
 4 & Its frequent portrayal in music, films, literature and popular culture has helped make the city and its landmarks known throughout the world. & \{{ \it SF }\} \\ \midrule 
 4 & San Francisco has developed a large tourist infrastructure with numerous hotels, restaurants, and top-notch convention facilities. & \{{ \it tourist, hotels, SF }\} \\ \midrule 
 4 & San Francisco is also one of the best places in the nation for other Asian cuisine: Korean, Thai, Indian, and Japanese. & \{{ \it SF }\} \\ \midrule 
 4 & Getting into your own car and heading off on a long road trip has an intrinsic appeal in its simplicity. & \{{ \it  }\} \\ \midrule 
 4 & Unlike larger vehicles, you are probably already familiar with driving your car and know its limitations. & \{{ \it  }\} \\ \midrule 
 4 & Setting up a tent on private property or in a town of any size can easily attract unwanted attention. & \{{ \it if }\} \\ \midrule 
 4 & In short, using your car is a great way to take a road trip but seldom in itself a way to "camp". & \{{ \it  }\} \\ \midrule 
 4 & Car camping is possible if you have a large minivan, SUV, Sedan or Station Wagon with seats that lay down. & \{{ \it if, minivan, SUV, sedan }\} \\ \midrule 
 4 & Accompanied by those officials, he assured Texas citizens that steps were being taken to protect the public's safety. & \{{ \it  }\} \\ \midrule 
 4 & Perry specifically said, "There are few places in the world better equipped to meet the challenge that is posed in this case." & \{{ \it Perry, of }\} \\
        \bottomrule
    \end{tabular}
    \label{tab:annotations60}
\end{table*}

\begin{table*}[]
    \centering
    \fontsize{8.4}{10}\selectfont
    \begin{tabular}{cp{10.5cm}p{3.5cm}}
        \toprule
        \textbf{signer} & \textbf{reference} & \textbf{fingerspelled terms} \\
        \midrule
 4 & The governor also stated, "Today, we learned that some school aged children have been identified as having had contact with the patient." & \{{ \it  }\} \\ \midrule 
 4 & He went on to say, "This case is serious. Rest assured that our system is working as well as it should." & \{{ \it case }\} \\ \midrule 
 4 & Hostels cater primarily to young people – a typical guest is in their twenties – but you can often find older travellers there too. & \{{ \it hostels, hotels, hostels }\} \\ \midrule 
 4 & Families with children are a rare sight, but some hostels do allow them in private rooms. & \{{ \it rare, hostels }\} \\ \midrule 
 4 & Lodin also said officials decided to cancel the runoff in order to save Afghans the expense and security risk of another election. & \{{ \it Lodin, Afghans, security risk }\} \\ \midrule 
 4 & Diplomats said that they had found enough ambiguity in the Afghan constitution to determine the runoff as unnecessary. & \{{ \it diplomat, Afghan }\} \\ \midrule 
 4 & This contradicts earlier reports, which said that cancelling the runoff would have been against the constitution. & \{{ \it  }\} \\ \midrule 
 4 & If you want to fly around the world completely in the Southern Hemisphere, the choice of flights and destinations is limited due to the lack of transoceanic routes. & \{{ \it hemisphere, transoceanic route }\} \\ \midrule 
 4 & No airline alliance covers all three ocean crossings in the Southern Hemisphere (and SkyTeam covers none of the crossings). & \{{ \it hemisphere, sky }\} \\ \midrule 
 4 & However, Star Alliance covers everything except the eastern South Pacific from Santiago de Chile to Tahiti, which is a LATAM Oneworld flight. & \{{ \it so, Star Alliance, Pacific, Santiago de Chile, Tahiti, LATAM }\} \\ \midrule 
 4 & This flight is not the only option if you want to skip the South Pacific and the west coast of South America. (see below) & \{{ \it Pacific, coast, of, map }\} \\ \midrule 
 4 & However, in June 1956, Krushchev's promises were put to the test when riots in Poland, where workers were protesting against food shortages and wage cuts, turned into a general protest against Communism. & \{{ \it June, Khrushchev, wage, communism }\} \\ \midrule 
 4 & Although in the end, Krushchev sent in tanks to restore order, he did give way to some economic demands and agreed to appoint the popular Wladyslaw Gomulka as the new prime minister. & \{{ \it so, Khrushchev, tank, order, Wladyslaw Gomulka, prime minister }\} \\ \midrule 
 4 & She gained critical acclaim during her time in Atlanta and was recognized for innovative urban education. & \{{ \it ed }\} \\ \midrule 
 4 & In 2009 she was awarded the title of National Superintendent of the Year. & \{{ \it supt, of }\} \\ \midrule 
 4 & At the time of the award, Atlanta schools had seen a large improvement on test scores. & \{{ \it  }\} \\ \midrule 
 4 & Shortly after, The Atlanta Journal-Constitution published a report showing problems with test results. & \{{ \it Journal }\} \\ \midrule 
 4 & The report showed test scores had increased implausibly fast, and alleged the school internally detected problems but didn't act on the findings. & \{{ \it  }\} \\ \midrule 
 4 & Evidence thereafter indicated test papers were tampered with Hall, along with 34 other education officials, was indicted in 2013. & \{{ \it Hall, so, Hall, ed }\} \\ \midrule 
 4 & One of strong points of Charlotte, North Carolina, is that it has an abundance of high-quality options for families. & \{{ \it Charlotte, NC }\} \\ \midrule 
 4 & Residents from other areas often cite family-friendliness as a primary reason for moving there, and visitors often find the city easy to enjoy with children around. & \{{ \it  }\} \\ \midrule 
 4 & In the past 20 years, the amount of child-friendly options in Uptown Charlotte has grown exponentially. & \{{ \it  }\} \\ \midrule 
 4 & Taxis are not generally used by families in Charlotte, though they can be of some use in certain circumstances. & \{{ \it taxi, by, so, be, taxi }\} \\ \midrule 
 4 & There is a surcharge for having more than 2 passengers, so this option might be more expensive than necessary. & \{{ \it taxi, if }\} \\ \midrule 
 4 & "Luckily nothing happened to me, but I saw a macabre scene, as people tried to break windows in order to get out. & \{{ \it  }\} \\ \midrule 
 4 & People were hitting the panes with chairs, but the windows were unbreakable. & \{{ \it  }\} \\ \midrule 
 4 & One of the panes finally broke, and they started to get out by the window," said survivor Franciszek Kowal. & \{{ \it Franciszek Kowal }\} \\ \midrule 
 4 & 21-year-old Jesus joined Manchester City last year in January 2017 from Brazilian club Palmeiras for a reported fee of £27 million. & \{{ \it Jesus, Manchester City, Jan, club, Palmeiras, fee, of }\} \\
        \bottomrule
    \end{tabular}
    \label{tab:annotations61}
\end{table*}

\begin{table*}[]
    \centering
    \fontsize{8.4}{10}\selectfont
    \begin{tabular}{cp{10.5cm}p{3.5cm}}
        \toprule
        \textbf{signer} & \textbf{reference} & \textbf{fingerspelled terms} \\
        \midrule
 4 & Since then, the Brazilian has featured in 53 matches for the club in all competitions and has scored 24 goals. & \{{ \it club, goals }\} \\ \midrule 
 4 & Peter Lenz, a 13-year-old motorcycle racer, has died after being involved in a crash at the Indianapolis Motor Speedway. & \{{ \it Peter Lenz, Indianapolis, motor, way }\} \\ \midrule 
 4 & While on his warm-up lap, Lenz fell off his bike, and was then struck by fellow racer Xavier Zayat. & \{{ \it up, by, Xavier Zayat }\} \\ \midrule 
 4 & He was immediately attended to by the on-track medical staff and transported to a local hospital where he later died. & \{{ \it  }\} \\ \midrule 
 4 & Zayat was unhurt in the accident. & \{{ \it  }\} \\ \midrule 
 4 & Permits must be reserved in advance. You must have a permit to stay overnight at Sirena. & \{{ \it Sirena }\} \\ \midrule 
 4 & Sirena is the only ranger station that offers dormitory lodging and hot meals in addition to camping. La Leona, San Pedrillo, and Los Patos offer only camping with no food service. & \{{ \it Sirena, ranger station, La Leona, San Pedrillo, Los Patos }\} \\ \midrule 
 4 & It is possible to secure park permits directly from the Ranger Station in Puerto Jiménez, but they do not accept credit cards & \{{ \it park, ranger station, Puerto Jimenez }\} \\ \midrule 
 4 & The Park Service (MINAE) does not issue park permits more than one month in advance of anticipated arrival. & \{{ \it park, MINAE, park, date }\} \\ \midrule 
 4 & CafeNet El Sol offers a reservation service for a fee of US\$30, or \$10 for one-day passes; details on their Corcovado page. & \{{ \it CafeNet El Sol, fee, or, pass, details, Corcovado }\} \\ \midrule 
 4 & The Irish government is stressing the urgency of parliamentary legislation to rectify the situation. & \{{ \it  }\} \\ \midrule 
 4 & "It is now important from both a public health and criminal justice perspective that the legislation be enacted as soon as possible", said a government spokesperson. & \{{ \it criminal justice }\} \\ \midrule 
 4 & The Health Minister expressed concern both for the welfare of individuals taking advantage of the temporary legality of the substances involved, and for drug-related convictions handed down since the now-unconstitutional changes came into effect. & \{{ \it minister, welfare, of, of, effect }\} \\ \midrule 
 4 & Traditionally, the heir to the throne would go straight into the military after finishing school. & \{{ \it heir, throne }\} \\ \midrule 
 4 & However, Charles went to university at Trinity College, Cambridge where he studied Anthropology and Archaeology, and later History, earning a 2:2 (a lower second class degree). & \{{ \it Charles, at, Trinity, Cambridge, he, anthropology, archaeology, score }\} \\ \midrule 
 4 & Charles was the first member of the British Royal Family to be awarded a degree. & \{{ \it so, Charles, of }\} \\ \midrule 
 4 & Yes! King Tutankhamun, sometimes referred to as "King Tut" or "The Boy King", is one of the most well known ancient Egyptian kings in modern times. & \{{ \it Tutankhamun, Tut, or, Tut }\} \\ \midrule 
 4 & Interestingly, he was not considered to be very important in ancient times and was not recorded on most ancient king lists. & \{{ \it  }\} \\ \midrule 
 4 & However, the discovery of his tomb in 1922 made him a celebrity. While many tombs of the past were robbed, this tomb was left virtually undisturbed. & \{{ \it tomb }\} \\ \midrule 
 4 & Most of the items buried with Tutankhamun have been well preserved, including thousands of artefacts made from precious metals and rare stones. & \{{ \it of, artifacts, rare }\} \\ \midrule 
 4 & Tropical Storm Danielle, fourth named storm of the 2010 Atlantic hurricane season, has formed in the eastern Atlantic Ocean. & \{{ \it tropical storm, Danielle, Atlantic, Atlantic Ocean }\} \\ \midrule 
 4 & The storm, located approximately 3,000 miles from Miami, Florida, has maximum sustained winds of 40 mph (64 kph). & \{{ \it miles, Miami, Fla, mph, kph }\} \\ \midrule 
 4 & Scientists at the National Hurricane Center predict that Danielle will strengthen to a hurricane by Wednesday. & \{{ \it Danielle, by }\} \\ \midrule 
 4 & As the storm is far from landfall, it remains difficult to assess potential impact to the United States or Caribbean. & \{{ \it US }\} \\ \midrule 
 4 & Nadal's head to head record against the Canadian is 7–2. & \{{ \it Nadal's }\} \\ \midrule 
 4 & He recently lost against Raonic in the Brisbane Open. & \{{ \it Raonic, Brisbane }\} \\ \midrule 
 4 & Nadal bagged 88\% net points in the match winning 76 points in the first serve. & \{{ \it Nadal }\} \\ \midrule 
 4 & After the match, King of Clay said, "I am just excited about being back in the final rounds of the most important events. I am here to try to win this." & \{{ \it of, clay, of }\} \\
        \bottomrule
    \end{tabular}
    \label{tab:annotations62}
\end{table*}

\end{document}